\documentclass{article}

 \usepackage[main, final]{neurips_2026}

\usepackage[utf8]{inputenc} 
\usepackage[T1]{fontenc}    
\usepackage{hyperref}       
\usepackage{url}            
\usepackage{booktabs}       
\usepackage{amsfonts}       
\usepackage{nicefrac}       
\usepackage{microtype}      
\usepackage{xcolor}         
\usepackage{graphicx}

\usepackage{amsmath}
\usepackage{amssymb}
\usepackage{mathtools}
\usepackage{amsthm}
\usepackage{algorithm}
\usepackage{algpseudocode}
\usepackage[capitalize,noabbrev]{cleveref}
\usepackage{subcaption}

\theoremstyle{plain}
\newtheorem{theorem}{Theorem}[section]
\newtheorem{proposition}[theorem]{Proposition}

\theoremstyle{definition}

\theoremstyle{remark}

\title{Rethinking Ratio-Based Trust Regions for Policy Optimization in Multi-Agent Reinforcement Learning}

\author{%
  Chulabhaya Wijesundara \\
  Northeastern University; STR \\  
  \And
  Andrea Baisero \\ University of California, Irvine \\  
  \And
  Zhongheng Li \\
  STR \\  
  \AND
  Gregory Castañón \\
  STR \\  
  \And
  Alan Carlin \\
  STR \\  
  \And
  Christopher Amato \\
  Northeastern University \\  
}

\begin{document}
\maketitle
\begin{abstract}
Centralized training with decentralized execution (CTDE) is a standard framework for cooperative multi-agent policy-gradient reinforcement learning, allowing agents to learn from joint information while acting from local observations. Ratio-based trust-region methods such as Multi-Agent Proximal Policy Optimization (MAPPO) and Multi-Agent Simple Policy Optimization (MASPO) update decentralized actors using per-agent probability ratios weighted by joint advantage estimates. Teammate non-stationarity increases the variance of these advantages, which in turn increases the variance in the local ratio updates.
This exposes two method-specific failure modes: MAPPO's additive clipping removes gradients for outlier samples and weakens recovery from policy drift, while MASPO's soft quadratic penalty can allow probability collapse. We introduce Multi-Agent Ratio Symmetry (MARS), a novel policy optimization objective that replaces these additive ratio-based trust-region mechanisms with a multiplicatively symmetric geometric barrier.
MARS preserves corrective gradients while assigning unbounded cost as probability ratios approach zero.
Across 47 tasks spanning eight multi-agent environments, including novel JAX benchmarks PaxMen and AeroJAX, MARS matches or exceeds MAPPO and MASPO in aggregate environment-level performance.
Ablations show that these gains arise from the geometry of the symmetric barrier rather than from flexible trust-region boundaries alone.
\end{abstract}

\section{Introduction}
\label{section:introduction}

Deep Multi-Agent Reinforcement Learning (MARL) has enabled agents to learn coordinated behavior in domains ranging from games \citep{ellis2023smacv2} to swarm robotics \citep{liu2024language} and transportation infrastructure \citep{saifullah2024multi}. A common framework for cooperative MARL is centralized training with decentralized execution (CTDE), where actors are trained using centralized information but execute from local observations. Within CTDE, ratio-based trust-region policy-gradient methods such as Multi-Agent Proximal Policy Optimization (MAPPO) \citep{yu2022surprising} and Multi-Agent Simple Policy Optimization (MASPO) \citep{selmonaj2025coordinated} are widely used because they are simple and empirically robust.

CTDE creates a distinct challenge for ratio-based policy updates. In MAPPO and MASPO, each decentralized actor is updated using a per-agent probability ratio, while the update direction is weighted by an advantage estimate from a centralized critic. Because this critic evaluates outcomes under simultaneously changing teammate policies, its advantage estimates can have high variance \citep{lyu2023centralized, foerster2018counterfactual}. These high-variance joint advantages are then applied directly to local ratio updates, producing large increases or decreases in individual action probabilities.

Existing additive trust-region mechanisms are brittle in this regime. MAPPO clips per-agent ratios to a fixed interval around the old policy. When a sample crosses the clipping boundary in the improving direction, its policy-gradient contribution is zeroed. In CTDE, this can remove corrective information from precisely the samples most affected by high-variance joint advantages, allowing policy drift to accumulate across minibatch updates.

MASPO addresses this truncation by replacing hard clipping with a soft quadratic penalty, so samples outside the nominal trust region continue to provide gradients. However, because the penalty is defined by additive distance from the old policy, it remains finite as the probability ratio approaches zero, corresponding to probability extinction for the sampled action. A large negative joint advantage can therefore dominate the penalty and continue decreasing the probability of a sampled action, even when that negative signal reflects teammate exploration or a transient coordination failure. This can reduce exploration over actions that remain useful for coordination \citep{agarwal2021theory}.

The shared limitation is that MAPPO and MASPO measure ratio changes additively around one, even though probability ratios represent relative changes in action probability. Additive bounds treat equal numerical changes above and below one as symmetric, but expansion and contraction are naturally symmetric only as multiplicative inverses. In MAPPO, this makes the clipping thresholds geometrically asymmetric: contraction can proceed farther than expansion before truncation. In MASPO, the same additive geometry appears as a quadratic penalty whose cost remains finite as the probability ratio approaches zero. Under high-variance joint advantages, noisy negative signals can suppress useful actions more aggressively than intended, while the objective provides either no corrective gradient after clipping or only finite resistance near probability collapse.

We introduce \textbf{Multi-Agent Ratio Symmetry (MARS)}, a CTDE policy optimization objective that replaces additive ratio control with a multiplicatively symmetric geometric barrier. MARS preserves smooth gradients over the valid ratio domain and assigns unbounded cost as probability ratios approach zero. It therefore retains gradient information from outlier samples while allowing negative-advantage actions to be down-weighted without making ratio extinction optimal.

\paragraph{Contributions.}
We contribute: (i) MARS, a CTDE-focused full-gradient trust-region objective for cooperative MARL based on a multiplicatively symmetric geometric barrier; (ii) theoretical results showing that MARS preserves corrective gradients and imposes an infinite barrier as probability ratios approach zero; (iii) two JAX-native benchmarks, PaxMen for coordinated exploration and AeroJAX for continuous 3D aerial combat; and (iv) an eight-environment, 47-task evaluation showing that MARS matches or exceeds MAPPO and MASPO in aggregate environment-level performance, with ablations attributing the gains to the curvature of the symmetric barrier rather than flexible trust-region boundaries alone.

\begin{figure*}[t]
    \centering
    \includegraphics[width=1.0\linewidth]{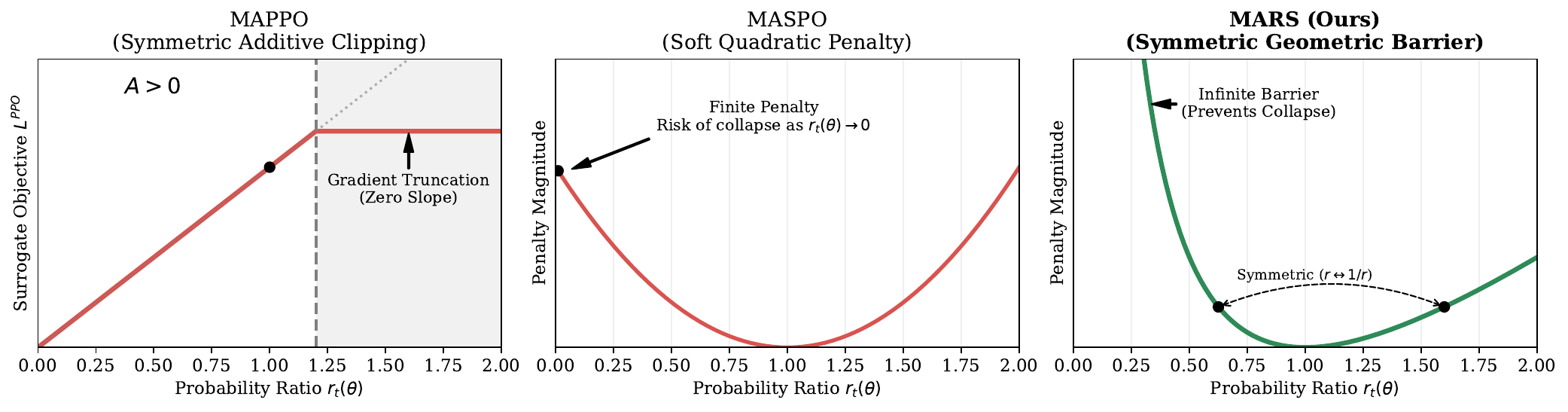}
    \caption{\textit{Comparison of ratio-based trust region mechanisms.} MAPPO's additive clipping creates zero-gradient regions after ratio outliers cross the clipping bounds. MASPO restores gradient flow with a soft quadratic penalty, but its cost remains finite as $r\to 0$. MARS replaces additive ratio control with a multiplicatively symmetric geometric barrier that preserves gradients on $r>0$ and diverges as $r\to 0$, making probability extinction non-optimal for the per-sample ratio surrogate.}
    \label{fig:objective_comparison}
\end{figure*}

\section{Background}
\label{section:background}

\subsection{Problem Formulation}
\label{subsection:problem_formulation}

We model cooperative MARL as a decentralized partially observable Markov decision process (Dec-POMDP) \citep{oliehoek2016concise}
$\mathcal{M}=\langle \mathcal{I},\mathcal{S},\{\mathcal{A}_i\}_{i\in\mathcal{I}},\{\mathcal{O}_i\}_{i\in\mathcal{I}},T,O,R,b_0,\gamma\rangle$.
Here, $\mathcal{I}=\{1,\dots,n\}$ is the set of agents, $\mathcal{S}$ is the state space, and each agent $i$ has action and observation spaces $\mathcal{A}_i$ and $\mathcal{O}_i$.
At timestep $t$, agents take a joint action
$\mathbf{a}_t=(a_{1,t},\dots,a_{n,t})$, the environment transitions according to
$s_{t+1}\sim T(s_t,\mathbf{a}_t)$, and each agent receives a local observation $o_{i,t+1}$.
All agents share the reward $r_t=R(s_t,\mathbf{a}_t)$.

The objective is to learn decentralized policies
$\pi_i(a_{i,t}\mid h_{i,t})$, where $h_{i,t}$ is
agent $i$'s local action-observation history, that maximize
$J=\mathbb{E}\!\left[\sum_{t=0}^{\infty}\gamma^t r_t\right]$.
In centralized training with decentralized execution (CTDE), actors condition only on local histories at execution time, while training may use centralized information to estimate value functions or advantages.

\subsection{Multi-Agent Proximal Policy Optimization}
\label{subsection:mappo}

Multi-Agent Proximal Policy Optimization (MAPPO) \citep{yu2022surprising} adapts the clipped ratio objective of PPO \citep{schulman2017proximal} to CTDE by combining decentralized actors with a centralized critic.
For each agent $i$, MAPPO forms the per-agent probability ratio
\begin{equation}
r_{i,t}(\theta)
=
\frac{\pi_\theta(a_{i,t}\mid h_{i,t})}
{\pi_{\theta_{\mathrm{old}}}(a_{i,t}\mid h_{i,t})},
\end{equation}
and weights the policy update by a joint advantage estimate $\widehat{\mathbf{A}}_t$.
The per-agent MAPPO objective is
\begin{equation}
\mathcal{L}_{i,t}^{\mathrm{MAPPO}}(\theta)
=
\min\Big(
r_{i,t}(\theta)\widehat{\mathbf{A}}_t,\,
\mathrm{clip}(r_{i,t}(\theta),1-\epsilon,1+\epsilon)\widehat{\mathbf{A}}_t
\Big),
\end{equation}
where $\epsilon$ controls the clipping range.

\subsection{Multi-Agent Simple Policy Optimization}
\label{subsection:maspo}

Multi-Agent Simple Policy Optimization (MASPO) \citep{selmonaj2025coordinated} extends the soft-penalty objective of Simple Policy Optimization (SPO) \citep{xie2025simple} to CTDE MARL.
Using the same per-agent ratio $r_{i,t}(\theta)$ and joint advantage estimate $\widehat{\mathbf{A}}_t$, MASPO replaces clipping with an additive quadratic penalty:
\begin{equation}
\mathcal{L}_{i,t}^{\mathrm{MASPO}}(\theta)
=
r_{i,t}(\theta)\widehat{\mathbf{A}}_t
-
\frac{|\widehat{\mathbf{A}}_t|}{2\epsilon}
\left(r_{i,t}(\theta)-1\right)^2.
\end{equation}
The quadratic term softly penalizes deviations of the per-agent ratio from one while retaining gradient flow outside the nominal trust region.

\section{Multi-Agent Ratio Symmetry}
\label{section:multi_agent_ratio_symmetry}

\subsection{Motivation: Joint Advantage Variance Exposes Ratio-Control Failures}
\label{subsection:motivation}

The preceding section defined MAPPO and MASPO as per-agent ratio objectives weighted by a joint advantage estimate. This combination is central to CTDE policy optimization: each decentralized actor is updated through a local probability ratio, while the learning signal reflects the joint behavior of all agents. As teammates adapt, the same local action can receive different advantage estimates across updates because the critic evaluates outcomes under a non-stationary joint policy \citep{lyu2023centralized, foerster2018counterfactual}. Thus, high-variance joint advantage estimates are applied directly to per-agent probability ratios.

\subsubsection{Gradient Truncation in MAPPO}
\label{subsubsection:mappo_gradient_truncation}

In MAPPO, the additive trust region is enforced through hard clipping bounds $[1-\epsilon,1+\epsilon]$, as shown in \cref{fig:objective_comparison}. The clipped objective removes gradients when a sample crosses the clipping boundary in the improving direction. For the per-agent, single-step clipped objective, the ratio-gradient is
\begin{equation}
    \frac{\partial \mathcal{L}_{i,t}^{\mathrm{MAPPO}}}{\partial r_{i,t}}
    =
    \begin{cases}
      \widehat{\mathbf{A}}_t,
      & \text{if } |r_{i,t}(\theta)-1| < \epsilon
      \text{ or }
      \mathrm{sign}(\widehat{\mathbf{A}}_t)(r_{i,t}(\theta)-1) < 0, \\
      0, & \text{otherwise}.
   \end{cases}
\end{equation}
See Appendix~\ref{appendix:subsection:ppo_clipped_objective_gradient} for the derivation.

In CTDE, this truncation is consequential because ratio outliers are often produced by high-variance joint advantage estimates. Once a transition is clipped, its objective term no longer provides a gradient that pulls the ratio back toward the trust region. Since policy parameters are shared across transitions in a minibatch, updates from other active samples can still move the clipped ratio further from the old policy. Without a smooth restoring signal for clipped outliers to counteract this drift, MAPPO allows the policy to drift arbitrarily far from the constraint, undermining the monotonic improvement guarantee \citep{wang2020truly, xie2025simple, engstrom2019implementation}.

\subsubsection{Finite Penalty in MASPO}
\label{subsubsection:maspo_finite_penalty}

MASPO removes MAPPO's hard clipping and restores gradient flow by replacing the clipped objective with a soft quadratic penalty,
\begin{equation}
    \psi_{\mathrm{MASPO}}(r) = (r-1)^2.
\end{equation}
This penalty is defined using additive distance from one. As a result, it assigns finite cost to probability extinction:
\begin{equation}
    \lim_{r\to 0^+}(r-1)^2 = 1.
\end{equation}
Thus, MASPO maintains gradients outside the nominal trust region, but its penalty remains bounded as the probability ratio approaches zero.

This bounded cost matters under joint advantage noise. A large negative joint advantage can dominate the finite penalty and continue decreasing the probability of the sampled action. Because the joint advantage may reflect teammate exploration or temporary coordination failure, this update can prematurely suppress actions that remain useful for future coordination, contributing to near-deterministic policies \citep{agarwal2021theory}.

\subsubsection{Additive Ratio Geometry as a Common Structure}
\label{subsubsection:additive_geometry}

MAPPO and MASPO fail through different mechanisms, but both regulate probability-ratio change using additive distance from $r=1$. In MAPPO, the additive interval $[1-\epsilon,1+\epsilon]$ determines where clipping occurs. This interval is symmetric as a numerical distance from one, but it is not symmetric as a relative change in probability. For example, with $\epsilon=0.2$, the lower bound $0.8$ corresponds to a $20\%$ contraction, whose inverse expansion is $1/0.8=1.25$, while the upper bound permits expansion only to $1.2$. Thus, before clipping occurs, MAPPO permits a stronger multiplicative contraction than expansion. Once the lower boundary is crossed for a negative-advantage sample, the clipped objective removes the gradient, so the same additive boundary both permits geometrically uneven contraction and then removes the corrective signal after the sample becomes an outlier.

MASPO uses the same additive parameterization in soft form: its quadratic penalty depends on distance from $r=1$. This restores gradient flow outside the nominal trust region, but the penalty remains finite as $r\to0$. As a result, a sufficiently large negative joint advantage can continue pushing the sampled-action ratio toward zero. Thus, MAPPO's truncation and MASPO's collapse are distinct failures, but both arise from controlling relative probability changes through additive distance around $r=1$.

The geometric issue is clearer in log-ratio space. Probability ratios encode relative change: expansion by $r$ is naturally inverted by contraction by $1/r$. These inverse changes are symmetric in $u=\log r$, since $r$ and $1/r$ become $u$ and $-u$. Additive neighborhoods around one do not preserve this symmetry, while the collapse limit $r\to0$ corresponds to $u\to-\infty$. This motivates replacing additive ratio control with a regularizer defined by multiplicative symmetry rather than numerical distance from one.

\subsection{The Symmetric Ratio Objective}
\label{subsection:symmetric_ratio_objective}

The preceding analysis suggests three criteria for a CTDE-compatible ratio regularizer:
\begin{enumerate}
    \item \textbf{Smooth corrective gradients:} The objective should remain differentiable across the valid ratio domain so that outlier samples continue to provide gradient information rather than being removed by clipping.
    \item \textbf{Unbounded extinction barrier:} The penalty should diverge as $r\to 0^+$, ensuring that no finite negative joint advantage can make probability extinction optimal.
    \item \textbf{Multiplicative symmetry:} The penalty should satisfy $\psi(r)=\psi(1/r)$, treating expansion and contraction as inverse changes in probability rather than as unequal additive deviations from one.
\end{enumerate}

\subsubsection{The Multi-Agent Ratio Symmetry Penalty}
\label{subsubsection:mars_penalty}

Many ratio penalties could satisfy these criteria. We choose a geometric symmetrization of MASPO's additive quadratic penalty because it aligns with the multiplicative geometry of probability ratios while keeping the construction directly comparable to MASPO. In particular, the geometric mean treats a ratio and its inverse as paired deviations, preserves local quadratic behavior near $r=1$, and changes the penalty primarily by replacing additive distance with multiplicative symmetry.

Concretely, we construct the Multi-Agent Ratio Symmetry (MARS) penalty by symmetrizing the MASPO quadratic penalty under the transformation $r \mapsto 1/r$. Applying geometric symmetrization to $(r-1)^2$ yields
\begin{equation}
    \psi_{\mathrm{MARS}}(r)
    =
    r+\frac{1}{r}-2.
    \label{eq:mars_penalty}
\end{equation}
See Appendix~\ref{appendix:subsection:symmetric_objective_derivation} for the derivation. This penalty is zero at $r=1$, smooth for all $r>0$, and symmetric under inversion by design. 
The linear term penalizes unbounded expansion as $r\to\infty$, while the reciprocal term creates an unbounded barrier as $r\to 0^+$.

The derivative of the penalty with respect to the ratio is
\begin{equation}
    \frac{d}{dr}\psi_{\mathrm{MARS}}(r)
    =
    1-\frac{1}{r^2}.
\end{equation}
As $r\to 0^+$, the reciprocal-gradient term dominates. Therefore, when a large negative joint advantage pushes a sampled-action ratio downward, the MARS penalty contributes an increasingly strong opposing term in ratio space.

Now consider the per-agent, one-step ratio surrogate
\begin{equation}
    \mathcal{L}^{\mathrm{MARS}}_{i,t}(\theta)
    =
    r_{i,t}(\theta)\widehat{\mathbf{A}}_t
    -
    \alpha_t
    \psi_{\mathrm{MARS}}\!\big(r_{i,t}(\theta)\big),
    \label{eq:mars_base_objective}
\end{equation}
where $\alpha_t>0$ controls the penalty strength. The ratio-gradient is
\begin{equation}
    \frac{\partial \mathcal{L}^{\mathrm{MARS}}_{i,t}}{\partial r_{i,t}}
    =
    \widehat{\mathbf{A}}_t
    -
    \alpha_t
    \left(
    1-\frac{1}{r_{i,t}(\theta)^2}
    \right).
    \label{eq:mars_ratio_gradient}
\end{equation}
Equation~\ref{eq:mars_ratio_gradient} shows the key barrier mechanism. For negative joint advantages, the linear surrogate can favor reducing the sampled-action ratio, but as $r\to 0^+$ the reciprocal term in the MARS penalty dominates any finite advantage. Thus, MARS can reduce ratios for negative-advantage samples while preventing ratio extinction from becoming optimal. Proposition~\ref{prop:barrier} formalizes this limit behavior.

\begin{proposition}[Barrier Against Probability Extinction]
\label{prop:barrier}
Let $r>0$ and consider the per-agent, one-step MARS surrogate
\[
    \mathcal{L}(r)
    =
    r\widehat{\mathbf{A}}
    -
    \alpha\psi_{\mathrm{MARS}}(r),
\]
with $\alpha>0$ and $\psi_{\mathrm{MARS}}(r)=r+\frac{1}{r}-2$. Then
\[
    \lim_{r\to 0^+}\mathcal{L}(r)=-\infty,
    \qquad
    \lim_{r\to 0^+}
    \frac{\partial \mathcal{L}}{\partial r}
    =
    +\infty.
\]
Consequently, probability extinction is never an optimizer of the per-sample ratio surrogate, and the ratio-gradient points away from zero in a neighborhood of $r=0$. Full proof in Appendix~\ref{appendix:subsection:prop_barrier}.
\end{proposition}

Having established the barrier behavior near $r=0$, we now consider the second desired property: preserving gradient information for ratio outliers rather than creating clipped zero-gradient regions.

\begin{proposition}[No Gradient Truncation]
\label{prop:no_gradient_truncation}
Let $r>0$, $\alpha>0$, and $\widehat{\mathbf{A}}\in\mathbb{R}$ be finite. For the MARS surrogate
\[
    \mathcal{L}(r)
    =
    r\widehat{\mathbf{A}}
    -
    \alpha\psi_{\mathrm{MARS}}(r),
    \qquad
    \psi_{\mathrm{MARS}}(r)=r+\frac{1}{r}-2,
\]
the ratio-gradient
\[
    \frac{\partial \mathcal{L}}{\partial r}
    =
    \widehat{\mathbf{A}}
    -
    \alpha
    \left(
    1-\frac{1}{r^2}
    \right)
\]
has no zero-gradient interval on $(0,\infty)$.
Full proof in Appendix~\ref{appendix:subsection:no_gradient_truncation}.
\end{proposition}
Proposition~\ref{prop:no_gradient_truncation} shows that the MARS trust-region mechanism does not create MAPPO-style gradient truncation. Gradients may vanish at isolated stationary points, but there is no interval of ratios for which outlier samples stop contributing gradient information solely because they crossed a clipping boundary.

Together, Propositions~\ref{prop:barrier} and~\ref{prop:no_gradient_truncation} show that MARS assigns unbounded cost as probability ratios approach zero while avoiding gradient truncation.

\subsubsection{Trust Region Alignment}
\label{subsection:trust_region_alignment}

The MARS penalty defines the shape of the ratio regularizer, while the scalar penalty weight $\alpha_t$ determines how strongly that penalty is applied. We choose $\alpha_t$ so that the advantage term and penalty term balance at the prescribed target ratio, making that ratio the stationary point of the one-step surrogate. This target plays the same role as the clipping boundary in MAPPO or the $\epsilon$-aligned optimum in MASPO, but is specified directly in multiplicative ratio space rather than through additive bounds around one. This distinction matters because additive bounds of the form $(1\pm\epsilon)$ are not multiplicatively symmetric: the lower and upper bounds are not reciprocal. We therefore define separate target ratios for expansion and contraction:
\begin{itemize}
    \item \textbf{Expansion target ($B_{\mathrm{upper}}>1$):} the intended boundary for up-weighting actions with $\widehat{\mathbf{A}}_t\ge 0$.
    \item \textbf{Contraction target ($B_{\mathrm{lower}}<1$):} the intended boundary for down-weighting actions with $\widehat{\mathbf{A}}_t<0$.
\end{itemize}

\paragraph{Target-ratio alignment for $\alpha_t$.}
Let $r_{\mathrm{target}}=B_{\mathrm{upper}}$ for $\widehat{\mathbf{A}}_t\ge0$ and $r_{\mathrm{target}}=B_{\mathrm{lower}}$ otherwise.
We set $\alpha_t$ by requiring the ratio-gradient of the one-step surrogate to vanish at $r_{\mathrm{target}}$. Since
\[
\frac{\partial \mathcal{L}^{\mathrm{MARS}}_{i,t}}{\partial r_{i,t}}
=
\widehat{\mathbf{A}}_t
-
\alpha_t
\left(
1-\frac{1}{r_{i,t}^2}
\right),
\qquad
\left.
\frac{\partial \mathcal{L}^{\mathrm{MARS}}_{i,t}}{\partial r_{i,t}}
\right|_{r_{i,t}=r_{\mathrm{target}}}
=0,
\]
we obtain
\begin{equation}
    \alpha_t
    =
    \frac{\widehat{\mathbf{A}}_t}
    {1-r_{\mathrm{target}}^{-2}}
    =
    \begin{cases}
        \frac{\widehat{\mathbf{A}}_t}{1-B_{\mathrm{upper}}^{-2}}
        & \text{if } \widehat{\mathbf{A}}_t \ge 0,\\[6pt]
        \frac{\widehat{\mathbf{A}}_t}{1-B_{\mathrm{lower}}^{-2}}
        & \text{if } \widehat{\mathbf{A}}_t < 0.
    \end{cases}
    \label{eq:mars_target_scaling}
\end{equation}
When $B_{\mathrm{upper}}>1$ and $B_{\mathrm{lower}}<1$, this yields $\alpha_t\ge 0$ in both regimes. The standard $\epsilon$-aligned MASPO scaling is recovered as the additive special case $r_{\mathrm{target}}=1+\mathrm{sign}(\widehat{\mathbf{A}}_t)\epsilon$; see Appendix~\ref{appendix:subsection:trust_region_boundaries}.


\paragraph{Final objective.}
We now have the two ingredients needed to instantiate the MARS surrogate: a multiplicatively symmetric penalty shape and a penalty weight $\alpha_t$ calibrated to the desired expansion or contraction target. Combining the penalty in \cref{eq:mars_penalty} and the aligned weight in \cref{eq:mars_target_scaling} with the base ratio surrogate in \cref{eq:mars_base_objective} gives the final per-agent MARS objective:
\begin{equation}
\begin{split}
    \mathcal{L}^{\mathrm{MARS}}_{i,t}(\theta)
    &=
    r_{i,t}(\theta)\widehat{\mathbf{A}}_t
    -
    \alpha_t
    \left(
    r_{i,t}(\theta)
    +
    \frac{1}{r_{i,t}(\theta)}
    -
    2
    \right).
\end{split}
\label{eq:mars_final_objective}
\end{equation}
Thus, MARS keeps the standard ratio-weighted policy-gradient term but replaces additive clipping or additive quadratic penalties with a calibrated multiplicatively symmetric barrier. The full actor objective averages this surrogate over agents and timesteps in the same CTDE training loop as MAPPO and MASPO.



\section{Results}
\label{section:results}
\subsection{Experimental Setup}
\label{subsection:experimental_setup}

We use the JAX-based Mava library \citep{de2021mava} to compare MARS against standard MAPPO and MASPO. To isolate whether MARS gains arise from the symmetric barrier itself rather than from allowing separate expansion and contraction limits, we include four boundary-flexibility ablations. The asymmetric MAPPO ablation replaces the standard clipping interval $[1-\epsilon,1+\epsilon]$ with $[1-\epsilon_{\mathrm{lower}},1+\epsilon_{\mathrm{upper}}]$. The asymmetric MASPO ablation keeps the MASPO quadratic penalty but scales it using a sign-dependent target, $r_{\mathrm{target}}=1+\epsilon_{\mathrm{upper}}$ for $\widehat{\mathbf{A}}_t\ge0$ and $r_{\mathrm{target}}=1-\epsilon_{\mathrm{lower}}$ otherwise. The MARS ablations keep the MARS barrier fixed but constrain the target ratios: multiplicatively symmetric MARS uses $B_{\mathrm{lower}}=1/B_{\mathrm{upper}}$, while additively symmetric MARS uses $B_{\mathrm{upper}}=1+\epsilon$ and $B_{\mathrm{lower}}=1-\epsilon$. Exact objectives and search spaces are given in Appendix~\ref{appendix:subsubsection:algorithm_specific_hyperparameters}.

All comparisons therefore keep the simultaneous per-agent actor-update structure fixed and vary only the ratio objective or its trust-region targets. Sequential-update methods such as HATRPO and HAPPO \citep{kuba2022trust} modify the agent update order and factorize the multi-agent policy improvement step, making them complementary to the ratio-geometry question studied here rather than direct controls for this objective-level comparison. 

All methods share the same common hyperparameter search space, with trust-region parameters tuned separately using Optuna's Tree-structured Parzen Estimator \citep{akiba2019optuna}. Methods with one tunable trust-region parameter receive 40 trials per task, while methods with separate upper and lower ratio bounds receive 80 trials per task to account for the larger search space. Agents share policy parameters and observe their agent ID; full hyperparameter ranges and training lengths are given in Appendix~\ref{appendix:section:hyperparameters}.

We evaluate across 47 cooperative MARL tasks spanning eight environments. Six are drawn from established JAX-based benchmark suites: SMAX, RWARE, Connector, MPE, JaxNav, and Search and Rescue \citep{bonnet2024jumanji, rutherford2024jaxmarl}. We additionally introduce two JAX-native benchmarks designed to stress different multi-agent failure modes: \textbf{PaxMen}, a coordinated exploration benchmark with variable team sizes and stochastic maze layouts, and \textbf{AeroJAX}, a continuous-control aerial combat benchmark with attitude-dependent sensing and nonlinear 3D flight dynamics. Appendix~\ref{appendix:section:environments} details each environment and how PaxMen and AeroJAX differ from the prior benchmarks on which they build.

\subsection{Multi-Agent Benchmark Performance}
\label{subsection:multi_agent_benchmark_performance}
\begin{figure*}[t] 
    \centering
    
    \begin{subfigure}{1.0\linewidth} 
        \centering
        \includegraphics[width=0.4\linewidth]{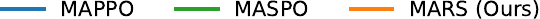}
    \end{subfigure}
    
    \begin{subfigure}{0.24\linewidth}
        \centering
        \includegraphics[width=\linewidth]{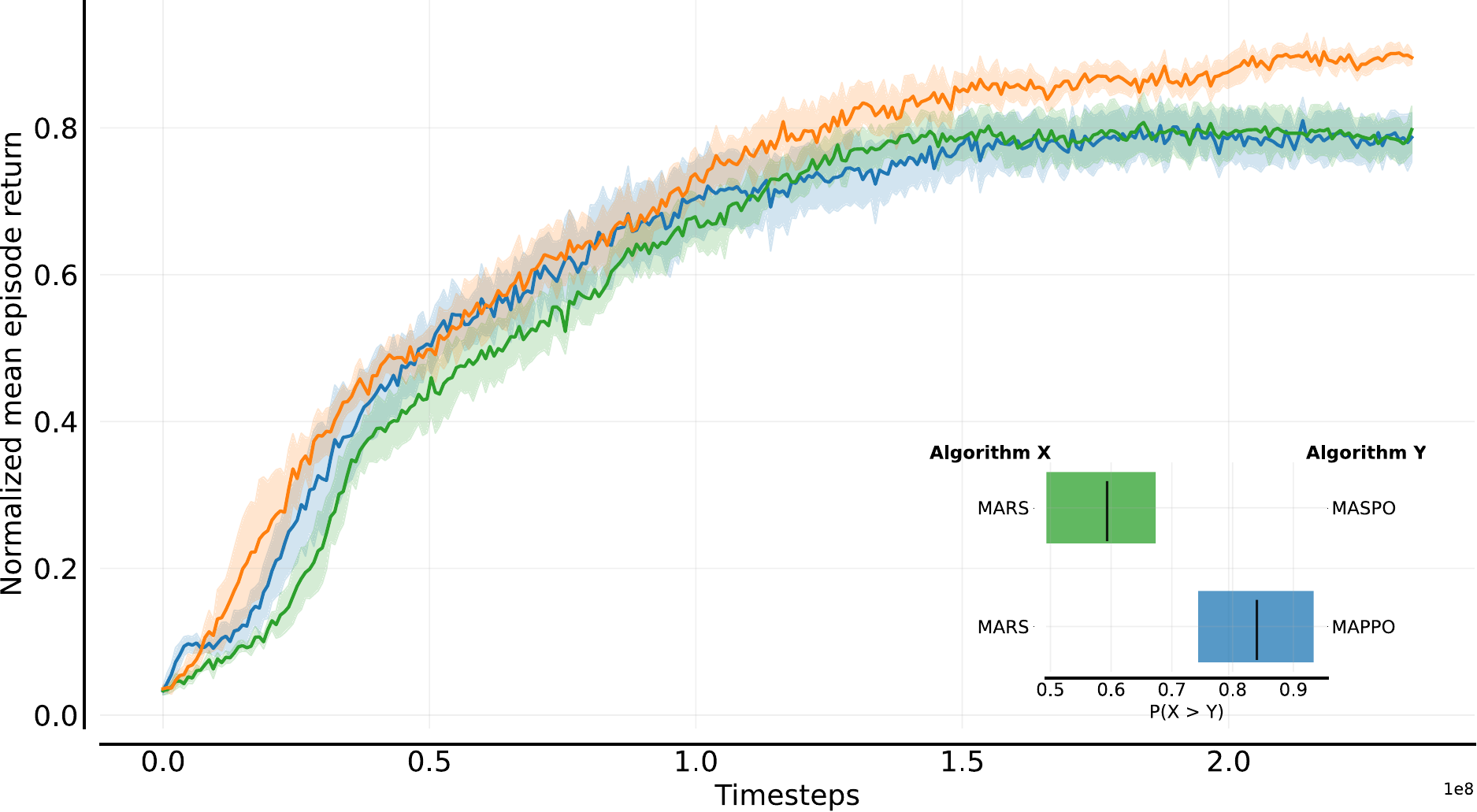}
        \caption{AeroJAX}
        \label{figure:agg_results:aerojax}
    \end{subfigure}
    \hfill 
    \begin{subfigure}{0.24\linewidth}
        \centering
        \includegraphics[width=\linewidth]{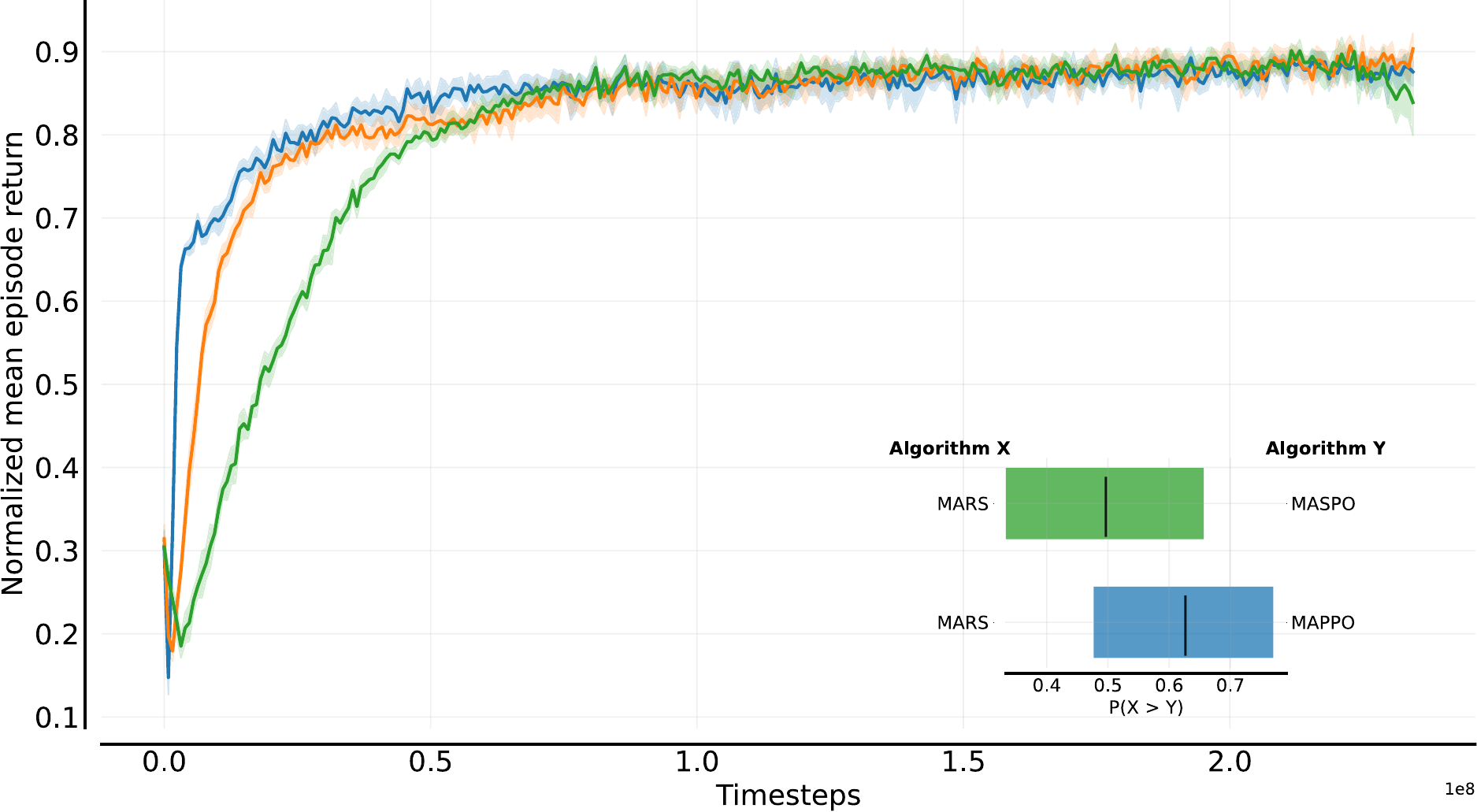}
        \caption{JaxNav}
        \label{figure:agg_results:jaxnav}
    \end{subfigure}
    \hfill
    \begin{subfigure}{0.24\linewidth}
        \centering
        \includegraphics[width=\linewidth]{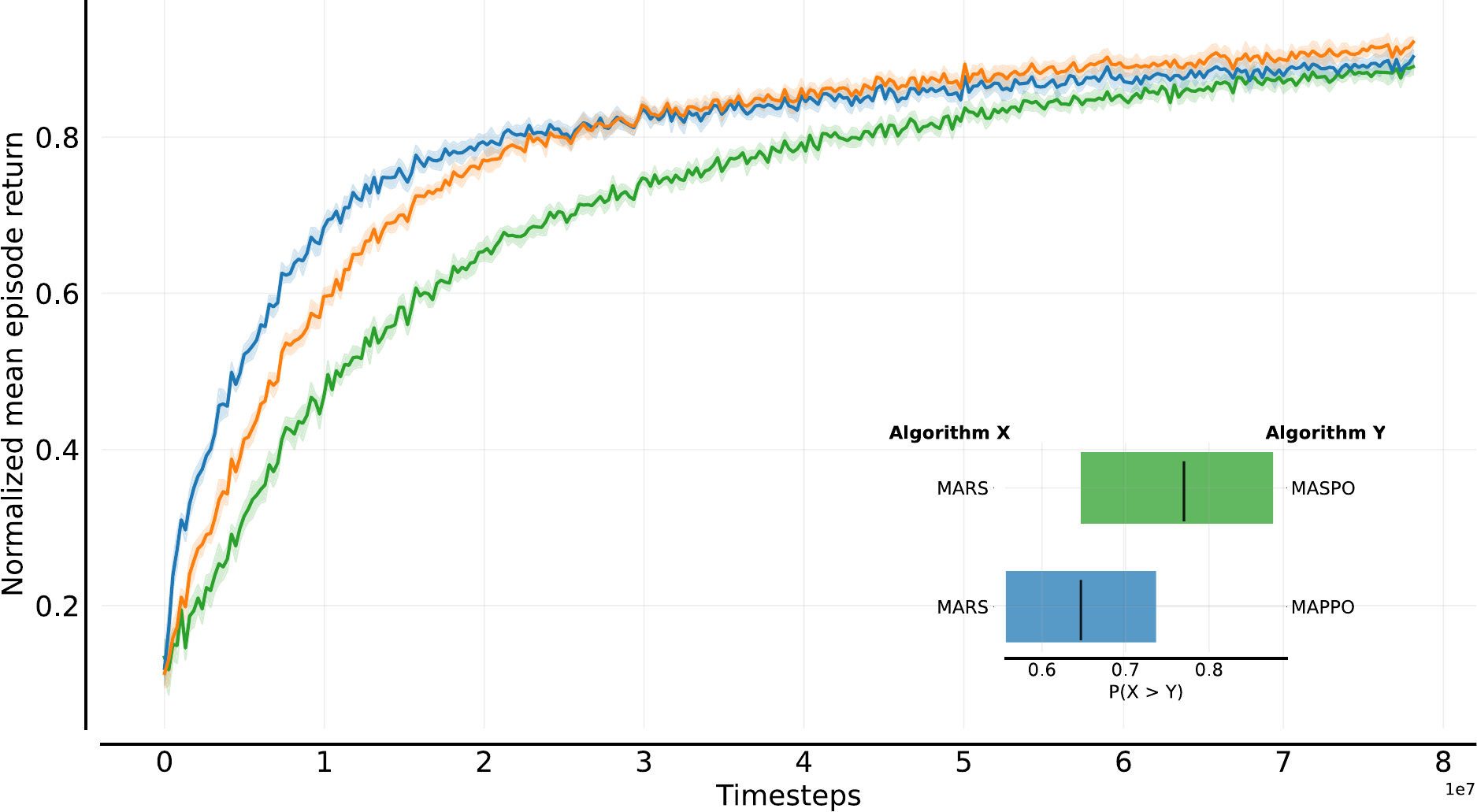}
        \caption{MPE}
        \label{figure:agg_results:mpe}
    \end{subfigure}
    \hfill
    \begin{subfigure}{0.24\linewidth}
        \centering
        \includegraphics[width=\linewidth]{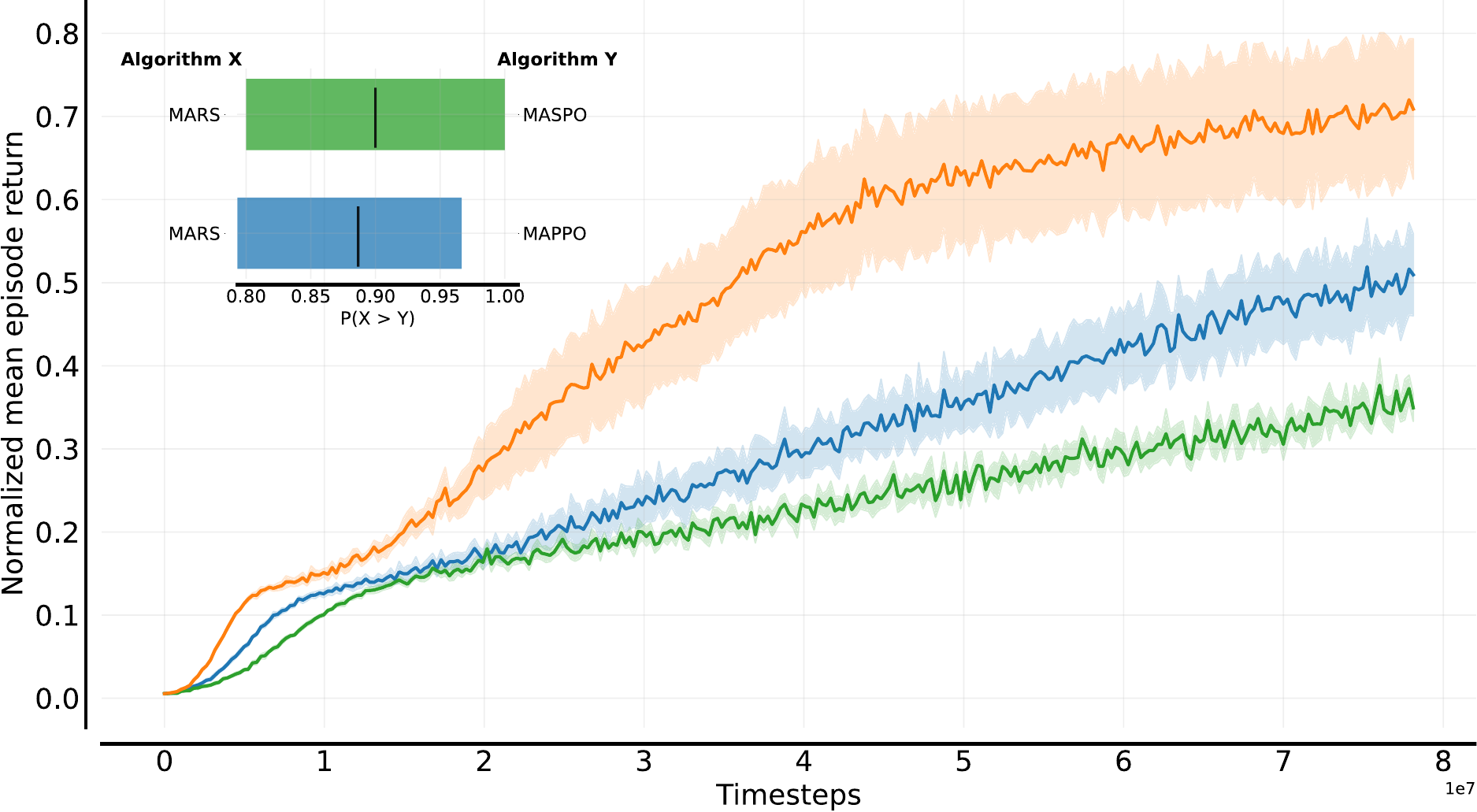}
        \caption{PaxMen}
        \label{figure:agg_results:paxmen}
    \end{subfigure}
    
    
    \hfill
    \begin{subfigure}{0.24\linewidth}
        \centering
        \includegraphics[width=\linewidth]{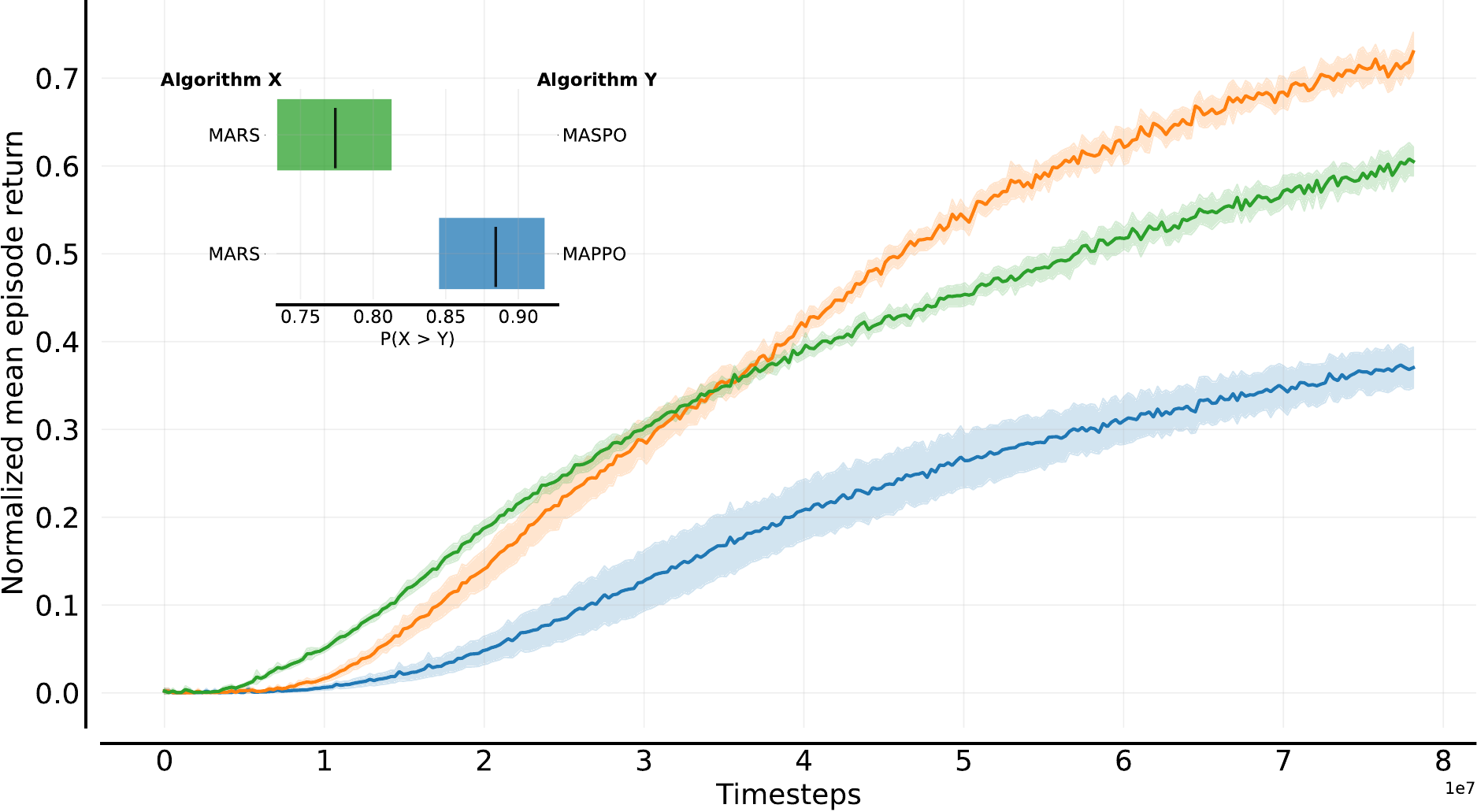}
        \caption{RWARE}
        \label{figure:agg_results:rware}
    \end{subfigure}
    \hfill
    \begin{subfigure}{0.24\linewidth}
        \centering
        \includegraphics[width=\linewidth]{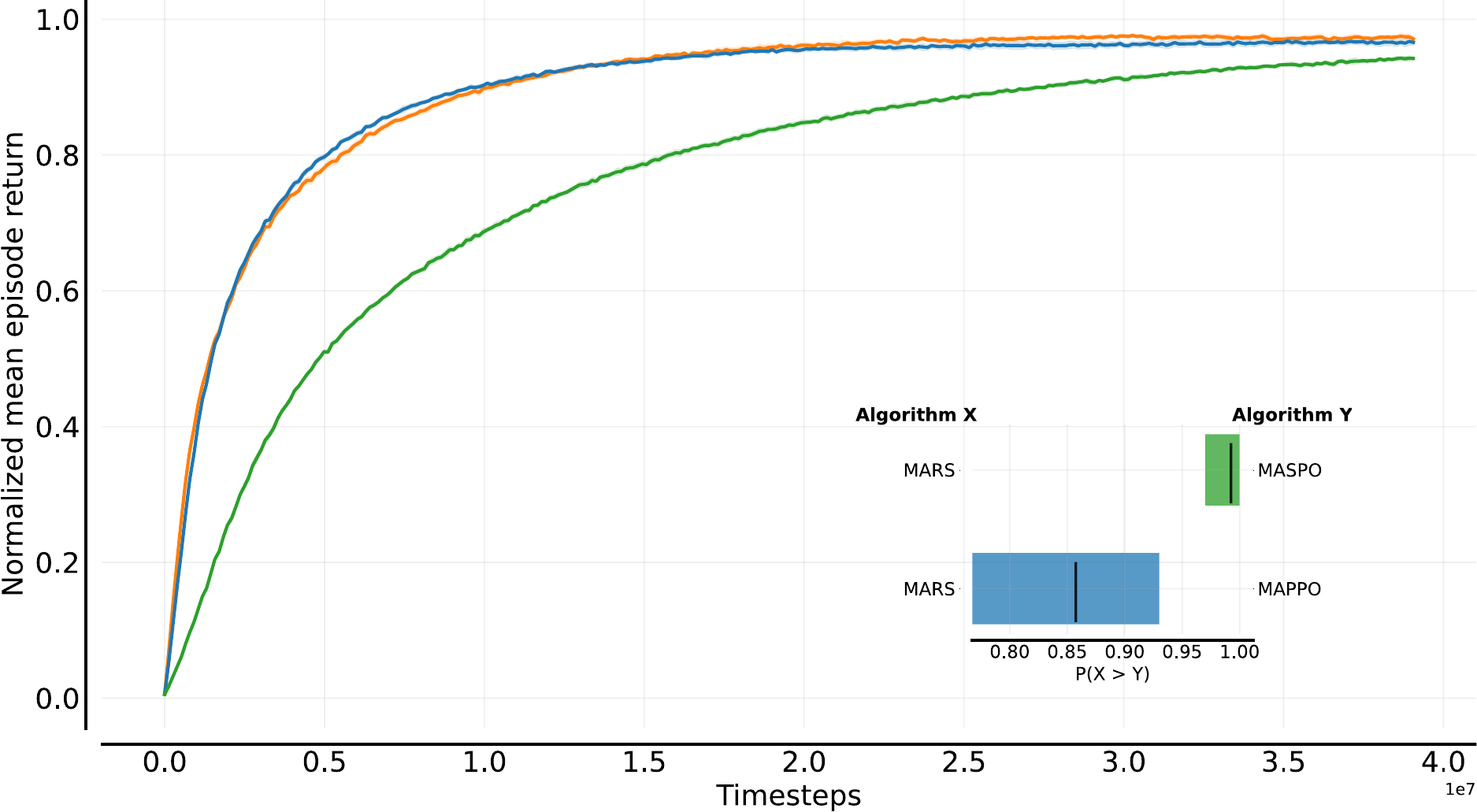}
        \caption{Search and Rescue}
        \label{figure:agg_results:s&r}
    \end{subfigure}
    \hfill
    \begin{subfigure}{0.24\linewidth}
        \centering
        \includegraphics[width=\linewidth]{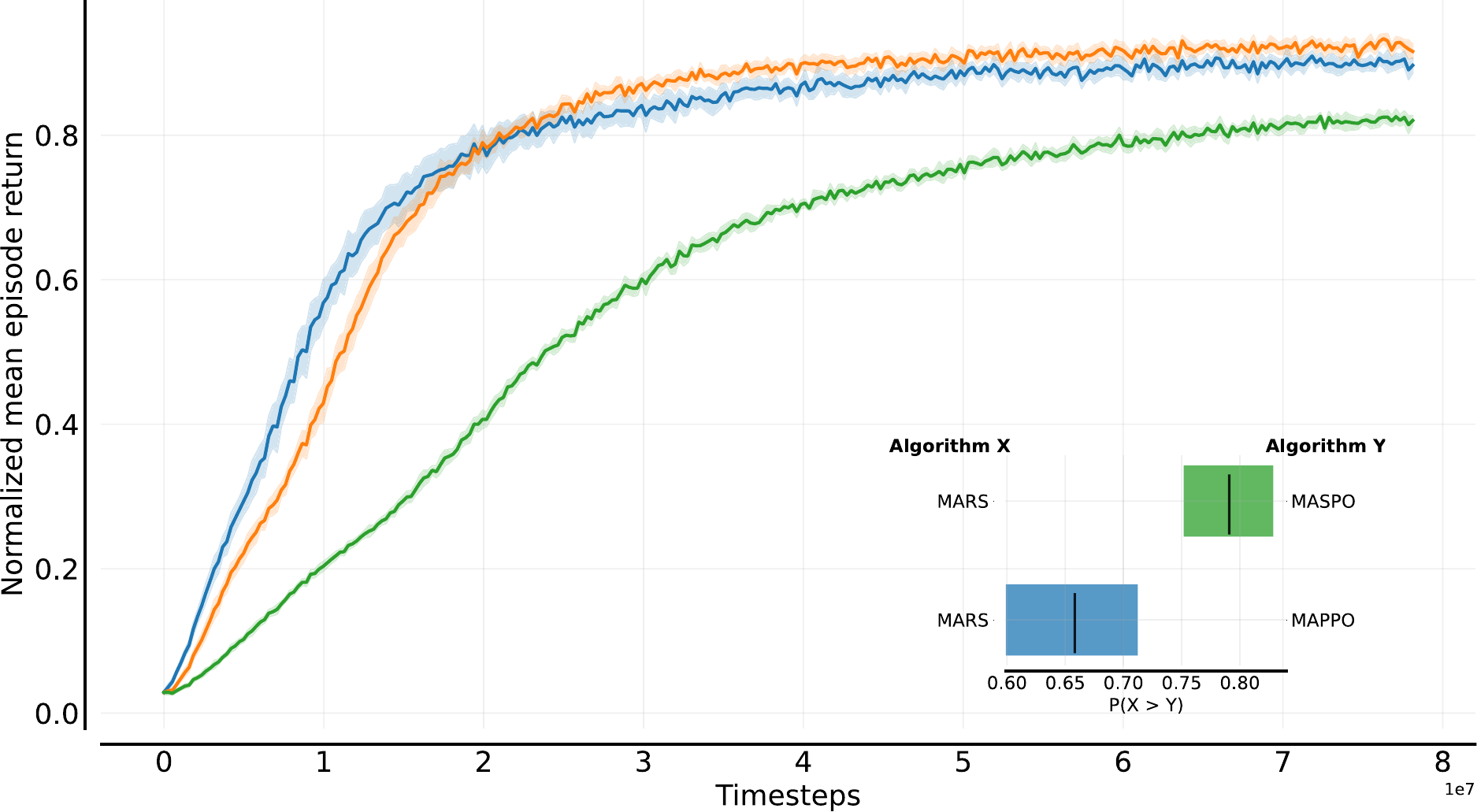}
        \caption{SMAX}
        \label{figure:agg_results:smax}
    \end{subfigure}
    \hfill
    \begin{subfigure}{0.24\linewidth}
        \centering
        \includegraphics[width=\linewidth]{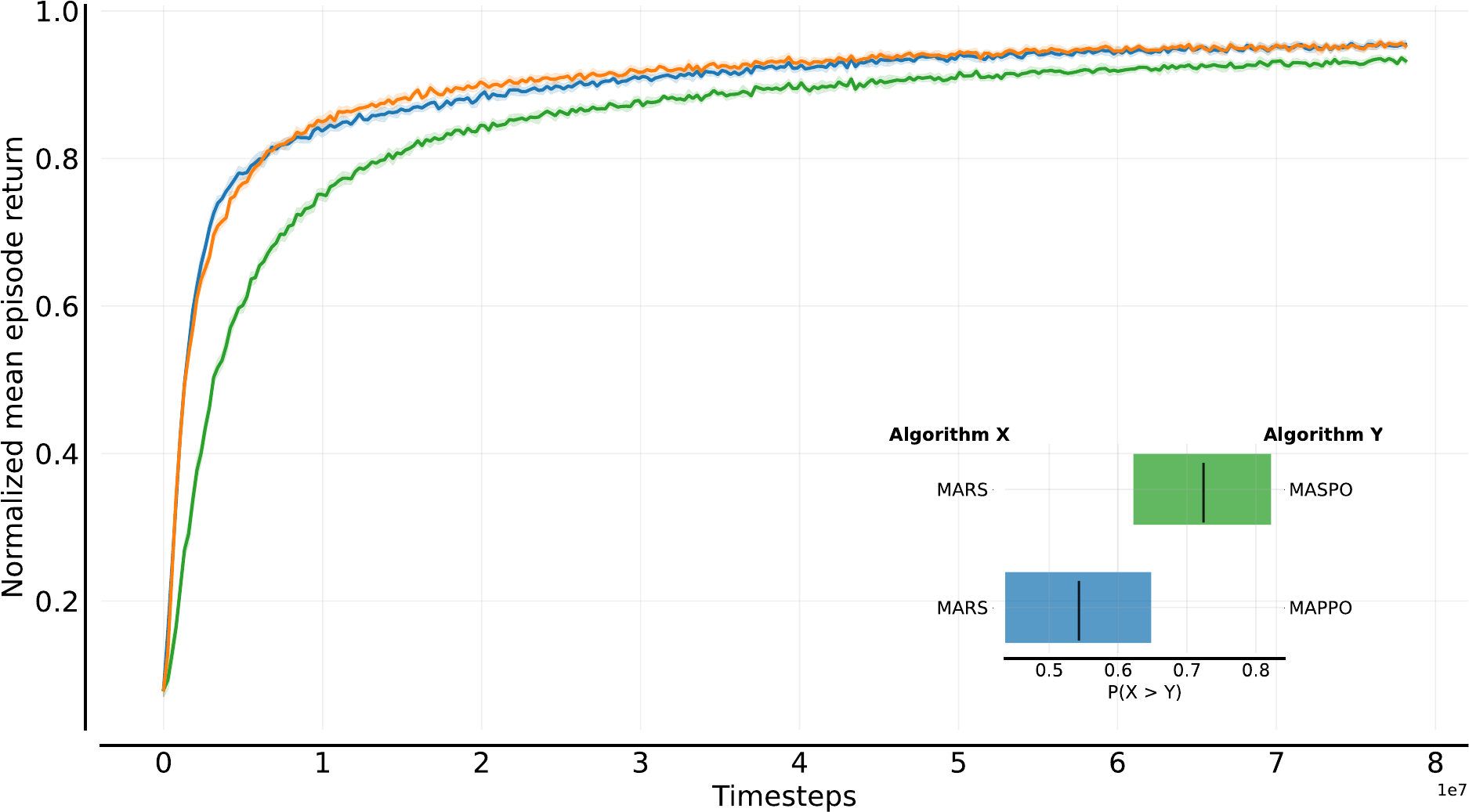}
        \caption{Connector}
        \label{figure:agg_results:connector}
    \end{subfigure}
    \caption{\textit{Aggregated learning performance and probability of improvement.} Results are aggregated across all tasks within each environment using per-task min-max normalization. Main plots depict the mean across 10 independent random seeds (shaded regions denote 95\% confidence intervals). Inset plots show the aggregate probability that MARS improves upon baselines; if the probability of improvement is higher than 0.5 and the confidence intervals do not contain 0.5, then the results are statistically significant \citep{agarwal2021deep}.}
    \label{figure:agg_results}
\end{figure*}

Across 47 tasks in eight cooperative MARL environments, MARS matches or exceeds MAPPO and MASPO in aggregate performance, as shown in \cref{tab:results}. \cref{figure:agg_results} summarizes environment-level learning curves and probability-of-improvement estimates, while Appendix ~\ref{appendix:section:per_task_performance} provides per-task learning curves for all 47 tasks. The probability-of-improvement insets in \cref{figure:agg_results} indicate statistically significant improvements over both baselines in six of the eight environments. These gains are most pronounced in two regimes: high-variance continuous control and discrete coordination tasks requiring sustained exploration.

\begin{table}[t]
    \centering
    \small
    \caption{\textit{Aggregate performance by environment.} We report the interquartile mean (IQM) of per-task min-max normalized absolute episode returns \citep{gorsane2022towards}, bounded by 95\% stratified bootstrap confidence intervals. Scores are aggregated across all tasks within each environment using per-task min-max normalization, then IQM is computed over 10 independent seeds. Bold entries denote the top score for each environment. An asterisk (*) marks results where the confidence interval overlaps with that of the highest score.}
    \label{tab:results}
    \begin{tabular*}{\columnwidth}{l@{\extracolsep{\fill}}ccc}
        \toprule
        Environment & MAPPO & MASPO & MARS \\
        \midrule
        AeroJAX & $0.82_{(0.79,0.85)}$ & $0.82_{(0.79,0.84)}$ & $\mathbf{0.91}_{(0.90,0.91)}$ \\
        Connector & $\mathbf{0.95}^*_{(0.94,0.95)}$ & $0.93_{(0.93,0.93)}$ & $\mathbf{0.95}^*_{(0.95,0.95)}$ \\
        JaxNav & $0.88_{(0.87,0.89)}$ & $\mathbf{0.89}^*_{(0.89,0.89)}$ & $\mathbf{0.89}^*_{(0.88,0.90)}$ \\
        MPE & $0.89_{(0.88,0.89)}$ & $0.88_{(0.88,0.89)}$ & $\mathbf{0.91}_{(0.91,0.92)}$ \\
        PaxMen & $0.49_{(0.43,0.54)}$ & $0.34_{(0.32,0.36)}$ & $\mathbf{0.72}_{(0.59,0.79)}$ \\
        RWARE & $0.36_{(0.34,0.40)}$ & $0.60_{(0.58,0.61)}$ & $\mathbf{0.71}_{(0.69,0.72)}$ \\
        S \& R & $0.97_{(0.97,0.97)}$ & $0.94_{(0.94,0.94)}$ & $\mathbf{0.98}_{(0.98,0.98)}$ \\
        SMAX & $0.89_{(0.89,0.90)}$ & $0.82_{(0.81,0.82)}$ & $\mathbf{0.92}_{(0.92,0.92)}$ \\
        \bottomrule
    \end{tabular*}
\end{table}

\paragraph{Stability in continuous control.}

In continuous-control environments, MARS shows improved training stability, particularly in AeroJAX and JaxNav, where policy updates interact with high-variance dynamics and dense ratio changes. In \cref{figure:agg_results}a--b, MAPPO and MASPO exhibit late-training degradation or plateauing, while MARS maintains higher returns through the end of training. Similar but smaller gains appear in MPE and Search and Rescue. These results are consistent with the mechanism proposed in \cref{section:multi_agent_ratio_symmetry}: preserving ratio-space gradients while imposing a barrier near probability extinction can reduce instability from high-variance joint advantages.

\paragraph{Coordination in discrete control.}

In discrete coordination tasks, MARS performs especially well in sparser-reward environments such as PaxMen and RWARE, where agents must discover coordinated action sequences before receiving useful feedback. Because the MARS penalty assigns unbounded cost near probability extinction, it discourages premature removal of sampled actions from the policy support. This is consistent with the stronger performance observed in \cref{figure:agg_results}d--e. In denser tasks such as SMAX and Connector, MARS also matches or improves over the baselines, suggesting that the same ratio-control mechanism remains useful even when rewards provide more frequent feedback.

\subsection{Ablating Boundary Flexibility and Barrier Geometry}
\label{subsection:ablation_analysis}

\begin{figure*}[t]
    \centering
    \includegraphics[width=0.78\textwidth]{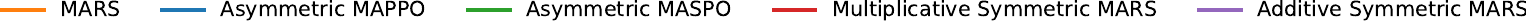}

    \begin{subfigure}[t]{0.315\textwidth}
        \centering
        \includegraphics[width=\linewidth]{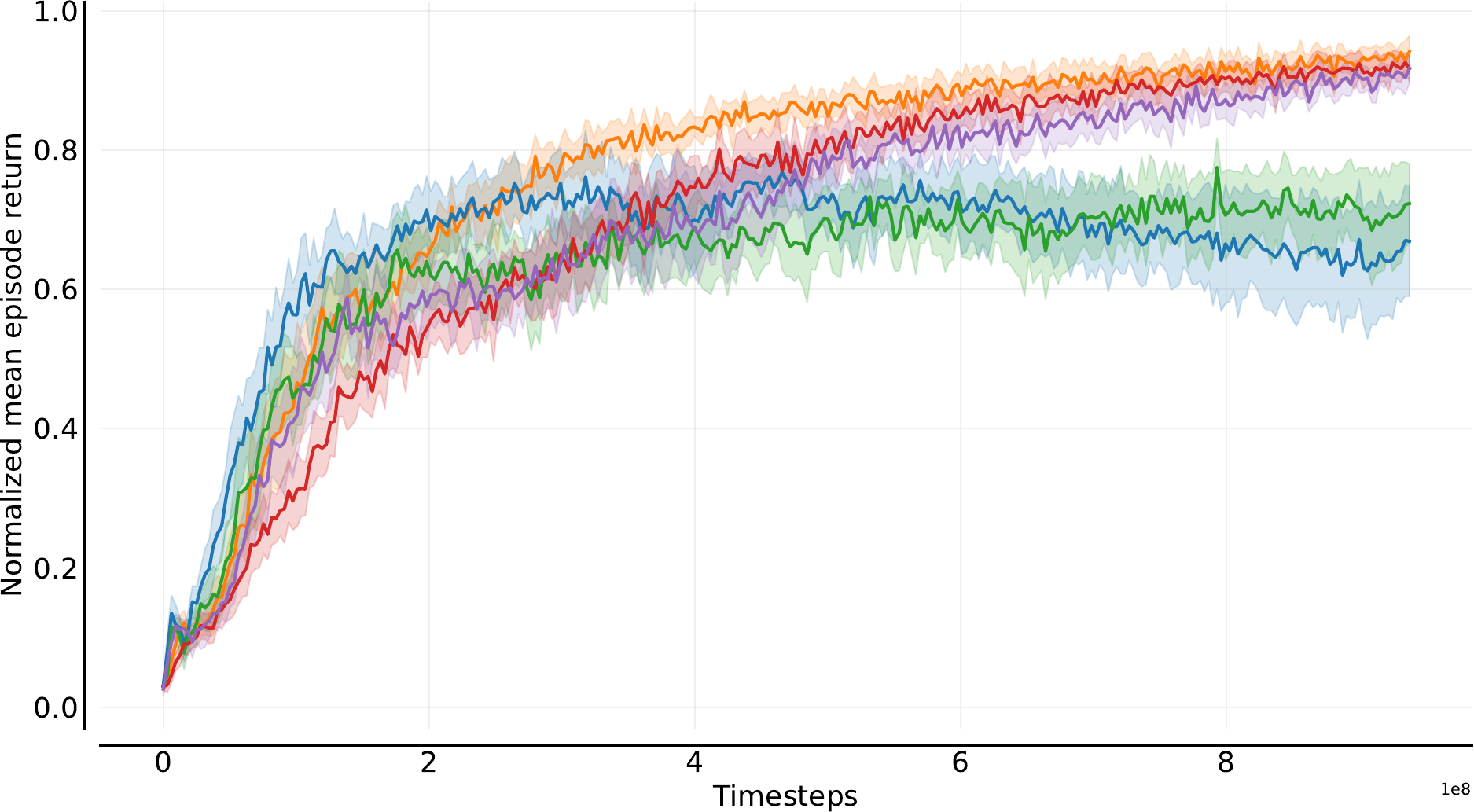}
        \caption{AeroJAX: returns}
        \label{figure:combined_ablation:aerojax_learning}
    \end{subfigure}\hfill%
    \begin{subfigure}[t]{0.315\textwidth}
        \centering
        \includegraphics[width=\linewidth]{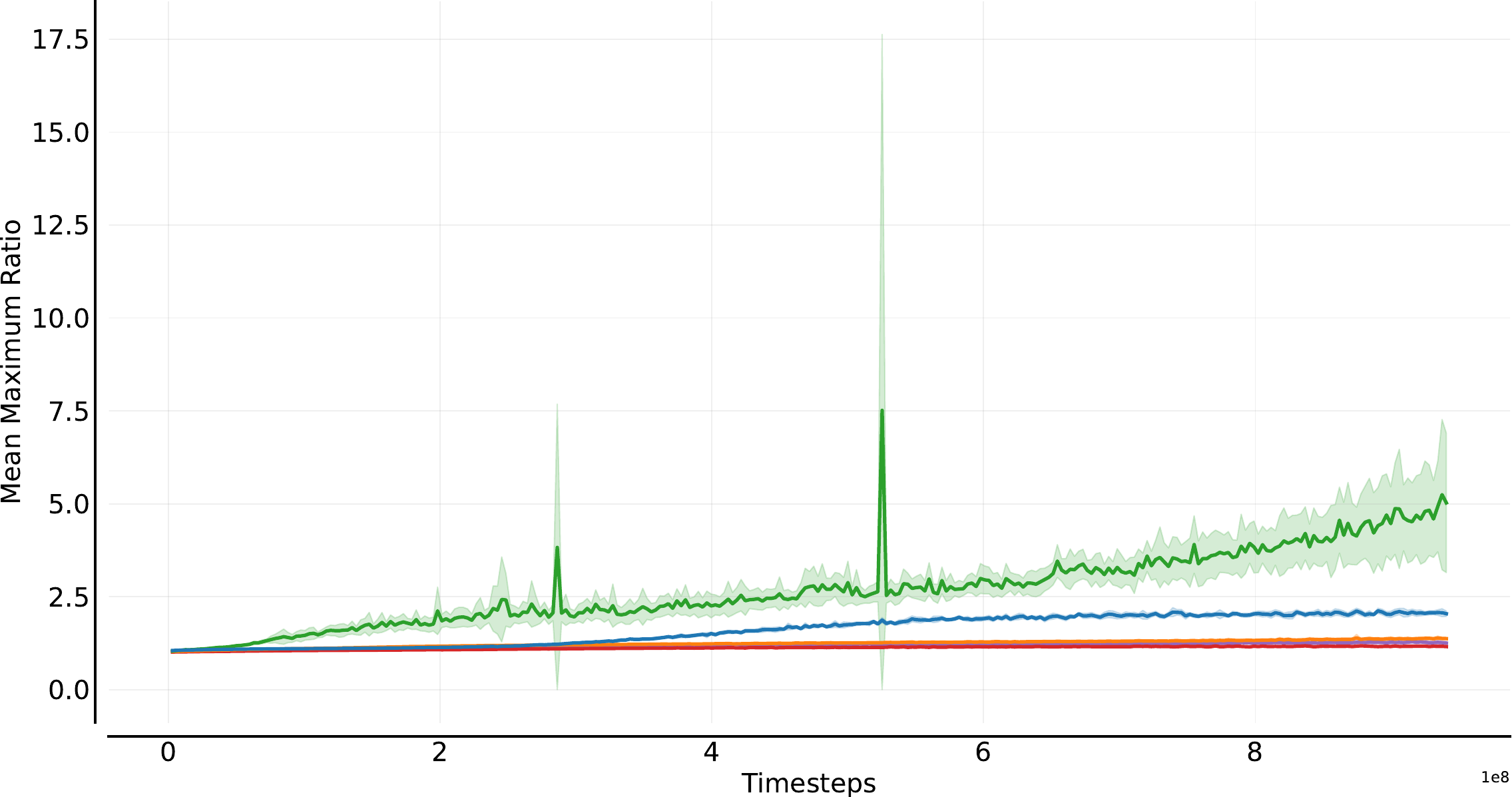}
        \caption{AeroJAX: max ratios}
        \label{figure:combined_ablation:aerojax_max_ratio}
    \end{subfigure}\hfill%
    \begin{subfigure}[t]{0.315\textwidth}
        \centering
        \includegraphics[width=\linewidth]{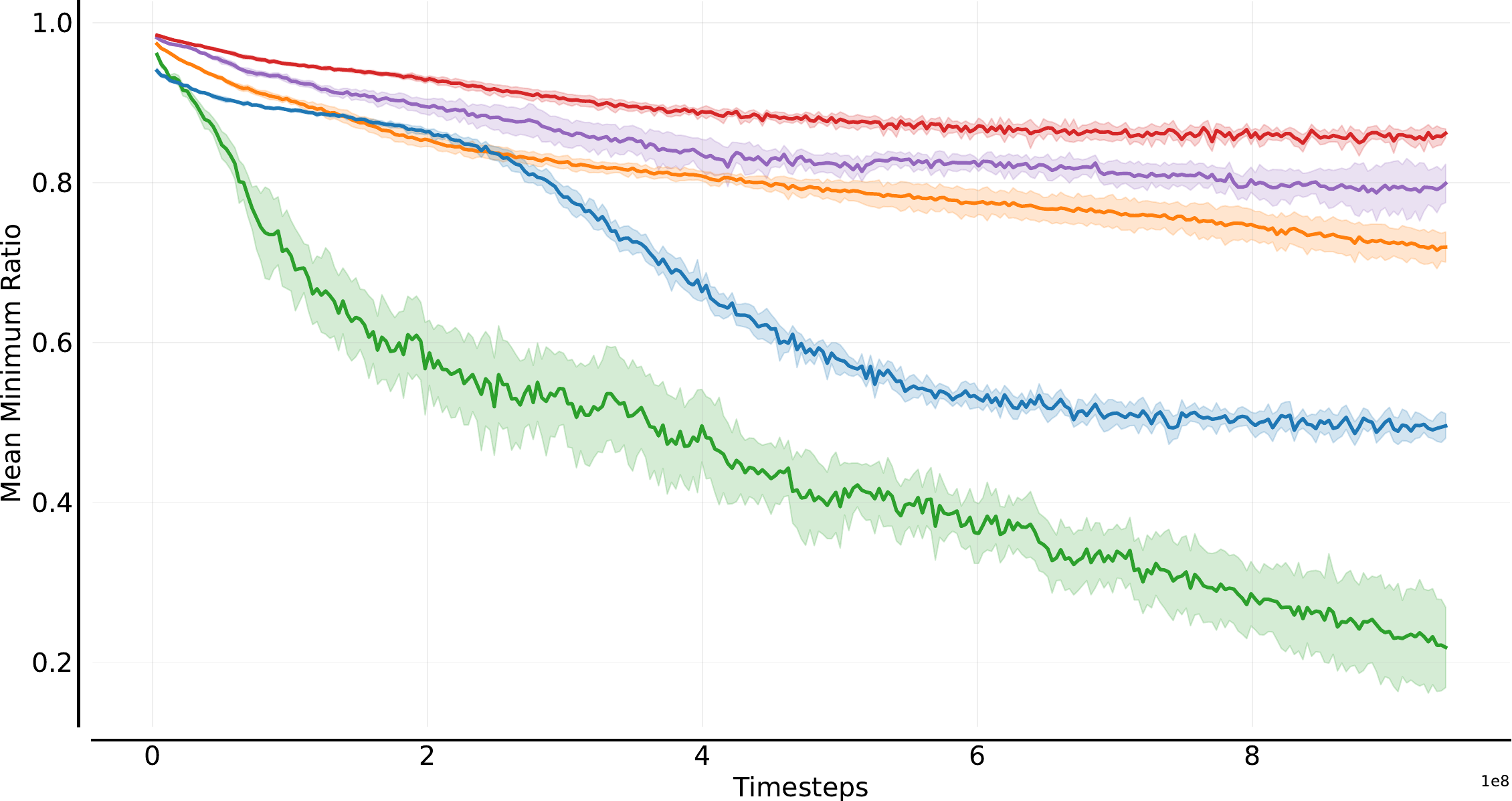}
        \caption{AeroJAX: min ratios}
        \label{figure:combined_ablation:aerojax_min_ratio}
    \end{subfigure}


    \begin{subfigure}[t]{0.315\textwidth}
        \centering
        \includegraphics[width=\linewidth]{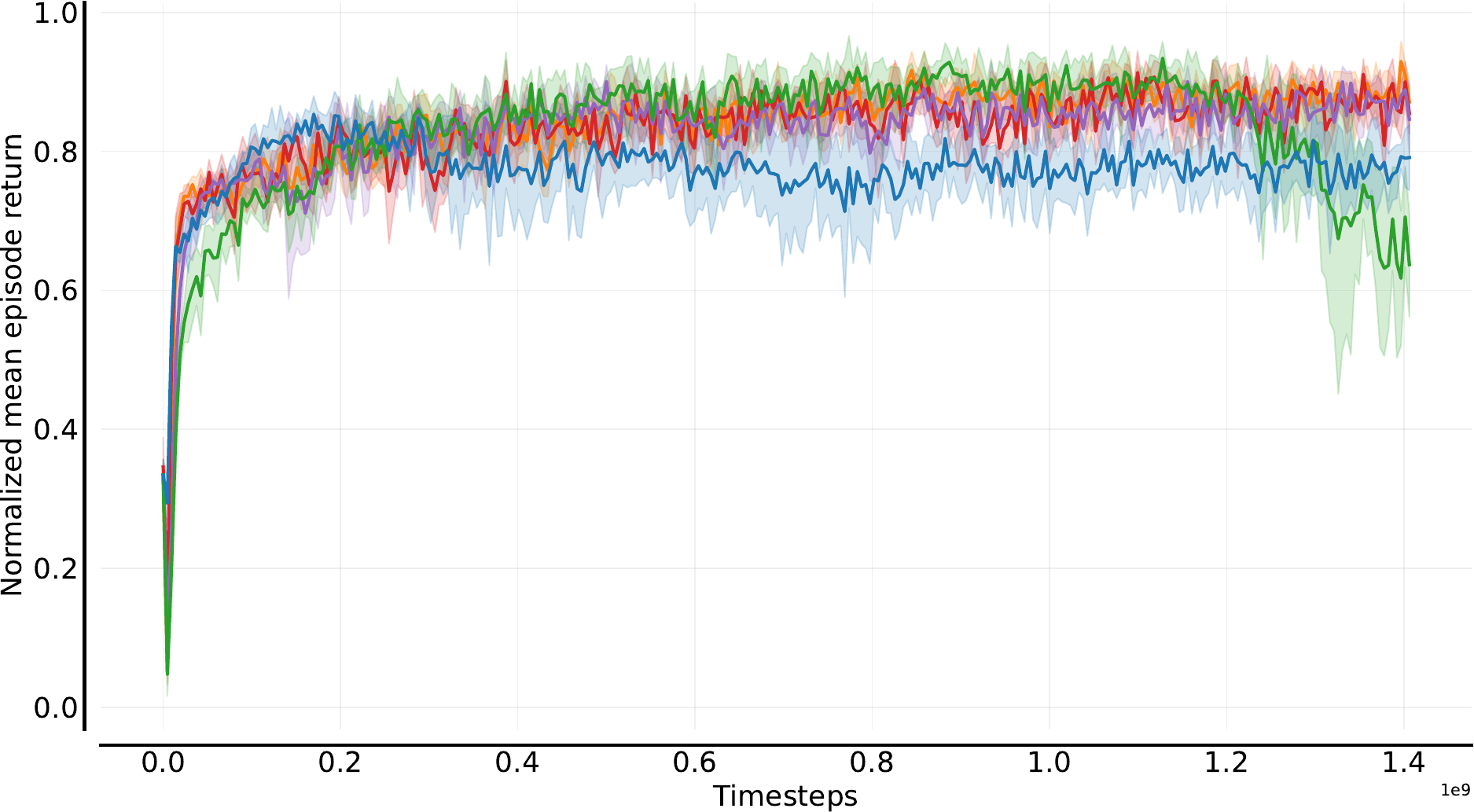}
        \caption{JaxNav: returns}
        \label{figure:combined_ablation:jaxnav_learning}
    \end{subfigure}\hfill%
    \begin{subfigure}[t]{0.315\textwidth}
        \centering
        \includegraphics[width=\linewidth]{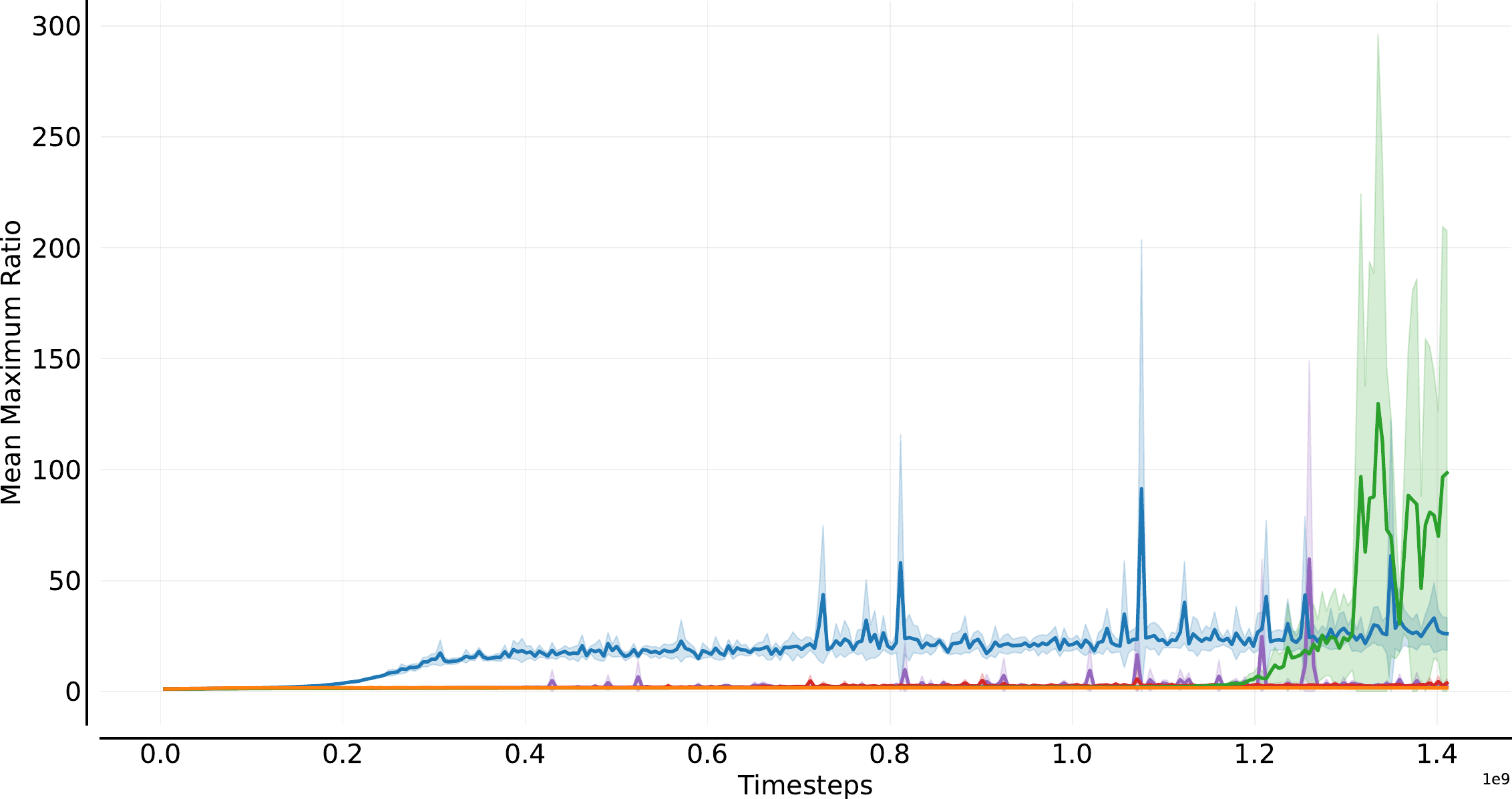}
        \caption{JaxNav: max ratios}
        \label{figure:combined_ablation:jaxnav_max_ratio}
    \end{subfigure}\hfill%
    \begin{subfigure}[t]{0.315\textwidth}
        \centering
        \includegraphics[width=\linewidth]{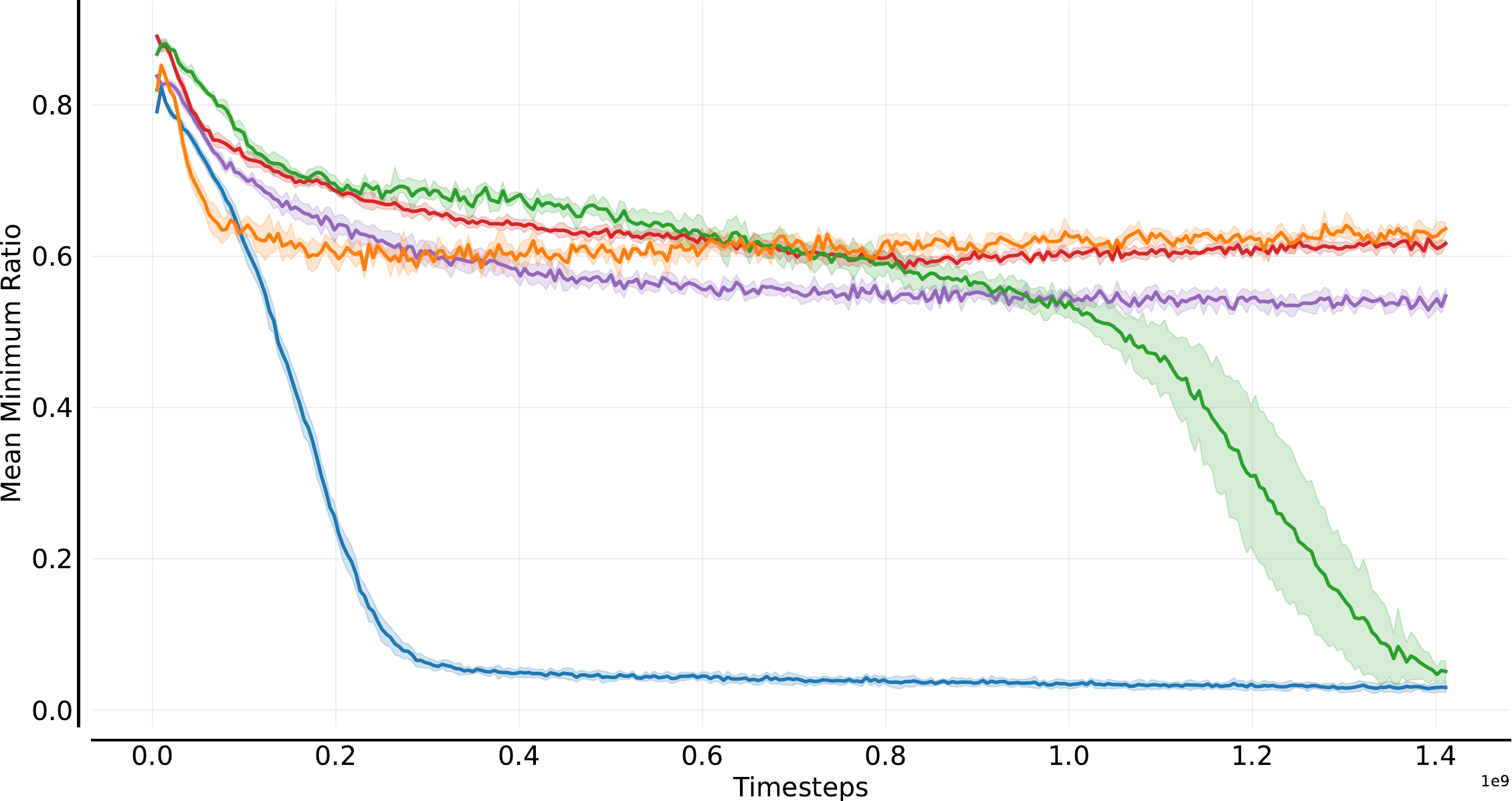}
        \caption{JaxNav: min ratios}
        \label{figure:combined_ablation:jaxnav_min_ratio}
    \end{subfigure}

    \caption{\textit{Ablation analysis across two continuous-control domains.}
    We compare MARS against boundary-flexibility controls: asymmetric MAPPO, asymmetric MASPO, and two MARS target-parameterization variants. Across AeroJAX 8v8 and JaxNav 11x11x4a, MARS variants maintain more stable ratio dynamics than the asymmetric baselines: minimum and maximum ratios spike in asymmetric MASPO, while asymmetric MAPPO's minimum ratios degrade. These failures are consistent with asymmetric MASPO's finite resistance to vanishing ratios and asymmetric MAPPO's clipped-gradient recovery failure. The stability of the MARS variants indicates that the geometric barrier is the main source of ratio stability, while flexible expansion and contraction targets primarily affect learning efficiency.}
    \label{figure:combined_ablation}
\end{figure*}

To separate objective geometry from boundary flexibility, we run long-horizon ablations on two high-variance continuous-control domains: AeroJAX 8v8 and JaxNav 11x11x4a. These tasks provide complementary diagnostics: AeroJAX combines continuous flight dynamics with multi-agent combat coordination, while JaxNav requires decentralized collision avoidance under local sensing. Figure~\ref{figure:combined_ablation} reports learning curves together with maximum- and minimum-ratio diagnostics for the ablation variants defined in Section~\ref{subsection:experimental_setup}.

\paragraph{Objective geometry.}
We first test whether MARS gains come from the geometric barrier rather than simply from using separately tuned expansion and contraction limits. In both domains, MARS maintains higher or more stable returns than the asymmetric baselines (Figures~\ref{figure:combined_ablation:aerojax_learning} and~\ref{figure:combined_ablation:jaxnav_learning}). The ratio diagnostics show that asymmetric MASPO develops declining minimum ratios and large maximum-ratio spikes, especially in AeroJAX (Figures~\ref{figure:combined_ablation:aerojax_max_ratio}--\ref{figure:combined_ablation:aerojax_min_ratio}), while asymmetric MAPPO exhibits degradation in minimum ratios despite avoiding the largest upward spikes. In JaxNav, asymmetric MAPPO shows a sharper early minimum-ratio drop, while MASPO develops later ratio instability (Figures~\ref{figure:combined_ablation:jaxnav_max_ratio}--\ref{figure:combined_ablation:jaxnav_min_ratio}). In contrast, the MARS variants maintain more stable ratio dynamics across both environments. These results support the claim that the MARS penalty shape, not only boundary flexibility, contributes to stability.

\paragraph{Domain-specific failure modes.}
The two environments show different versions of the same ratio-control problem. In AeroJAX, asymmetric MASPO shows maximum-ratio spikes together with decreasing minimum ratios, consistent with finite resistance as ratios approach zero. Asymmetric MAPPO degrades more gradually, consistent with clipped samples losing corrective gradients after leaving the nominal trust region. In JaxNav, the MAPPO failure is sharper: minimum ratios drop early in training, while MASPO learns initially but has unstable ratios. Thus, the MARS barrier stabilizes ratio dynamics across environments even when the baseline failure mode differs.


\paragraph{Boundary symmetry and parameterization.}
Finally, we compare MARS with target-parameterization ablations that keep the MARS penalty fixed while restricting how expansion and contraction targets are chosen. Multiplicatively symmetric MARS maintains stable ratio dynamics in both domains, especially in the minimum-ratio panels (Figures~\ref{figure:combined_ablation:aerojax_min_ratio} and~\ref{figure:combined_ablation:jaxnav_min_ratio}), indicating that the symmetric barrier is sufficient to avoid the most severe ratio pathologies. Additively symmetric MARS is also stable but learns more slowly, especially in AeroJAX (Figure~\ref{figure:combined_ablation:aerojax_learning}), suggesting that additive target choices are less efficient even with the MARS barrier. MARS with independently tuned expansion and contraction targets can learn faster or reach slightly higher returns, particularly in AeroJAX (Figure~\ref{figure:combined_ablation:aerojax_learning}), so target-ratio flexibility mainly affects learning speed rather than ratio stability.

\section{Conclusion}
\label{section:conclusion}

We introduced Multi-Agent Ratio Symmetry (MARS), a policy optimization objective for CTDE cooperative MARL. MARS addresses two limitations of additive ratio-based trust-region methods: clipped objectives can remove corrective gradients for ratio outliers, while finite additive penalties can allow probability collapse under large negative joint advantages. By replacing additive ratio control with a multiplicatively symmetric geometric barrier, MARS preserves ratio-space gradients and makes probability extinction non-optimal for the per-sample surrogate. Aggregating performance across 47 tasks in eight environments, MARS matches or outperforms MAPPO and MASPO, with ablations indicating that the curvature of the symmetric barrier is the primary source of stability. These results suggest that aligning ratio regularization with multiplicative probability geometry is a useful direction for robust multi-agent policy optimization.

\begin{ack}
This work was partially supported by the National Science Foundation (NSF) under Grant No.~2525087. The authors declare no competing interests.
\end{ack}

\bibliography{paper}
\bibliographystyle{icml2026}


\appendix

\section{Limitations}
\label{appendix:limitations}

MARS is designed for cooperative CTDE policy-gradient reinforcement learning with decentralized actors and joint advantage estimates. Its theoretical results characterize the per-sample ratio surrogate in ratio space, showing that the objective avoids clipped gradient plateaus and makes probability-ratio extinction non-optimal. These results do not constitute a global convergence guarantee for nonlinear neural-network training or a monotonic improvement guarantee for full multi-agent policy updates.

Our experiments focus on cooperative MARL settings with shared rewards and centralized critics. Although the benchmark suite spans 47 tasks across eight environments, it does not cover competitive, mixed-motive, or open-ended multi-agent settings. The behavior of MARS in settings with independent rewards, learned communication, population-based training, or very large heterogeneous agent populations remains an open question.

MARS also introduces target ratio parameters for expansion and contraction. We tune these parameters using the same protocol as the baselines and include ablations to separate objective geometry from boundary flexibility, but the method still requires selecting appropriate target ranges. Future work could study adaptive target selection, extensions to other multi-agent training paradigms, and stronger theoretical guarantees for full policy updates.

\section{Related Work: Divergence-Based Policy Optimization}
\label{appendix:related_work}

Divergence-based regularization has been explored in single-agent policy optimization. \citet{belousov2018fdivergenceconstrainedpolicyimprovement} study policy improvement under general (f)-divergence regularization, including a generator proportional to the MARS penalty ($\psi(r)=r+r^{-1}-2$). Similarly, AlphaPPO \citep{xu2023improving} replaces the KL regularizer in PPO-Penalty with an alpha-divergence family encompassing reverse-Pearson/Neyman geometry. These approaches establish precedents for the underlying divergence geometry but optimize policy-level objectives rather than advantage-conditioned, target-aligned sampled-ratio surrogates.

PPO-RPE \citep{kobayashi2020, kobayashi2023proximal} instead regularizes a transformed probability ratio, yielding a penalty that remains finite as the original ratio approaches zero and is not invariant under ratio inversion. In contrast, MARS applies an inversion-symmetric barrier directly to per-agent ratios within simultaneous CTDE updates, using centralized joint advantages and sample-dependent coefficients to align the surrogate with separate expansion and contraction targets.

\section{Proofs and Derivations}
\label{appendix:proofs_derivations}

\subsection{MAPPO Clipped Objective Gradient}
\label{appendix:subsection:ppo_clipped_objective_gradient}

The per-agent MAPPO clipped surrogate is
\begin{equation}
\mathcal{L}^{\mathrm{MAPPO}}_{i,t}(\theta)
=
\min\left(
r_{i,t}(\theta)\widehat{\mathbf{A}}_t,\,
\mathrm{clip}(r_{i,t}(\theta),1-\epsilon,1+\epsilon)\widehat{\mathbf{A}}_t
\right),
\end{equation}
where
\[
r_{i,t}(\theta)
=
\frac{\pi_\theta(a_{i,t}\mid h_{i,t})}
{\pi_{\theta_{\mathrm{old}}}(a_{i,t}\mid h_{i,t})}
\]
is the per-agent probability ratio and $\widehat{\mathbf{A}}_t$ is the joint advantage estimate. We derive the derivative of this objective with respect to $r_{i,t}$ away from the clipping kink points. At the clipping boundaries, the derivative can be interpreted using subgradients; these measure-zero cases do not affect the truncation argument.

\paragraph{Inside the trust region.}
If $|r_{i,t}(\theta)-1|\le \epsilon$, then
\[
\mathrm{clip}(r_{i,t}(\theta),1-\epsilon,1+\epsilon)=r_{i,t}(\theta).
\]
Therefore,
\begin{equation}
\mathcal{L}^{\mathrm{MAPPO}}_{i,t}
=
r_{i,t}(\theta)\widehat{\mathbf{A}}_t,
\qquad
\frac{\partial \mathcal{L}^{\mathrm{MAPPO}}_{i,t}}{\partial r_{i,t}}
=
\widehat{\mathbf{A}}_t.
\end{equation}

\paragraph{Above the trust region.}
If $r_{i,t}(\theta)>1+\epsilon$, then the clipped ratio is $1+\epsilon$, and
\[
\mathcal{L}^{\mathrm{MAPPO}}_{i,t}
=
\min\left(
r_{i,t}\widehat{\mathbf{A}}_t,\,
(1+\epsilon)\widehat{\mathbf{A}}_t
\right).
\]
If $\widehat{\mathbf{A}}_t>0$, the clipped term is smaller, so the objective is constant with respect to $r_{i,t}$ and the derivative is zero. If $\widehat{\mathbf{A}}_t<0$, the unclipped linear term is smaller, so the derivative is $\widehat{\mathbf{A}}_t$.

\paragraph{Below the trust region.}
If $r_{i,t}(\theta)<1-\epsilon$, then the clipped ratio is $1-\epsilon$, and
\[
\mathcal{L}^{\mathrm{MAPPO}}_{i,t}
=
\min\left(
r_{i,t}\widehat{\mathbf{A}}_t,\,
(1-\epsilon)\widehat{\mathbf{A}}_t
\right).
\]
If $\widehat{\mathbf{A}}_t>0$, the unclipped linear term is smaller, so the derivative is $\widehat{\mathbf{A}}_t$. If $\widehat{\mathbf{A}}_t<0$, the clipped term is smaller, so the objective is constant with respect to $r_{i,t}$ and the derivative is zero.

Combining these cases gives
\begin{equation}
    \frac{\partial \mathcal{L}_{i,t}^{\mathrm{MAPPO}}}{\partial r_{i,t}}
    =
    \begin{cases}
      \widehat{\mathbf{A}}_t,
      & \text{if } |r_{i,t}(\theta)-1| < \epsilon
      \text{ or }
      \mathrm{sign}(\widehat{\mathbf{A}}_t)(r_{i,t}(\theta)-1) < 0, \\
      0, & \text{otherwise}.
   \end{cases}
\end{equation}
Thus, MAPPO removes the ratio-gradient exactly for samples that cross the clipping boundary in the advantage-improving direction.

\subsection{Symmetric Penalty Derivation}
\label{appendix:subsection:symmetric_objective_derivation}

MASPO uses an additive quadratic penalty on the probability ratio,
\begin{equation}
\psi_{\mathrm{MASPO}}(r)=(r-1)^2.
\end{equation}
For probability ratios, the natural inverse of expansion by $r$ is contraction by $1/r$. To construct a penalty invariant under ratio inversion, we symmetrize the MASPO penalty by taking the geometric mean of the penalty evaluated at $r$ and at $1/r$:
\begin{equation}
\psi_{\mathrm{MARS}}(r)
=
\sqrt{
\psi_{\mathrm{MASPO}}(r)
\psi_{\mathrm{MASPO}}(1/r)
},
\qquad r>0.
\end{equation}
Now,
\begin{equation}
\psi_{\mathrm{MASPO}}(1/r)
=
\left(\frac{1}{r}-1\right)^2
=
\left(\frac{1-r}{r}\right)^2
=
\frac{(r-1)^2}{r^2}.
\end{equation}
Substituting,
\begin{equation}
\psi_{\mathrm{MARS}}(r)
=
\sqrt{
(r-1)^2
\frac{(r-1)^2}{r^2}
}
=
\sqrt{
\frac{(r-1)^4}{r^2}
}.
\end{equation}
Since $r>0$ and $(r-1)^2\ge 0$,
\begin{equation}
\sqrt{
\frac{(r-1)^4}{r^2}
}
=
\frac{(r-1)^2}{r}.
\end{equation}
Expanding yields
\begin{equation}
\psi_{\mathrm{MARS}}(r)
=
\frac{(r-1)^2}{r}
=
\frac{r^2-2r+1}{r}
=
r+\frac{1}{r}-2.
\end{equation}
This penalty is zero at $r=1$, symmetric under inversion, and diverges as $r\to 0^+$.

\subsection{Proof of Proposition~\ref{prop:barrier}: Barrier Against Probability Extinction}
\label{appendix:subsection:prop_barrier}

Let $r\in(0,\infty)$, let $\widehat{\mathbf{A}}\in\mathbb{R}$ be finite, and let $\alpha>0$. Consider
\begin{equation}
\mathcal{L}(r)
=
r\widehat{\mathbf{A}}
-
\alpha\psi_{\mathrm{MARS}}(r),
\qquad
\psi_{\mathrm{MARS}}(r)
=
r+\frac{1}{r}-2.
\end{equation}

Substituting the penalty gives
\begin{equation}
\mathcal{L}(r)
=
r\widehat{\mathbf{A}}
-
\alpha r
-
\frac{\alpha}{r}
+
2\alpha.
\end{equation}

\paragraph{Objective limit.}
As $r\to 0^+$, the terms $r\widehat{\mathbf{A}}$ and $-\alpha r$ vanish, and the constant $2\alpha$ remains finite. The dominant term is $-\alpha/r$. Since $\alpha>0$,
\begin{equation}
\lim_{r\to 0^+}
-\frac{\alpha}{r}
=
-\infty.
\end{equation}
Therefore,
\begin{equation}
\lim_{r\to 0^+}
\mathcal{L}(r)
=
-\infty.
\end{equation}

\paragraph{Gradient limit.}
Differentiating with respect to $r$,
\begin{equation}
\frac{\partial \mathcal{L}}{\partial r}
=
\widehat{\mathbf{A}}
-
\alpha
\left(
1-\frac{1}{r^2}
\right)
=
\widehat{\mathbf{A}}-\alpha+\frac{\alpha}{r^2}.
\end{equation}
As $r\to 0^+$, the finite terms $\widehat{\mathbf{A}}-\alpha$ are dominated by $\alpha/r^2$. Since $\alpha>0$,
\begin{equation}
\lim_{r\to 0^+}
\frac{\alpha}{r^2}
=
+\infty.
\end{equation}
Hence,
\begin{equation}
\lim_{r\to 0^+}
\frac{\partial \mathcal{L}}{\partial r}
=
+\infty.
\end{equation}

Thus, probability extinction is not an optimizer of the per-sample ratio surrogate, and in a neighborhood of $r=0$ the ratio-gradient points toward increasing $r$.

\subsection{Proof of Proposition~\ref{prop:no_gradient_truncation}: No Gradient Truncation}
\label{appendix:subsection:no_gradient_truncation}

Let $r\in(0,\infty)$, let $\widehat{\mathbf{A}}\in\mathbb{R}$ be finite, and let $\alpha>0$. The MARS surrogate is
\begin{equation}
\mathcal{L}(r)
=
r\widehat{\mathbf{A}}
-
\alpha
\left(
r+\frac{1}{r}-2
\right).
\end{equation}

\paragraph{Smoothness.}
The function $\mathcal{L}$ is a linear combination of polynomial terms and the rational term $1/r$. The only singularity of $1/r$ occurs at $r=0$, which is outside the domain $(0,\infty)$. Therefore, $\mathcal{L}$ is $C^\infty$ on $(0,\infty)$.

\paragraph{No gradient truncation.}
The ratio-gradient is
\begin{equation}
\frac{\partial \mathcal{L}}{\partial r}
=
\widehat{\mathbf{A}}
-
\alpha
+
\frac{\alpha}{r^2}.
\end{equation}
Suppose the gradient vanishes. Then
\begin{equation}
\widehat{\mathbf{A}}
-
\alpha
+
\frac{\alpha}{r^2}
=
0,
\end{equation}
which implies
\begin{equation}
\frac{\alpha}{r^2}
=
\alpha-\widehat{\mathbf{A}}.
\end{equation}
If $\alpha-\widehat{\mathbf{A}}\le 0$, this equation has no solution on $r>0$. If $\alpha-\widehat{\mathbf{A}}>0$, then
\begin{equation}
r
=
\sqrt{
\frac{\alpha}{\alpha-\widehat{\mathbf{A}}}
},
\end{equation}
which gives exactly one positive stationary point.

Therefore, the zero-gradient set is either empty or a singleton. In particular, there is no nontrivial interval on which $\partial \mathcal{L}/\partial r=0$. Hence, unlike a clipped objective, the MARS trust-region mechanism introduces no gradient truncation in ratio space.

\subsection{Trust Region Boundaries}
\label{appendix:subsection:trust_region_boundaries}

The MARS ratio-gradient for agent $i$ at timestep $t$ is
\begin{equation}
\frac{\partial \mathcal{L}_{i,t}^{\mathrm{MARS}}}{\partial r_{i,t}}
=
\widehat{\mathbf{A}}_t
-
\alpha_t
\left(
1-\frac{1}{r_{i,t}^2}
\right).
\end{equation}
To align the stationary point with an intended trust-region boundary, we choose a target ratio $r_{\mathrm{target}}>0$ with $r_{\mathrm{target}}\neq 1$ and enforce
\begin{equation}
\left.
\frac{\partial \mathcal{L}_{i,t}^{\mathrm{MARS}}}{\partial r_{i,t}}
\right|_{r_{i,t}=r_{\mathrm{target}}}
=
0.
\end{equation}
This gives
\begin{equation}
0
=
\widehat{\mathbf{A}}_t
-
\alpha_t
\left(
1-r_{\mathrm{target}}^{-2}
\right),
\end{equation}
and therefore
\begin{equation}
\alpha_t
=
\frac{\widehat{\mathbf{A}}_t}
{1-r_{\mathrm{target}}^{-2}}.
\end{equation}

\paragraph{Additive $\epsilon$-aligned target.}
If the target is expressed using the standard additive boundary,
\begin{equation}
r_{\mathrm{target}}
=
1+\mathrm{sign}(\widehat{\mathbf{A}}_t)\epsilon,
\end{equation}
then
\begin{equation}
\alpha_t
=
\frac{\widehat{\mathbf{A}}_t}
{
1-
\left(
1+\mathrm{sign}(\widehat{\mathbf{A}}_t)\epsilon
\right)^{-2}
}.
\end{equation}

\paragraph{Multiplicative target ratios.}
For MARS, we specify the expansion and contraction targets directly:
\begin{equation}
r_{\mathrm{target}} =
\begin{cases}
B_{\mathrm{upper}}, & \widehat{\mathbf{A}}_t \ge 0,\\
B_{\mathrm{lower}}, & \widehat{\mathbf{A}}_t < 0.
\end{cases}
\end{equation}
This yields
\begin{equation}
\alpha_t
=
\begin{cases}
\dfrac{\widehat{\mathbf{A}}_t}{1-B_{\mathrm{upper}}^{-2}},
& \widehat{\mathbf{A}}_t>0,\\[8pt]
\dfrac{\widehat{\mathbf{A}}_t}{1-B_{\mathrm{lower}}^{-2}},
& \widehat{\mathbf{A}}_t<0.
\end{cases}
\end{equation}
When $B_{\mathrm{upper}}>1$ and $B_{\mathrm{lower}}<1$, $\alpha_t>0$ in both nonzero-advantage regimes. For $\widehat{\mathbf{A}}_t>0$, the numerator and denominator are both positive. For $\widehat{\mathbf{A}}_t<0$, the numerator is negative and the denominator is negative because $B_{\mathrm{lower}}^{-2}>1$. If $\widehat{\mathbf{A}}_t=0$, the corresponding policy-gradient term is zero; in implementation, this sample can be skipped for the actor update or assigned $\alpha_t=0$.

\section{Environments}
\label{appendix:section:environments}

\subsection{AeroJAX}
\label{appendix:subsection:aerojax}
\begin{figure}[H]
    \centering
    \includegraphics[width=0.5\linewidth]{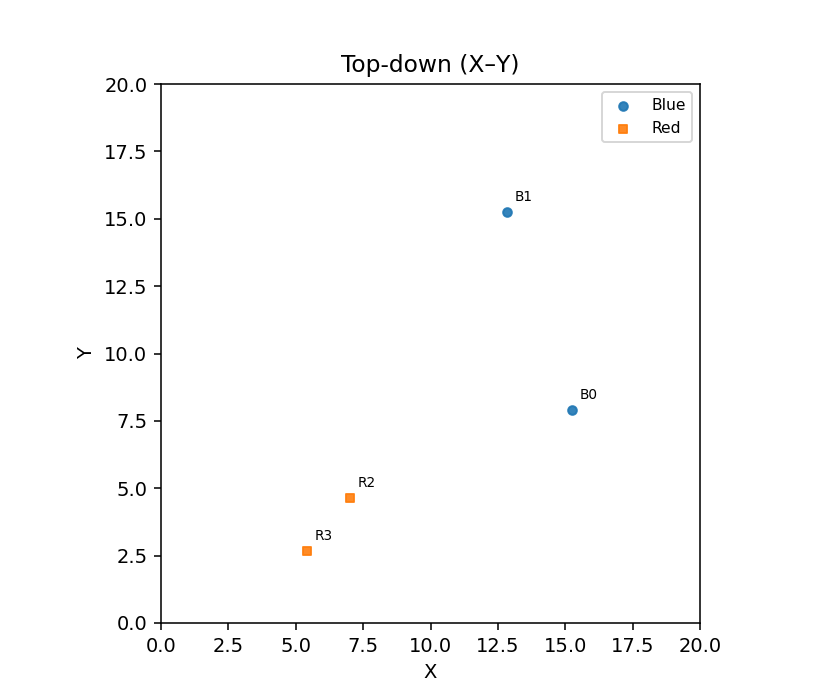}
    \caption{Rendering of a two versus two dogfighting AeroJAX scenario.}
    \label{fig:aerojax_render}
\end{figure}

AeroJAX is a JAX-native multi-agent aerial combat environment built from the Gigastep framework~\citep{lechner2023gigastep}. Whereas Gigastep emphasizes high-throughput, arcade-style multi-agent simulation, AeroJAX increases the fidelity of the control problem by introducing decoupled 3D flight dynamics and attitude-dependent combat sensing. Agents must manage six-degree-of-freedom maneuvering, energy loss, lift-induced drag, turn coordination, and throttle control while engaging heuristic opponents in continuous 3D dogfighting scenarios.

A key distinction from the original Gigastep setting is that sensing and combat are tied to aircraft attitude. AeroJAX uses view and weapons cones that rotate with each aircraft's roll and pitch, so maintaining target lock requires coordinated control of orientation, speed, and relative position rather than planar pursuit alone. This makes the benchmark distinct from simplified kinematic environments such as MPE, JaxNav, and Search and Rescue: the learning problem combines continuous control, nonlinear flight coupling, partial observability, and multi-agent combat coordination. As a result, AeroJAX provides a high-variance stress test for policy optimization stability under dense ratio changes and complex dynamics.

\paragraph{Observation space.}
Each agent observes the environment through a limited view cone defined by specific horizontal and vertical angles and a maximum depth. The observation vector includes ego-specific flight data (roll, pitch, angle of attack, flight path angle) and a sorted list of detected entities. Observed attributes for these entities include relative position, heading, speed, health, team alignment, and unit type. Detection is stochastic based on distance, but agents can share target information with allies within a defined communication range.

\paragraph{Action space.}
The environment utilizes a continuous multi-dimensional action space representing aircraft control surfaces. Agents output a 3-dimensional vector corresponding to aileron command (roll rate), elevator command (angle of attack rate), and throttle (thrust fraction).

\paragraph{Reward.}
Agents achieve sparse rewards for killing enemies and sparse penalties for ally deaths. Winning the engagement (ending the episode with more allies alive than enemies) also provides a sparse reward.

\subsection{PaxMen}
\label{appendix:subsection:paxmen}
\begin{figure}[H]
    \centering
    \includegraphics[width=0.5\linewidth]{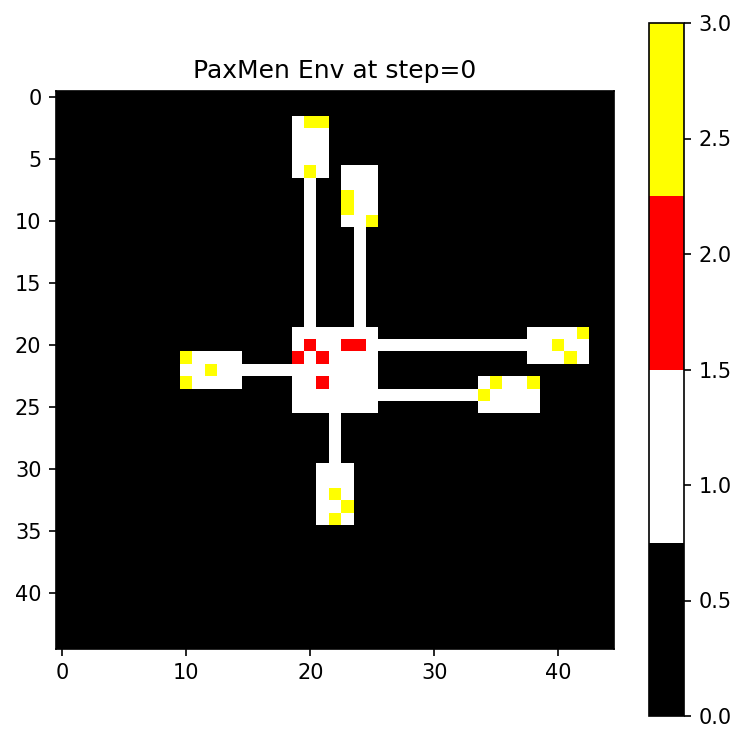}
    \caption{Rendering of an example six-agent PaxMen layout.}
    \label{fig:paxmen_render}
\end{figure}

PaxMen is a JAX-native cooperative exploration benchmark based on the PacMen environment~\citep{li2021celebrating}. The original PacMen task uses a shared multi-agent setting in which agents coordinate to collect dots in a structured maze. PaxMen extends this setting by supporting larger team sizes, including four-, five-, and six-agent configurations, and by introducing stochastic layout variation across episodes. Rather than training on a single static map, agents encounter a fixed set of maze topologies with varying corridor orientations and lengths.

This design makes PaxMen a diagnostic benchmark for decentralized coordinated exploration. Agents begin in a central hub and must disperse into branching corridors to maximize coverage without explicit communication. Because agents are homogeneous and share rewards, successful policies must break symmetry and adopt complementary roles; policies that send multiple agents down the same corridor waste exploration capacity. This distinguishes PaxMen from combat micro-management tasks such as SMAX and routing or transport benchmarks such as RWARE and Connector. PaxMen instead isolates the challenge of sustained exploration and role differentiation under partial observability.

\paragraph{Observation space.}
Each agent observes a partial, local view of the environment defined by a $5 \times 5$ grid centered on the agent. The observation is a flattened vector containing values representing walls, empty space, other agents, and dots.

\paragraph{Action space.}
The action space is discrete and consists of five actions: four cardinal movement
directions (Up, Down, Left, Right) and a dedicated Eat action. Movement into walls is invalid and results in no position change.

\paragraph{Reward.}
Agents receive a global reward calculated as the sum of all individual agent contributions at each timestep. A positive reward of +1.0 is generated for consuming a dot; if multiple agents consume the same dot simultaneously, the reward is split such that the total team gain remains +1.0. A small step penalty of -0.025 is applied to encourage efficient exploration. The dots respawn once all have been consumed.

\subsection{JaxNav}
\label{appendix:subsection:jaxnav}
\begin{figure}[H]
    \centering
    \includegraphics[width=0.4\linewidth]{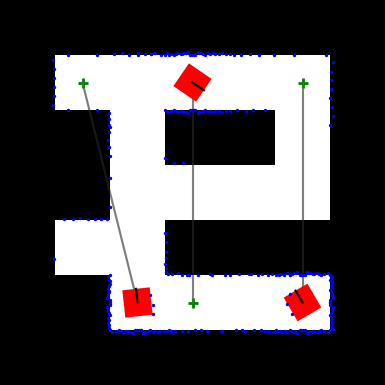}
    \caption{Rendering of an example three-agent JaxNav navigation scenario.}
    \label{fig:jaxnav_render}
\end{figure}
JaxNav is a JAX-based continuous 2D navigation environment for differential-drive robots, originally introduced by \citet{rutherford2024no} and included in the JaxMARL suite \citep{rutherford2024jaxmarl}. The environment tasks agents with navigating to specific goal regions while avoiding collisions with static obstacles and other dynamic agents. It relies exclusively on local range sensing without inter-agent communication, requiring agents to learn decentralized collision avoidance strategies.

\paragraph{Observation space.}
Each agent observes a local vector comprising three primary components: a set of 200 LiDAR range readings from a $360^\circ$ scan (normalized by the maximum range of 6m), the agent's current linear and angular velocities, and the relative goal direction in polar coordinates. The goal distance is clipped to the maximum LiDAR range to preserve local-only sensing constraints.

\paragraph{Action space.}
The environment features a continuous 2D action space governing differential-drive dynamics. Agents output a vector specifying the target linear velocity ($v$) and angular velocity ($\omega$), which are executed subject to acceleration limits and kinematic constraints.

\paragraph{Reward.}
Agents receive a dense shaped reward based on their progress toward the goal (weighted at 0.25) combined with a sparse bonus of +4.0 for successfully reaching the target radius (0.3m). To ensure safety, penalties are applied for collisions (-4.0) and for breaching a minimum safety distance (LiDAR proximity < 0.3m). A small step penalty of -0.01 is applied to encourage time-optimal paths. In the multi-agent setting, the final reward is a convex combination of the agent's individual reward and the collective team reward ($\lambda=0.5$).

\subsection{Search and Rescue}
\label{appendix:subsection:s&r}
\begin{figure}[H]
    \centering
    \includegraphics[width=0.5\linewidth]{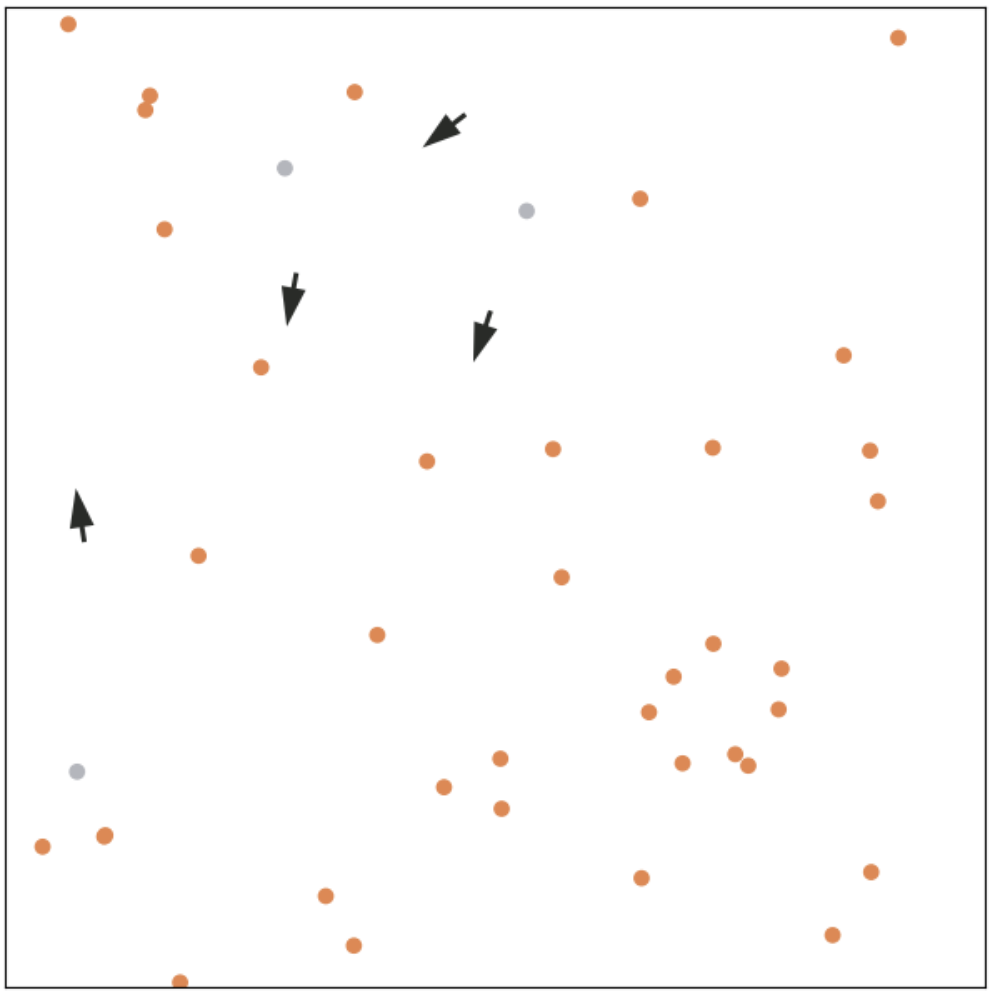}
    \caption{Rendering of an example four-agent Search and Rescue scenario.}
    \label{fig:sar_render}
\end{figure}
Search and Rescue is a continuous-space multi-agent coordination task from the Jumanji suite \citep{bonnet2024jumanji}. A team of homogeneous searchers operates within a 2D square arena featuring wrap-around (toroidal) boundaries. The objective is to locate a set of potentially moving targets within a fixed time horizon. A target is considered "found" when a searcher navigates within a specific contact radius while having the target within its view cone.

\paragraph{Observation space.}
Agents perceive the environment through a segmented, ray-based local view. The primary observation is a multi-channel ray array where each ray encodes the normalized distance to the nearest entity along that vector. Channels separate entity types into other searchers, found targets, and unfound targets. Additionally, the observation vector includes lightweight scalar context, such as the fraction of targets remaining and the current timestep fraction.

\paragraph{Action space.}
Agents operate in a continuous multi-dimensional action space. Each agent outputs a 2-dimensional vector determining its change in heading ($\Delta \theta$) and velocity ($\Delta v$). These commands are scaled by environment parameters to update the searcher's position and orientation, with speed clipped to a fixed operational interval.

\paragraph{Reward.}
Rewards are strictly event-based and issued for \emph{new} detections. An agent receives a reward of +1.0 when it detects a target that was previously unfound. If multiple agents discover the same target simultaneously, the reward is split evenly to maintain a constant team utility. By default, the magnitude of these rewards decays linearly over the episode duration to encourage rapid discovery.

\subsection{Other environments}
We refer to the Appendix in \citet{mahjoub2025sable} for thorough descriptions of the other environments we use (Connector, MPE, RWARE, and SMAX) rather than re-writing them here. 

\section{Hyperparameters}
\label{appendix:section:hyperparameters}

\subsection{Overview}
We use Optuna \citep{akiba2019optuna}, a Bayesian optimization framework, to systematically search for optimal hyperparameters across all algorithms. Our tuning procedure covers MARS, MAPPO, MASPO, and their ablation variants.

We employ the Tree-structured Parzen Estimator (TPE) sampler. MARS, asymmetric MAPPO, and asymmetric MASPO are allocated 80 trials for their tuning budget, while MAPPO, MASPO, and additive symmetric MARS and multiplicative symmetric MARS are allocated 40 trials for their tuning budget. This larger tuning budget accounts for the additional trust-region parameter in methods with independently tuned upper and lower bounds.

\subsection{Evaluation Protocol}

Each trial follows a rigorous multi-seed evaluation protocol to ensure statistical reliability:

\begin{enumerate}
    \item \textbf{Seeds:} Five fixed seeds are used for each trial
    \item \textbf{Sequential evaluation:} Seeds are evaluated one at a time with intermediate pruning checks
    \item \textbf{Performance metric:} Mean evaluation return averaged over the final 20\% of evaluation steps, which smooths end-of-training noise while focusing on final convergence quality
    \item \textbf{Trial objective:} Mean performance across all 5 seeds
    \item \textbf{Optimization direction:} Maximize evaluation return
\end{enumerate}

\subsection{Hyperparameter Search Spaces}

\subsubsection{Common Hyperparameters}

Table~\ref{tab:common_hyperparameters} shows the common hyperparameters tuned across all algorithms, inspired by the hyperparameter search space used by \citet{mahjoub2025sable}.

\begin{table}[h]
\centering
\caption{Common hyperparameters tuned across all algorithms.}
\label{tab:common_hyperparameters}
\begin{tabular}{ll}
\toprule
\textbf{Hyperparameter} & \textbf{Search Space} \\
\midrule
Actor learning rate & $\{10^{-4}, 2.5 \times 10^{-4}, 5 \times 10^{-4}\}$ \\
Critic learning rate & $\{10^{-4}, 2.5 \times 10^{-4}, 5 \times 10^{-4}\}$ \\
Entropy coefficient & $\{0.0, 10^{-5}, 10^{-2}\}$ \\
Value clipping epsilon & $\{0.05, 0.1, 0.2\}$ \\
Number of minibatches & $\{2, 4, 8\}$ \\
PPO epochs & $\{2, 4, 8\}$ \\
\bottomrule
\end{tabular}
\end{table}

\subsubsection{Algorithm-Specific Hyperparameters}
\label{appendix:subsubsection:algorithm_specific_hyperparameters}

All ablation variants use the same rollout, critic, optimizer, and shared-policy training loop as their corresponding base method. They differ only in the ratio objective or in how the trust-region target ratios are parameterized.

\paragraph{MARS.}
MARS uses the per-agent surrogate
\[
    \mathcal{L}^{\mathrm{MARS}}_{i,t}(\theta)
    =
    r_{i,t}(\theta)\widehat{\mathbf{A}}_t
    -
    \alpha_t
    \left(
        r_{i,t}(\theta)
        +
        \frac{1}{r_{i,t}(\theta)}
        -
        2
    \right),
\]
with independently tuned expansion and contraction targets
\[
r_{\mathrm{target}} =
\begin{cases}
B_{\mathrm{upper}}, & \widehat{\mathbf{A}}_t \ge 0,\\
B_{\mathrm{lower}}, & \widehat{\mathbf{A}}_t < 0,
\end{cases}
\qquad
\alpha_t =
\frac{\widehat{\mathbf{A}}_t}{1-r_{\mathrm{target}}^{-2}}.
\]
We tune
\[
B_{\mathrm{upper}} \in \{1.05, 1.1, 1.2, 2, 4, 6, 8, 10\},
\quad
B_{\mathrm{lower}} \in \{0.95, 0.9, 0.8, 0.7, 0.5, 0.3, 0.1\}.
\]

\paragraph{Multiplicatively symmetric MARS.}
This ablation keeps the MARS loss unchanged but constrains the target ratios to be reciprocal. For a single tuned target $b>1$,
\[
B_{\mathrm{upper}} = b,
\qquad
B_{\mathrm{lower}} = \frac{1}{b}.
\]
Thus expansion and contraction targets are multiplicative inverses. We tune
\[
b \in \{1.05, 1.1, 1.2, 1.5, 1.9\}.
\]

\paragraph{Additively symmetric MARS.}
This ablation also keeps the MARS loss unchanged but constrains the target ratios using MAPPO-style additive symmetry. For a single tuned target $b>1$, equivalently $\epsilon=b-1$,
\[
B_{\mathrm{upper}} = b = 1+\epsilon,
\qquad
B_{\mathrm{lower}} = 2-b = 1-\epsilon.
\]
Thus the only difference from MARS is the target-ratio parameterization used to compute $\alpha_t$. We tune
\[
b \in \{1.05, 1.1, 1.2, 1.5, 1.9\}.
\]

\paragraph{MAPPO.}
Standard MAPPO uses the clipping interval $[1-\epsilon,1+\epsilon]$:
\[
\mathcal{L}^{\mathrm{MAPPO}}_{i,t}(\theta)
=
\min\left(
r_{i,t}(\theta)\widehat{\mathbf{A}}_t,
\operatorname{clip}(r_{i,t}(\theta),1-\epsilon,1+\epsilon)
\widehat{\mathbf{A}}_t
\right).
\]
We tune
\[
\epsilon \in \{0.05, 0.1, 0.2\}.
\]

\paragraph{Asymmetric MAPPO.}
For the asymmetric MAPPO ablation, we replace the standard symmetric clipping interval with independently tuned lower and upper clipping bounds:
\[
\operatorname{clip}
\left(
r_{i,t}(\theta),
1-\epsilon_{\mathrm{lower}},
1+\epsilon_{\mathrm{upper}}
\right).
\]
The objective is otherwise identical to MAPPO. We tune
\[
\epsilon_{\mathrm{upper}} \in \{0.05, 0.1, 0.2, 1.0, 3.0, 5.0, 7.0, 9.0\},
\quad
\epsilon_{\mathrm{lower}} \in \{0.05, 0.1, 0.2, 0.3, 0.5, 0.7, 0.9\}.
\]

\paragraph{MASPO.}
Standard MASPO uses the quadratic penalty
\[
\mathcal{L}^{\mathrm{MASPO}}_{i,t}(\theta)
=
r_{i,t}(\theta)\widehat{\mathbf{A}}_t
-
\frac{|\widehat{\mathbf{A}}_t|}{2\epsilon}
\left(r_{i,t}(\theta)-1\right)^2.
\]
We tune the same single trust-region parameter as standard MAPPO:
\[
\epsilon \in \{0.05, 0.1, 0.2\}.
\]

\paragraph{Asymmetric MASPO.}
For the asymmetric MASPO ablation, we keep the MASPO quadratic penalty but choose its scale using a sign-dependent target ratio:
\[
r_{\mathrm{target}} =
\begin{cases}
1+\epsilon_{\mathrm{upper}}, & \widehat{\mathbf{A}}_t \ge 0,\\
1-\epsilon_{\mathrm{lower}}, & \widehat{\mathbf{A}}_t < 0.
\end{cases}
\]
Equivalently, the quadratic penalty coefficient is chosen so that the one-step MASPO surrogate has its stationary point at the corresponding upper or lower target:
\[
\mathcal{L}^{\mathrm{asym\text{-}MASPO}}_{i,t}(\theta)
=
r_{i,t}(\theta)\widehat{\mathbf{A}}_t
-
\frac{\widehat{\mathbf{A}}_t}
{2(r_{\mathrm{target}}-1)}
\left(r_{i,t}(\theta)-1\right)^2.
\]
This coefficient is positive for both $\widehat{\mathbf{A}}_t>0$ and $\widehat{\mathbf{A}}_t<0$ because $r_{\mathrm{target}}>1$ in the positive-advantage case and $r_{\mathrm{target}}<1$ in the negative-advantage case. We tune the same independently selected upper and lower values used for asymmetric MAPPO.

\paragraph{Ablation Fairness.} For fairness in our ablation study in \cref{subsection:ablation_analysis}, we re-tune MARS using the exact same asymmetric clipping bounds (converted to their target ratio forms as described in Appendix \cref{appendix:subsection:trust_region_boundaries}) as asymmetric MAPPO and asymmetric MASPO; in other words, all algorithms share an \textbf{identical} hyperparameter search space. This ensures that, within the boundary-flexibility ablation, the differentiating
factor between the three objectives is the objective itself.

\subsection{Fixed Hyperparameters}

To focus the hyperparameter search, we fix the following parameters across all trials (Table~\ref{tab:fixed_hyperparameters}).

\begin{table}[h]
\centering
\caption{Fixed hyperparameters across all tuning experiments.}
\label{tab:fixed_hyperparameters}
\begin{tabular}{ll}
\toprule
\textbf{Hyperparameter} & \textbf{Value} \\
\midrule
Agent ID encoding & True \\
Update batch size & 2 \\
Rollout length & 128 \\
Max gradient norm & 0.5 \\
GAE $\lambda$ & 0.95 \\
Discount factor ($\gamma$) & 0.99 \\
Value function coefficient & 0.5 \\
Learning rate decay & False \\
\bottomrule
\end{tabular}
\end{table}

\subsection{Training/Tuning Lengths}

We run each seed of each trial of each task for every environment except AeroJAX and JaxNav for approximately 40 million timesteps. Due to the higher complexity of AeroJAX and JaxNav, we allow for additional training time for those seeds, approximately 120 million timesteps. 

Final evaluation runs using tuned hyperparameters are run for twice the tuning length: approximately 240 million timesteps for AeroJAX and JaxNav and approximately 80 million
timesteps for all other environments. Ablation runs in Section~\ref{subsection:ablation_analysis}
are run for approximately one billion timesteps to evaluate long-horizon stability.

\section{Algorithm Pseudocode}
\label{appendix:section:algorithm_pseudocode}

Below is pseudocode for MARS in the same CTDE training loop used by MAPPO and MASPO, replacing only the actor objective. Although the pseudocode denotes the centralized critic as state-based, history-dependent critics such as $V(h)$ and $V(h,s)$ can offer theoretical advantages \citep{lyu2023centralized}. However, computational constraints in large-scale tasks often necessitate state-based critics $V(s)$. Our work utilizes a variety of critic inputs, depending on what was provided by the environment and the scale of the task. Additionally, the pseudocode writes the actor objective as a loss to be minimized, so the MARS surrogate and entropy bonus appear inside a leading negative sign.

\begin{algorithm}[H]
\caption{Recurrent Multi-Agent Ratio Symmetry (MARS) with CTDE}
\label{alg:mars_ctde}
\begin{algorithmic}[1]
\State \textbf{Input:} Initial policy parameters $\theta$, critic parameters $\phi$
\State \textbf{Hyperparameters:} Expansion target $B_{\text{upper}} > 1$, Contraction target $B_{\text{lower}} < 1$, learning rates $\eta_{\theta}, \eta_{\phi}$
\While{training not converged}
    \State \textbf{Data Collection (Decentralized Rollout):}
    \State Collect batch $\mathcal{D}$ by running policies $\pi_{\theta}(a_{i,t}|h_{i,t})$ for all agents $i \in \{1, \dots, N\}$.
    \State Store trajectories $\tau = \{(s_t, \mathbf{h}_t, \mathbf{a}_t, r_t, \pi_{\text{old}})\}_{t=0}^T$.
    \State \textit{// Note: Actors use local histories $h_{i,t}$; Critic uses global state $s_t$.}
    
    \State \textbf{Advantage Estimation (Centralized):}
    \State Compute joint advantages $\mathbf{\hat{A}}_t$ and value targets $R_t$ using GAE and centralized critic $V_{\phi}(s_t)$.
    
    \State \textbf{Optimization Loop:}
    \For{each epoch, minibatch $b \in \mathcal{D}$}
        \State Re-evaluate policies to get $\pi_{\theta}(a_{i,t}|h_{i,t})$ and ratios $r_{i,t}(\theta) = \frac{\pi_{\theta}(a_{i,t}|h_{i,t})}{\pi_{\text{old}}(a_{i,t}|h_{i,t})}$.
        
        \State \textbf{1. Trust Region Alignment:}
        \State $r_{\text{target}} \leftarrow \begin{cases} B_{\text{upper}} & \text{if } \mathbf{\hat{A}}_t \ge 0 \\ B_{\text{lower}} & \text{if } \mathbf{\hat{A}}_t < 0 \end{cases}$
        
        \State \textbf{2. Target-aligned penalty weight and barrier:}
        \State \textit{// Shared penalty weight derived from the joint advantage and target ratio}
        \State $\alpha_t \leftarrow \frac{\mathbf{\hat{A}}_t}{1 - (r_{\text{target}})^{-2}}$
        \State \textit{// Per-agent geometric barrier}
        \State $\psi_{\text{MARS}}(r_{i,t}) \leftarrow r_{i,t} + \frac{1}{r_{i,t}} - 2$
        
        \State \textbf{3. Independent Updates:}
        \State \textit{// Actor Update (Uses History $h_{i,t}$ and Joint Advantage $\mathbf{\hat{A}}_t$)}
        \State $\mathcal{L}_{\text{Actor}}(\theta) = -\mathbb{E}\left[ \frac{1}{N}\sum_{i=1}^N \left( r_{i,t}(\theta)\mathbf{\hat{A}}_t - \alpha_t \psi_{\text{MARS}}(r_{i,t}(\theta)) + \beta \mathcal{H}(\pi_{\theta}(\cdot|h_{i,t})) \right) \right]$
        \State $\theta \leftarrow \theta - \eta_{\theta} \nabla_{\theta} \mathcal{L}_{\text{Actor}}$
        
        \State \textit{// Critic Update (Uses State $s_t$)}
        \State $\mathcal{L}_{\text{Critic}}(\phi) = \mathbb{E}\left[ (V_{\phi}(s_t) - R_t)^2 \right]$
        \State $\phi \leftarrow \phi - \eta_{\phi} \nabla_{\phi} \mathcal{L}_{\text{Critic}}$
    \EndFor
\EndWhile
\end{algorithmic}
\end{algorithm}

\section{Per-Task Performance}
\label{appendix:section:per_task_performance}
Below we provide a learning curve breakdown at the per-task level:

\captionsetup{type=figure} 
\centering 
\includegraphics[width=0.4\textwidth]{figures/agg_learning_curves/learning_curve_legend.pdf}
\par

\begin{subfigure}{0.2\textwidth}
    \includegraphics[width=\linewidth]{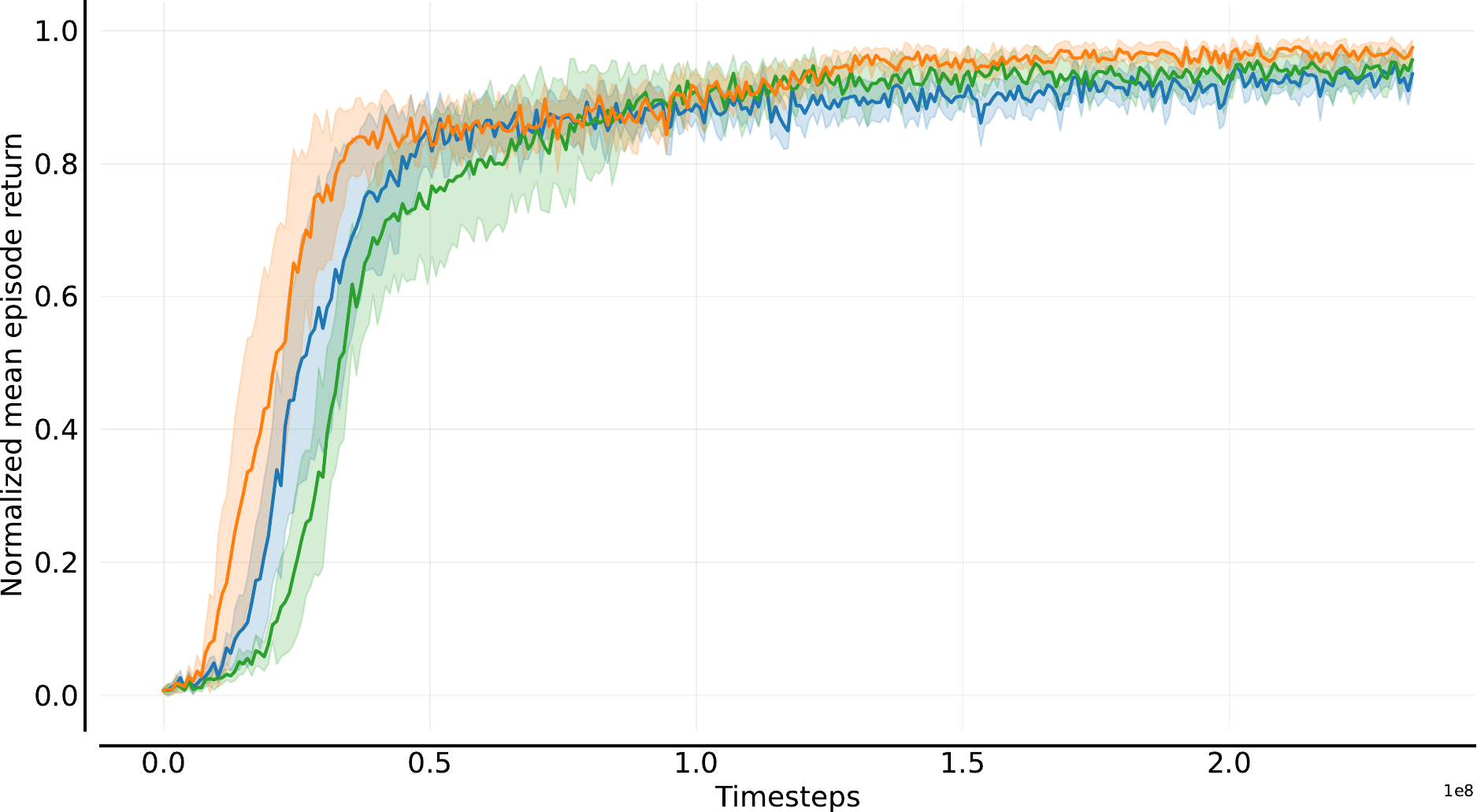}
    \caption*{\tiny AeroJAX Dogfight 2v2}
\end{subfigure}\hfill
\begin{subfigure}{0.2\textwidth}
    \includegraphics[width=\linewidth]{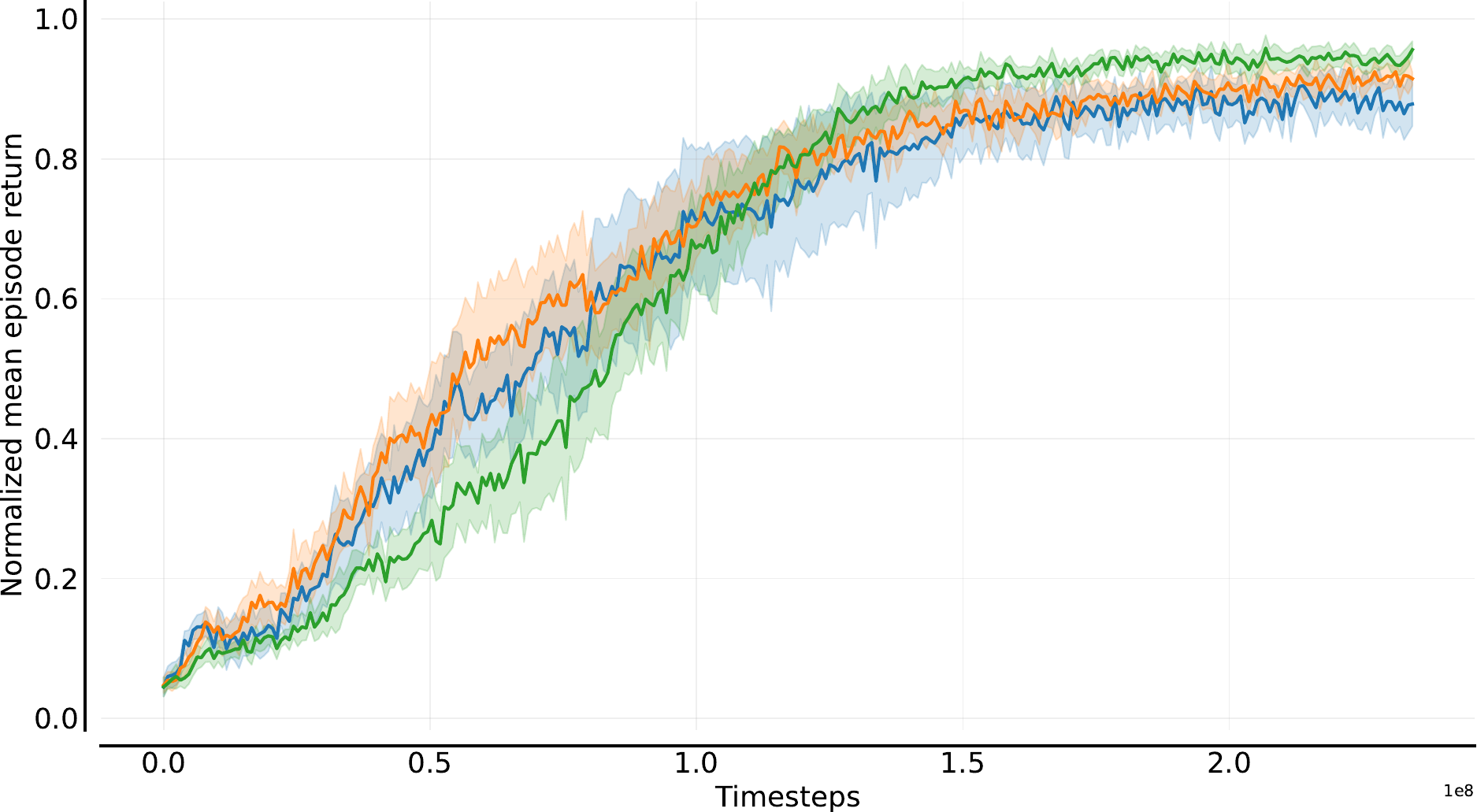}
    \caption*{\tiny AeroJAX Dogfight 4v4}
\end{subfigure}\hfill
\begin{subfigure}{0.2\textwidth}
    \includegraphics[width=\linewidth]{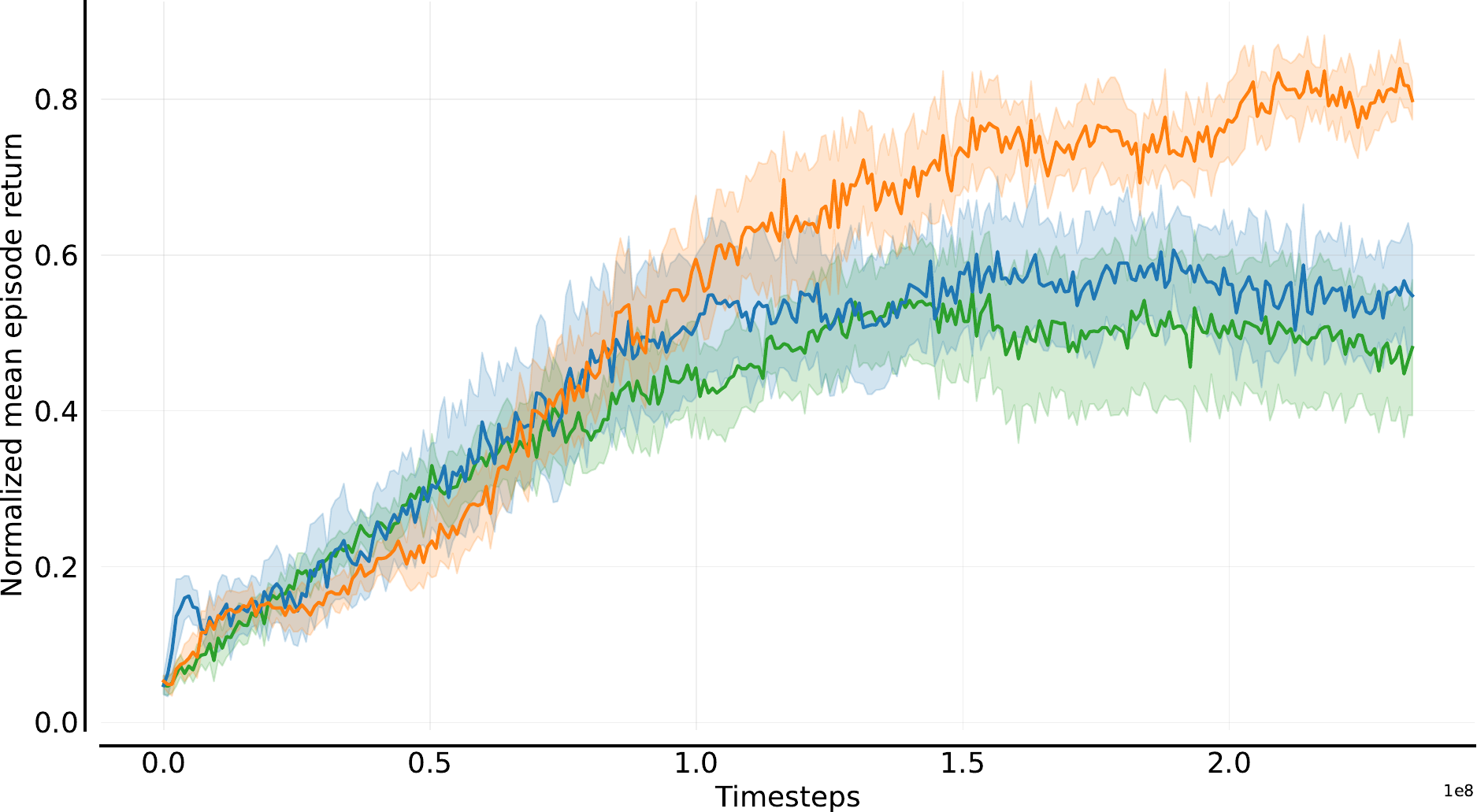}
    \caption*{\tiny AeroJAX Dogfight 8v8}
\end{subfigure}\hfill
\begin{subfigure}{0.2\textwidth}
    \includegraphics[width=\linewidth]{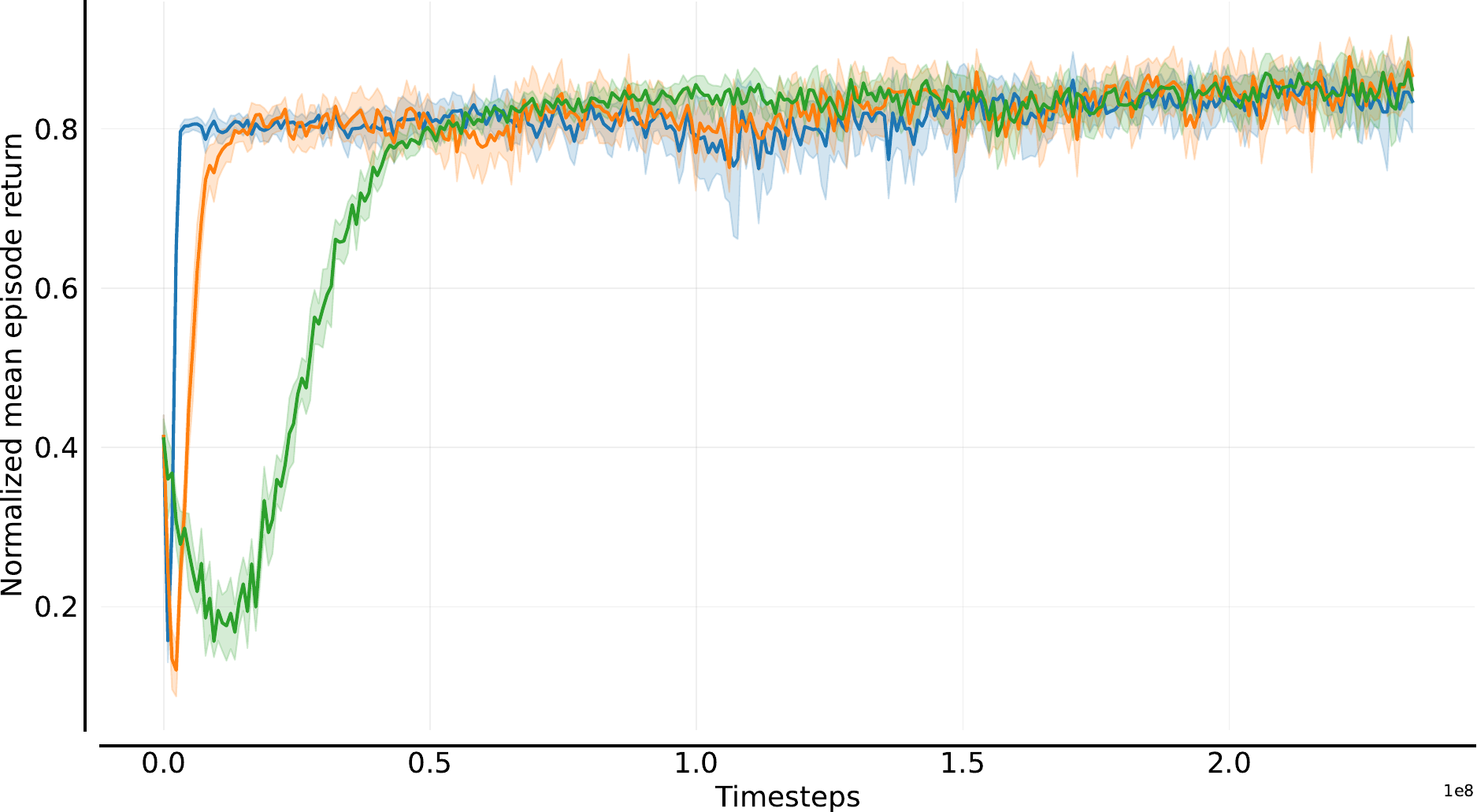}
    \caption*{\tiny JaxNav 11x11x4a}
\end{subfigure}\hfill
\begin{subfigure}{0.2\textwidth}
    \includegraphics[width=\linewidth]{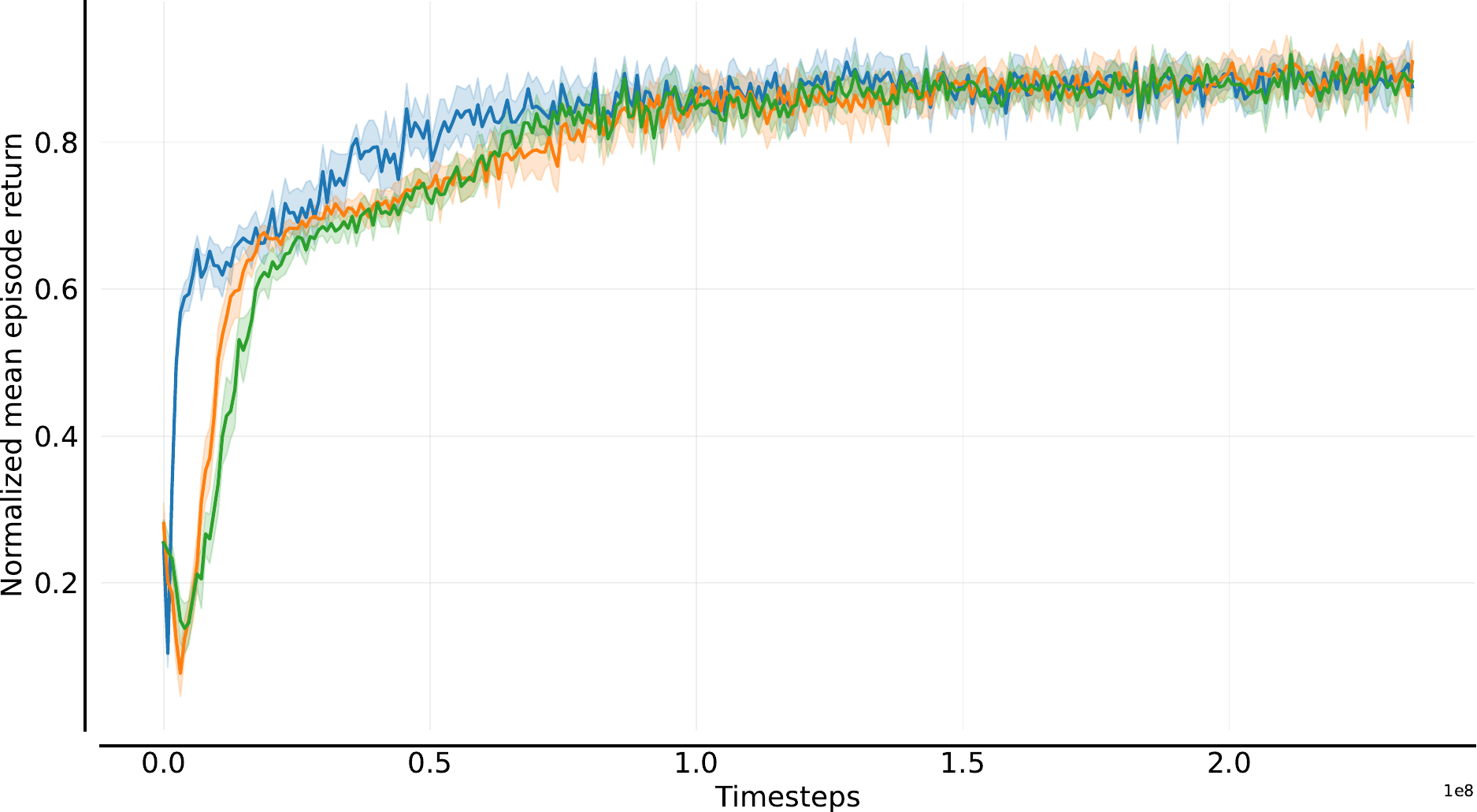}
    \caption*{\tiny JaxNav 9x9x3a}
\end{subfigure}\hfill
\par\smallskip 

\begin{subfigure}{0.2\textwidth}
    \includegraphics[width=\linewidth]{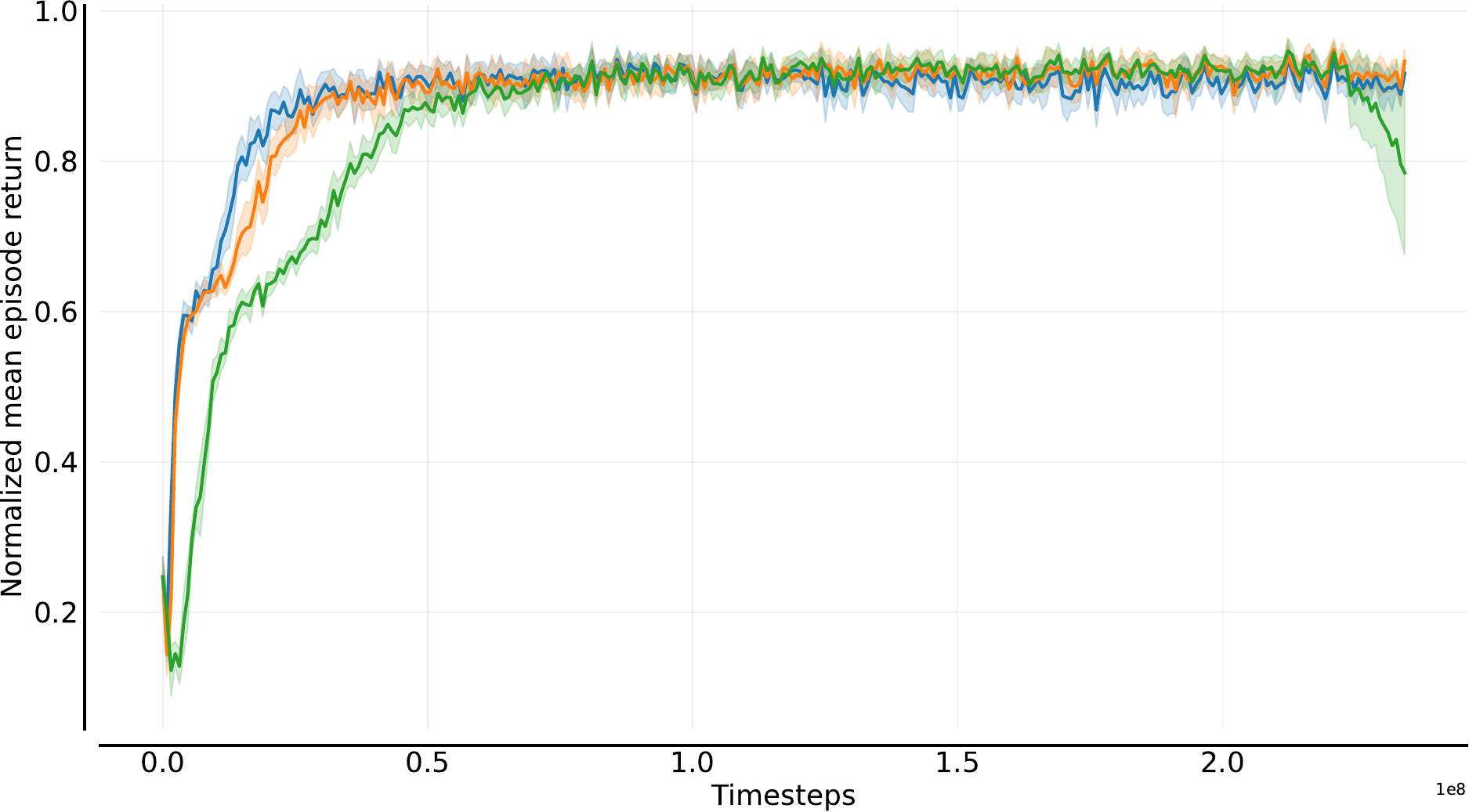}
    \caption*{\tiny JaxNav 8x8x2a}
\end{subfigure}\hfill
\begin{subfigure}{0.2\textwidth}
    \includegraphics[width=\linewidth]{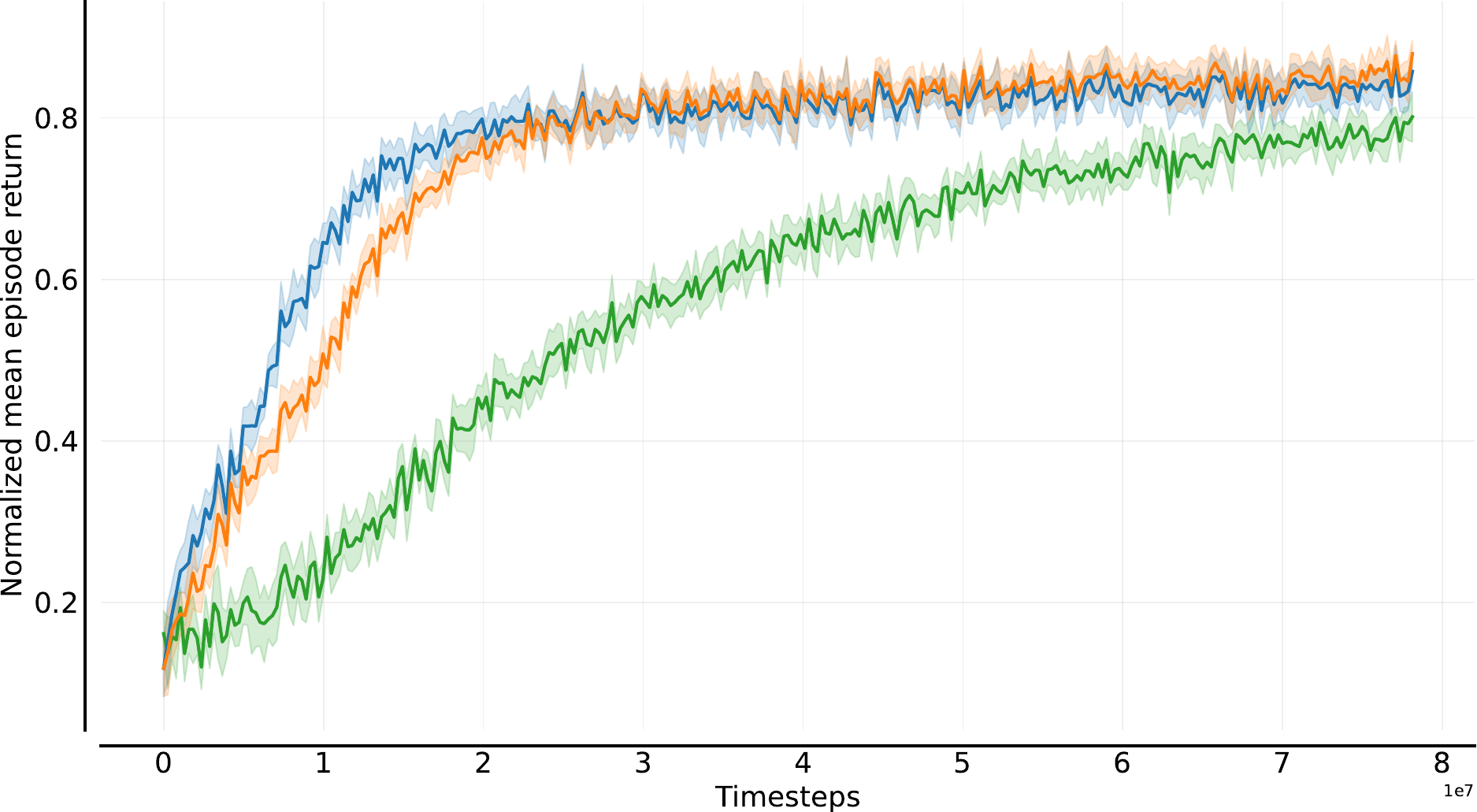}
    \caption*{\tiny MPE Simple Spread 10ag}
\end{subfigure}\hfill
\begin{subfigure}{0.2\textwidth}
    \includegraphics[width=\linewidth]{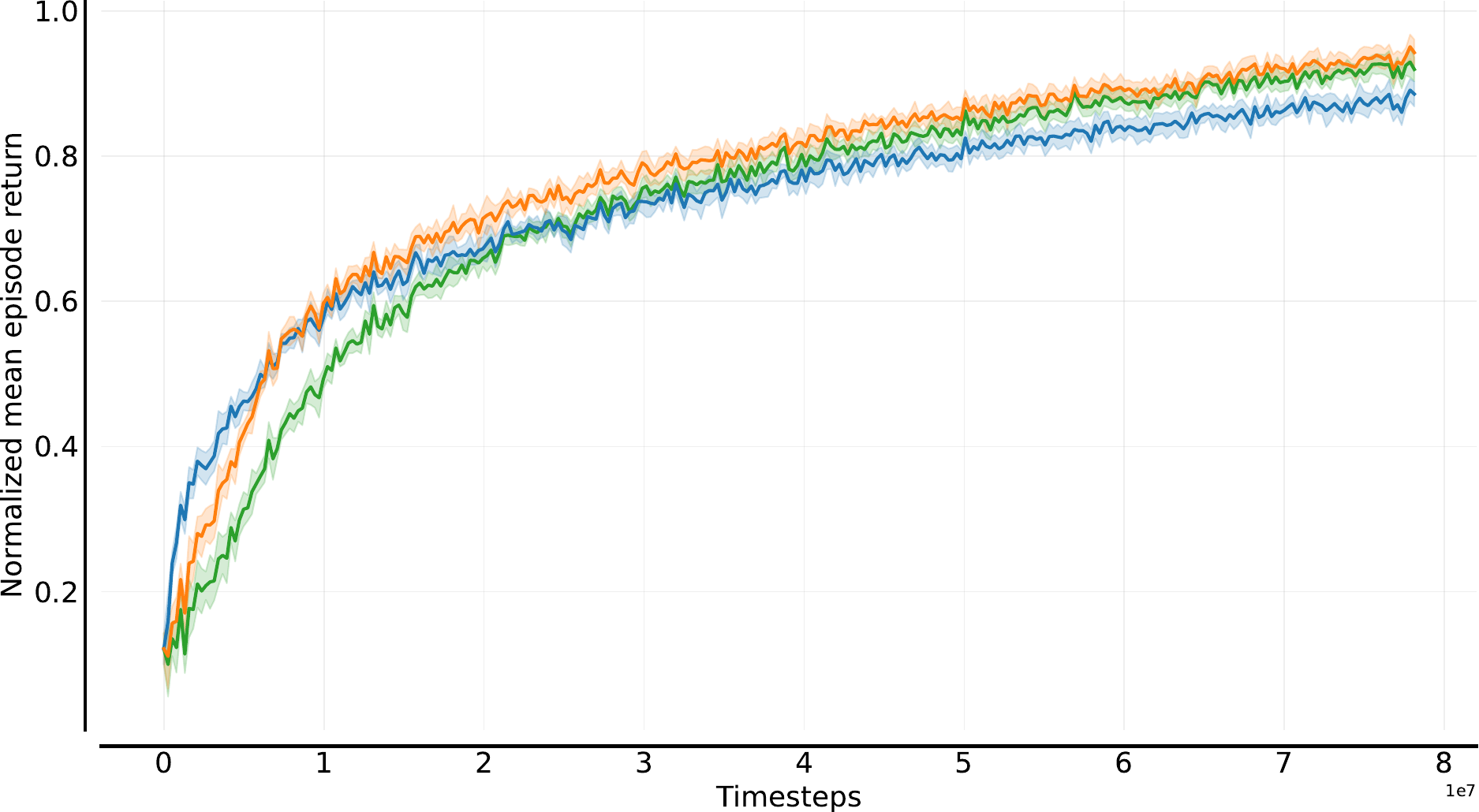}
    \caption*{\tiny MPE Simple Spread 5ag}
\end{subfigure}\hfill
\begin{subfigure}{0.2\textwidth}
    \includegraphics[width=\linewidth]{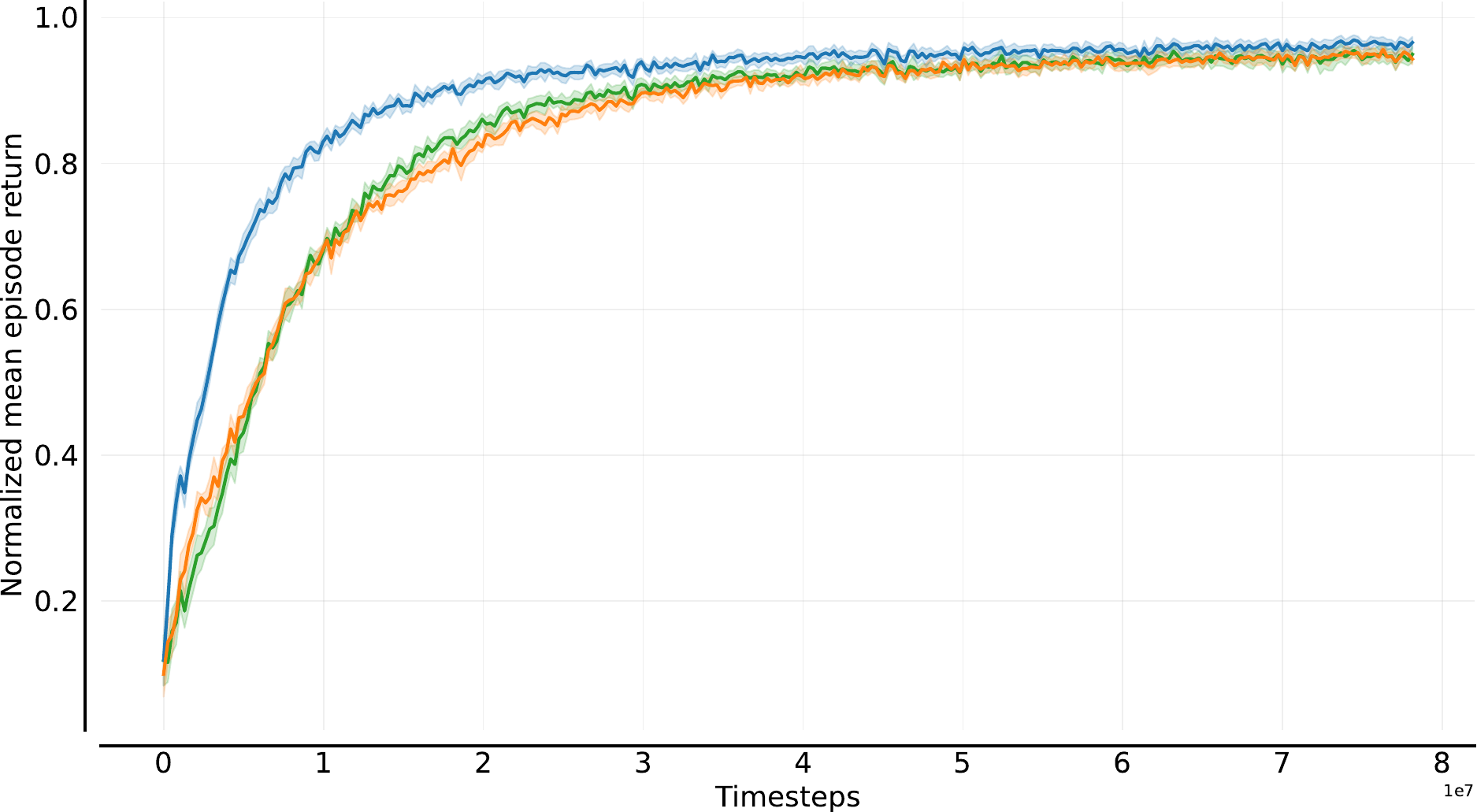}
    \caption*{\tiny MPE Simple Spread 3ag}
\end{subfigure}\hfill
\begin{subfigure}{0.2\textwidth}
    \includegraphics[width=\linewidth]{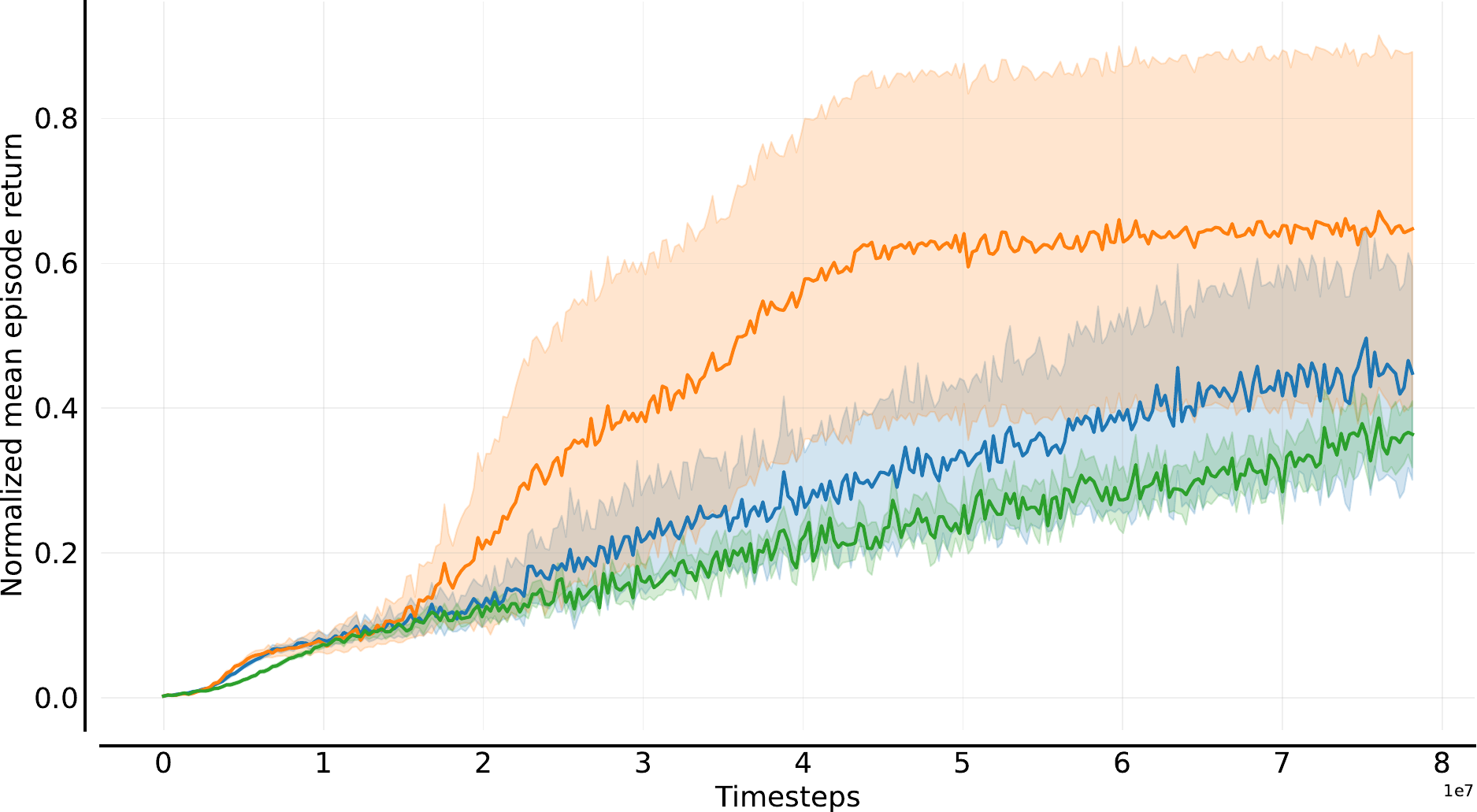}
    \caption*{\tiny PaxMen 4a}
\end{subfigure}\hfill
\par\smallskip 

\begin{subfigure}{0.2\textwidth}
    \includegraphics[width=\linewidth]{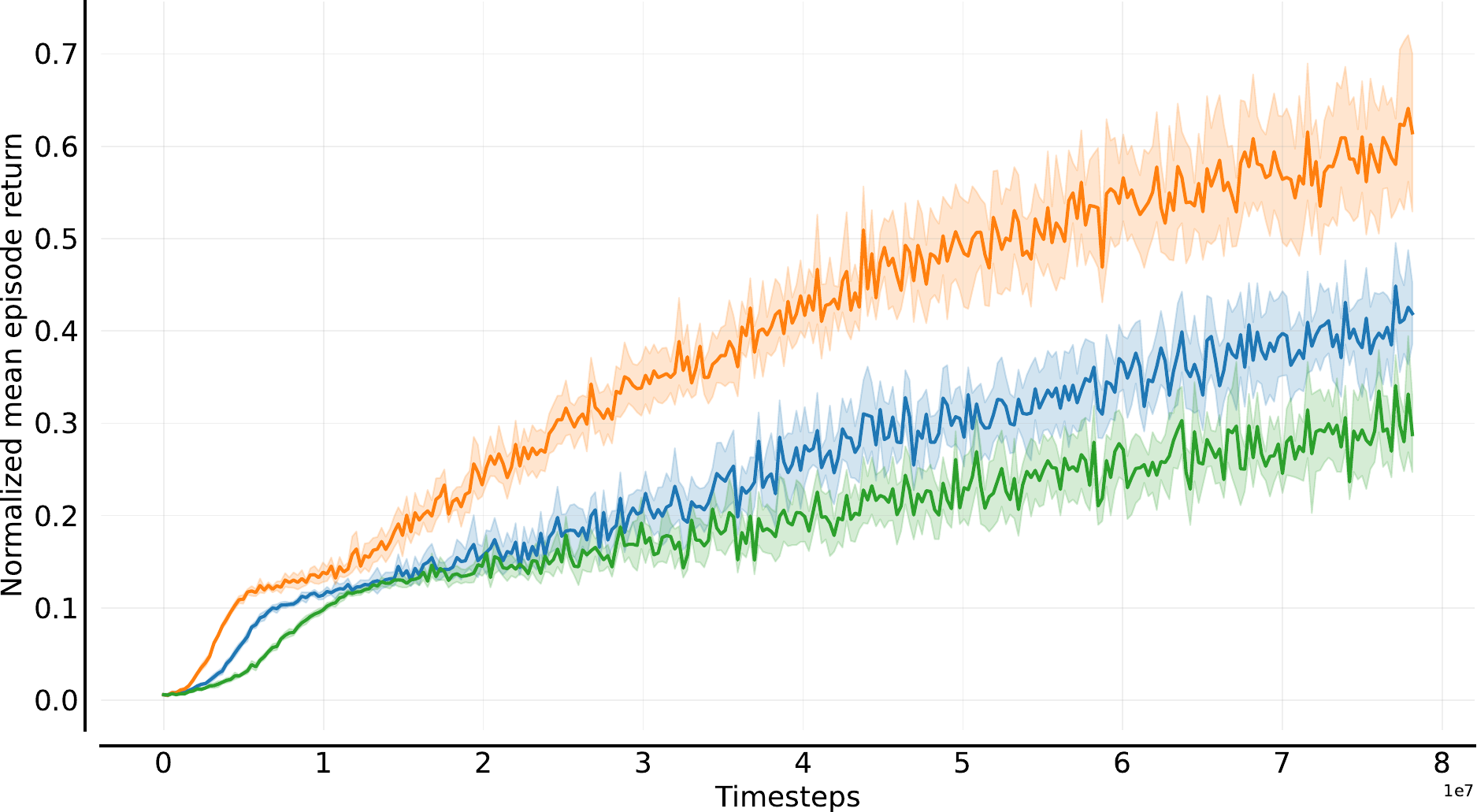}
    \caption*{\tiny PaxMen 5a}
\end{subfigure}\hfill
\begin{subfigure}{0.2\textwidth}
    \includegraphics[width=\linewidth]{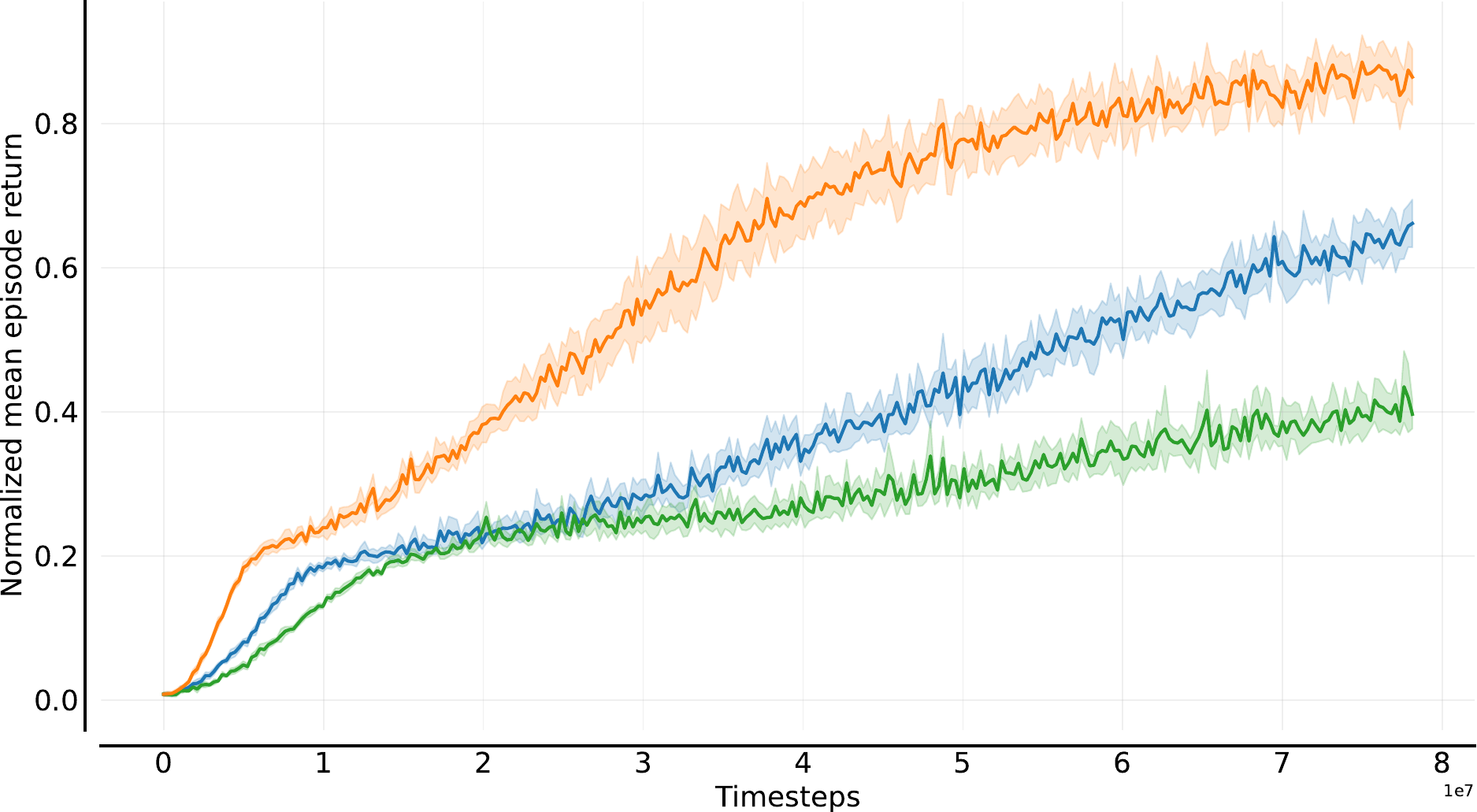}
    \caption*{\tiny PaxMen 6a}
\end{subfigure}\hfill
\begin{subfigure}{0.2\textwidth}
    \includegraphics[width=\linewidth]{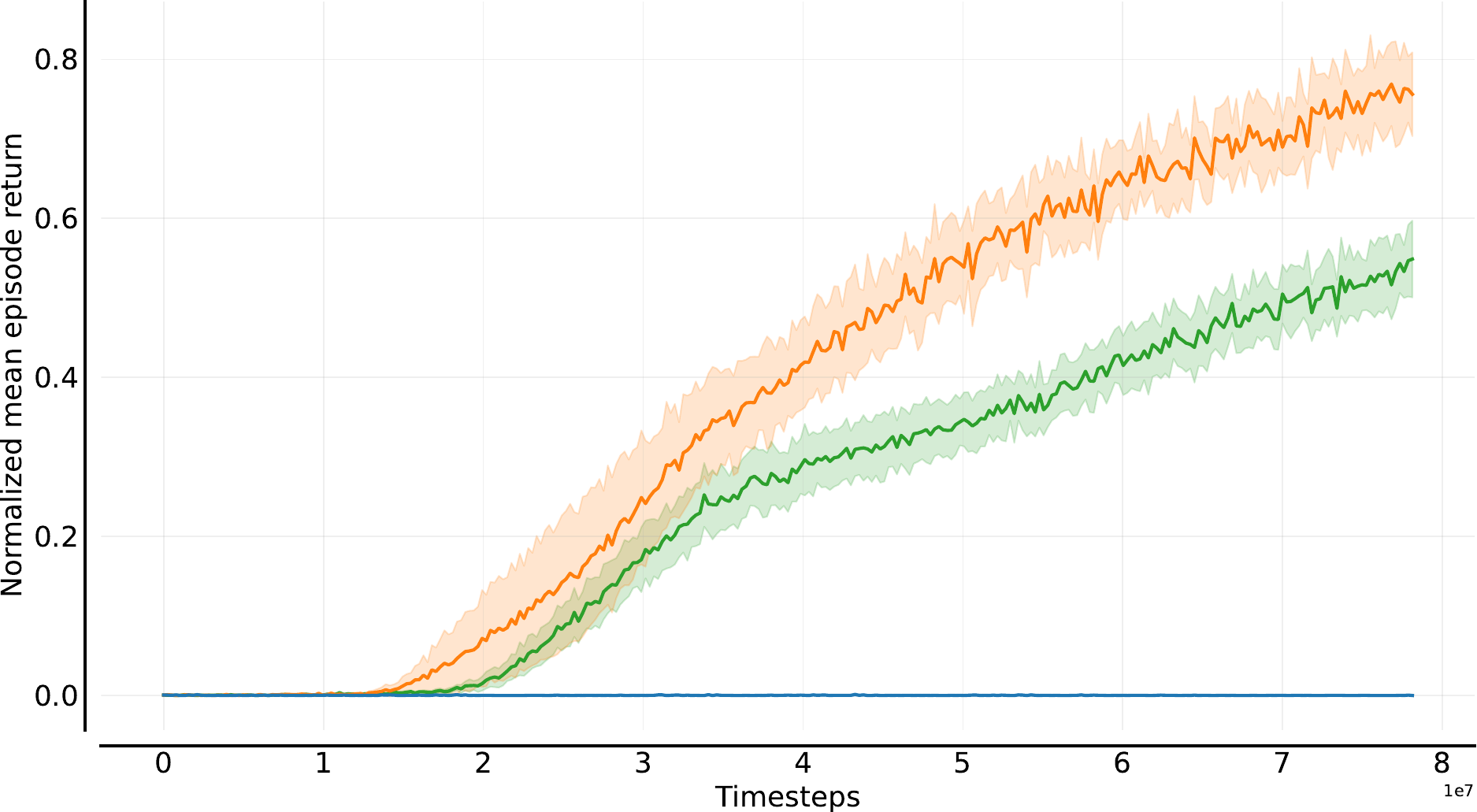}
    \caption*{\tiny RWARE Large 4ag}
\end{subfigure}\hfill
\begin{subfigure}{0.2\textwidth}
    \includegraphics[width=\linewidth]{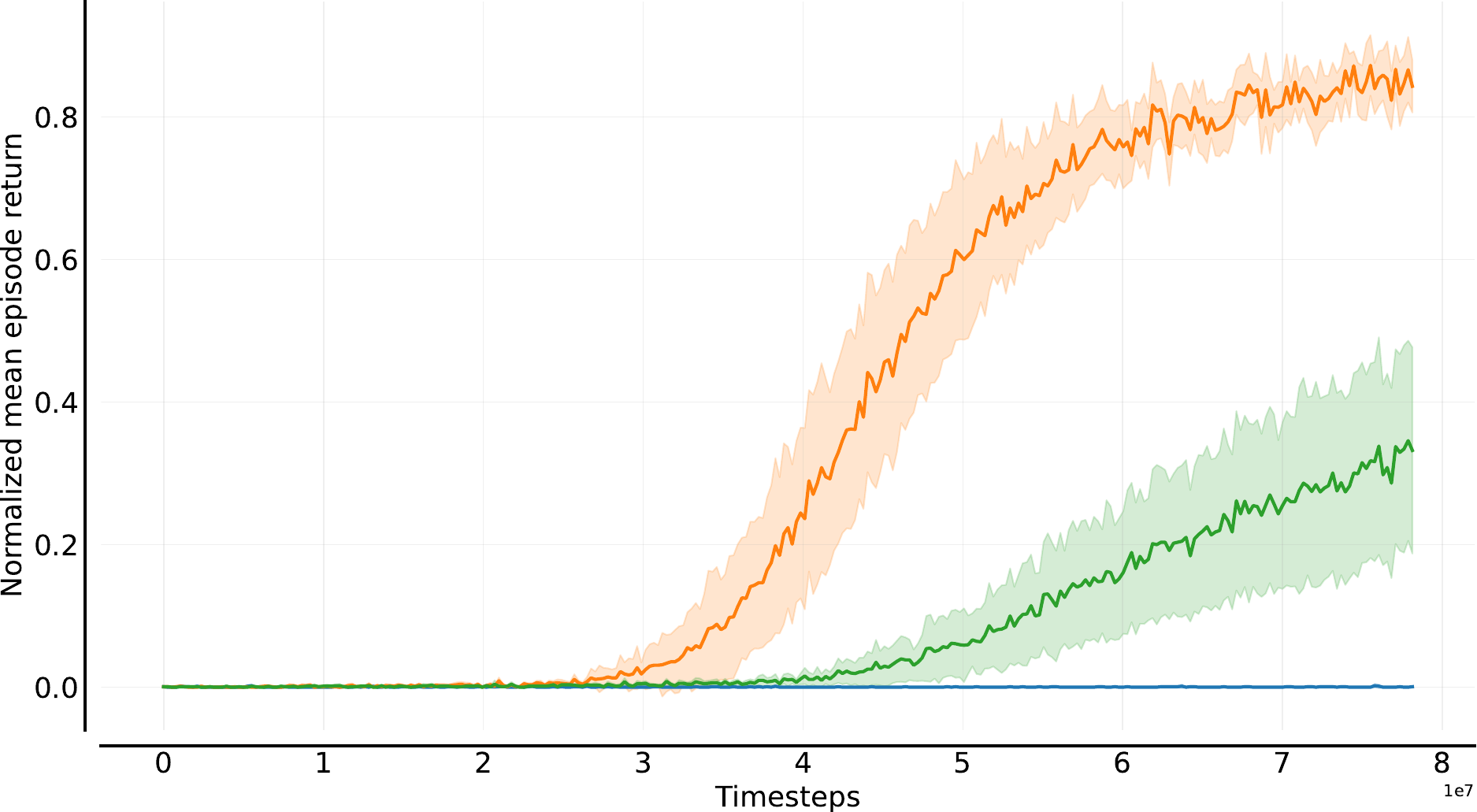}
    \caption*{\tiny RWARE Large 4ag Hard}
\end{subfigure}\hfill
\begin{subfigure}{0.2\textwidth}
    \includegraphics[width=\linewidth]{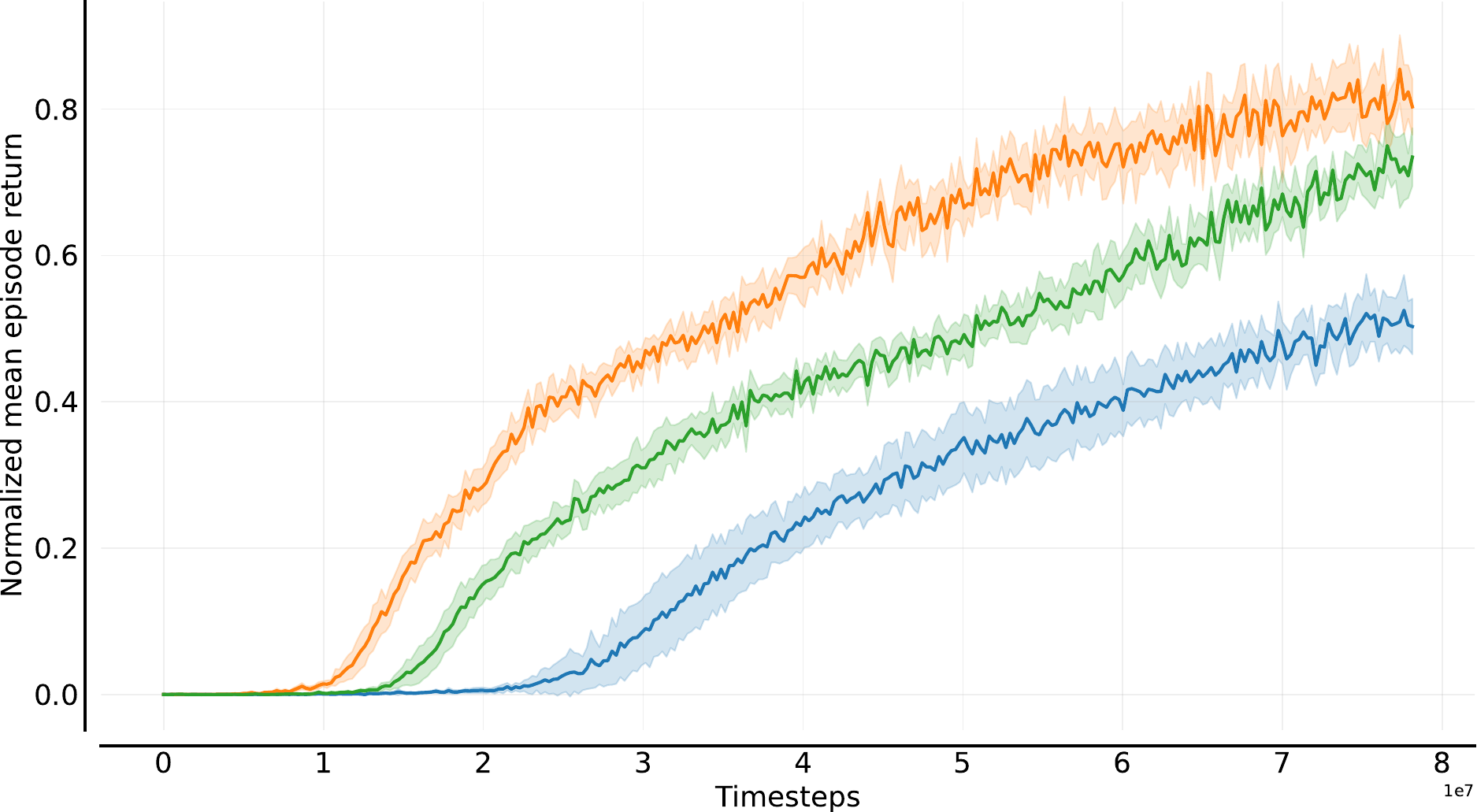}
    \caption*{\tiny RWARE Large 8ag}
\end{subfigure}\hfill
\par\smallskip 

\begin{subfigure}{0.2\textwidth}
    \includegraphics[width=\linewidth]{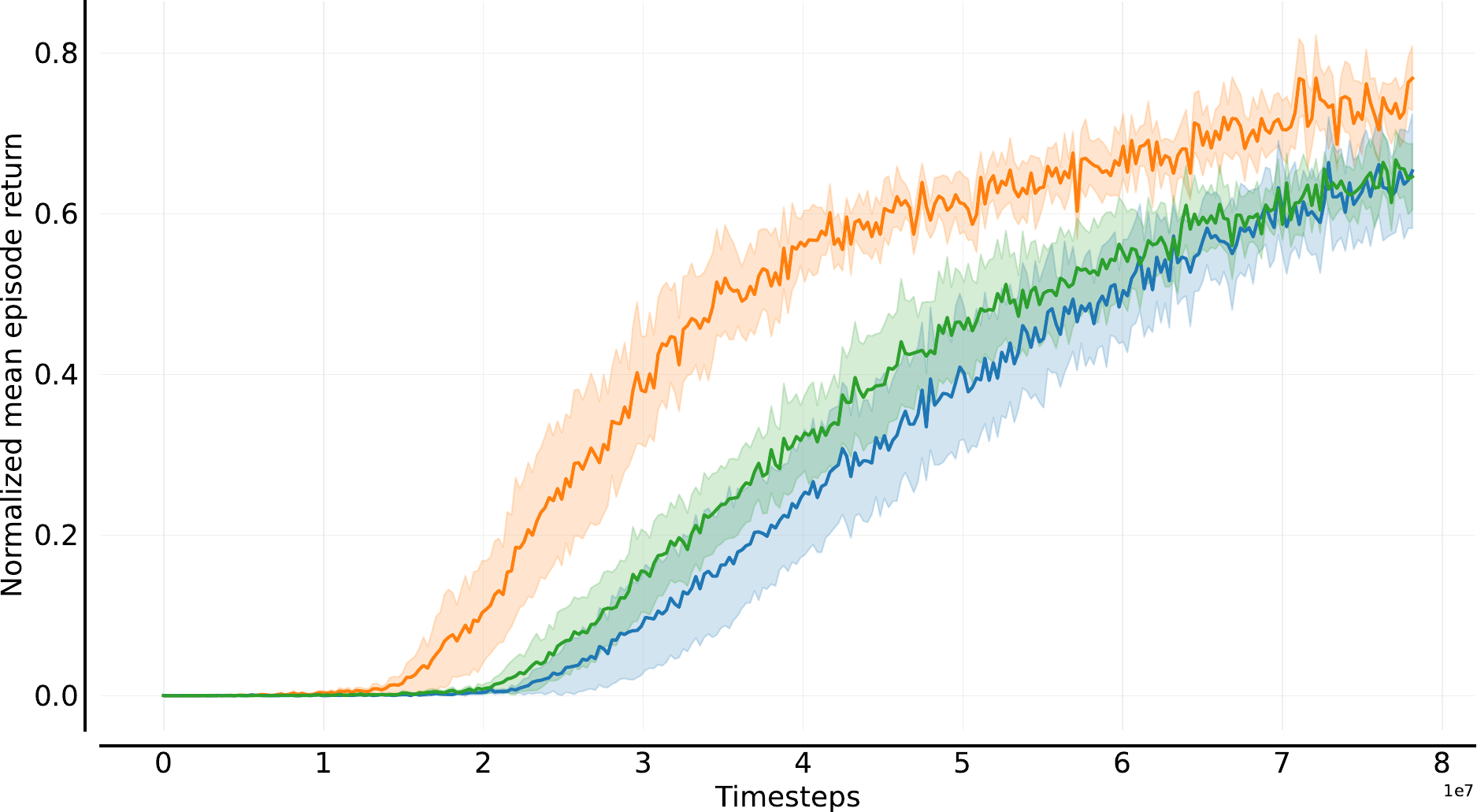}
    \caption*{\tiny RWARE Large 8ag Hard}
\end{subfigure}\hfill
\begin{subfigure}{0.2\textwidth}
    \includegraphics[width=\linewidth]{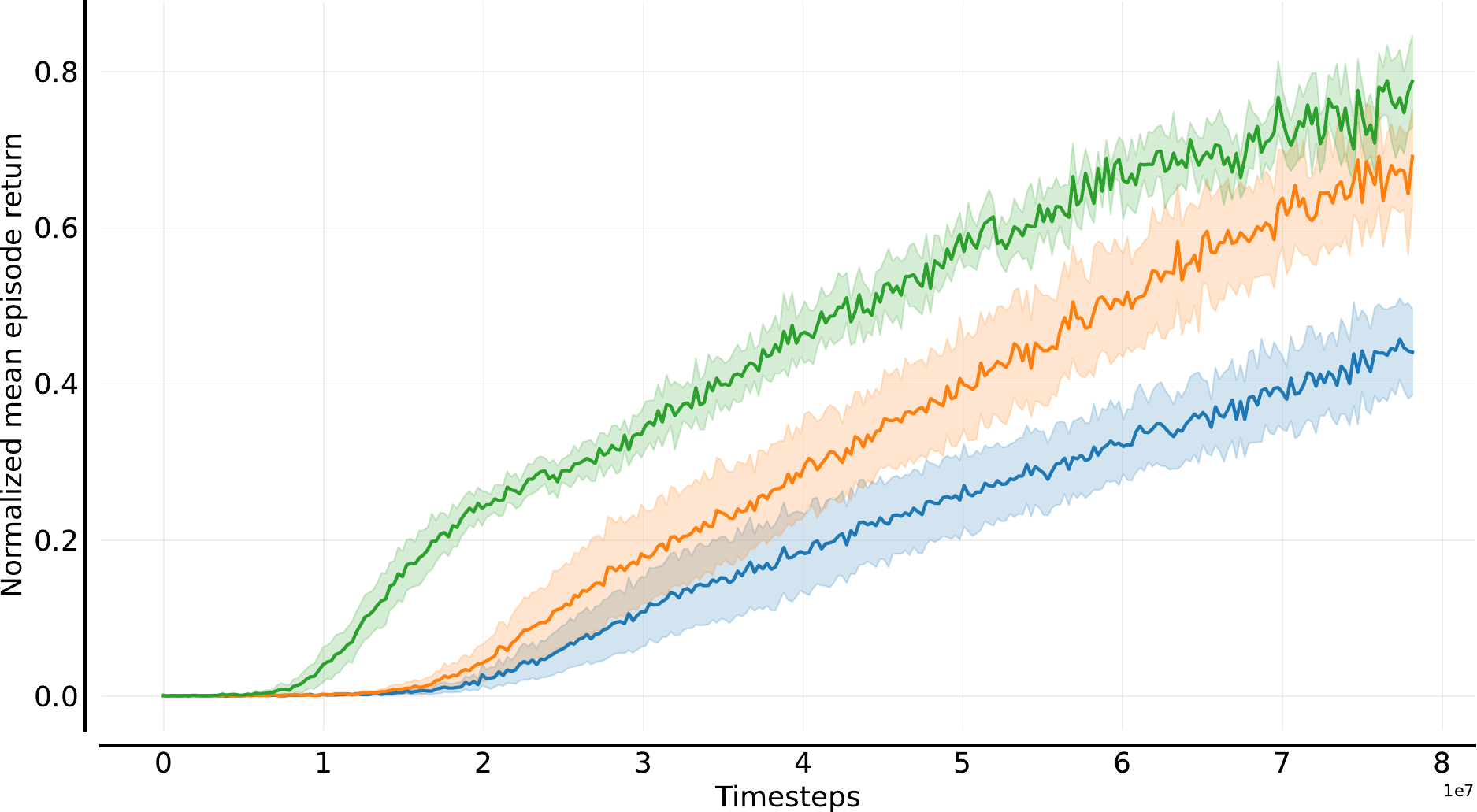}
    \caption*{\tiny RWARE Medium 4ag}
\end{subfigure}\hfill
\begin{subfigure}{0.2\textwidth}
    \includegraphics[width=\linewidth]{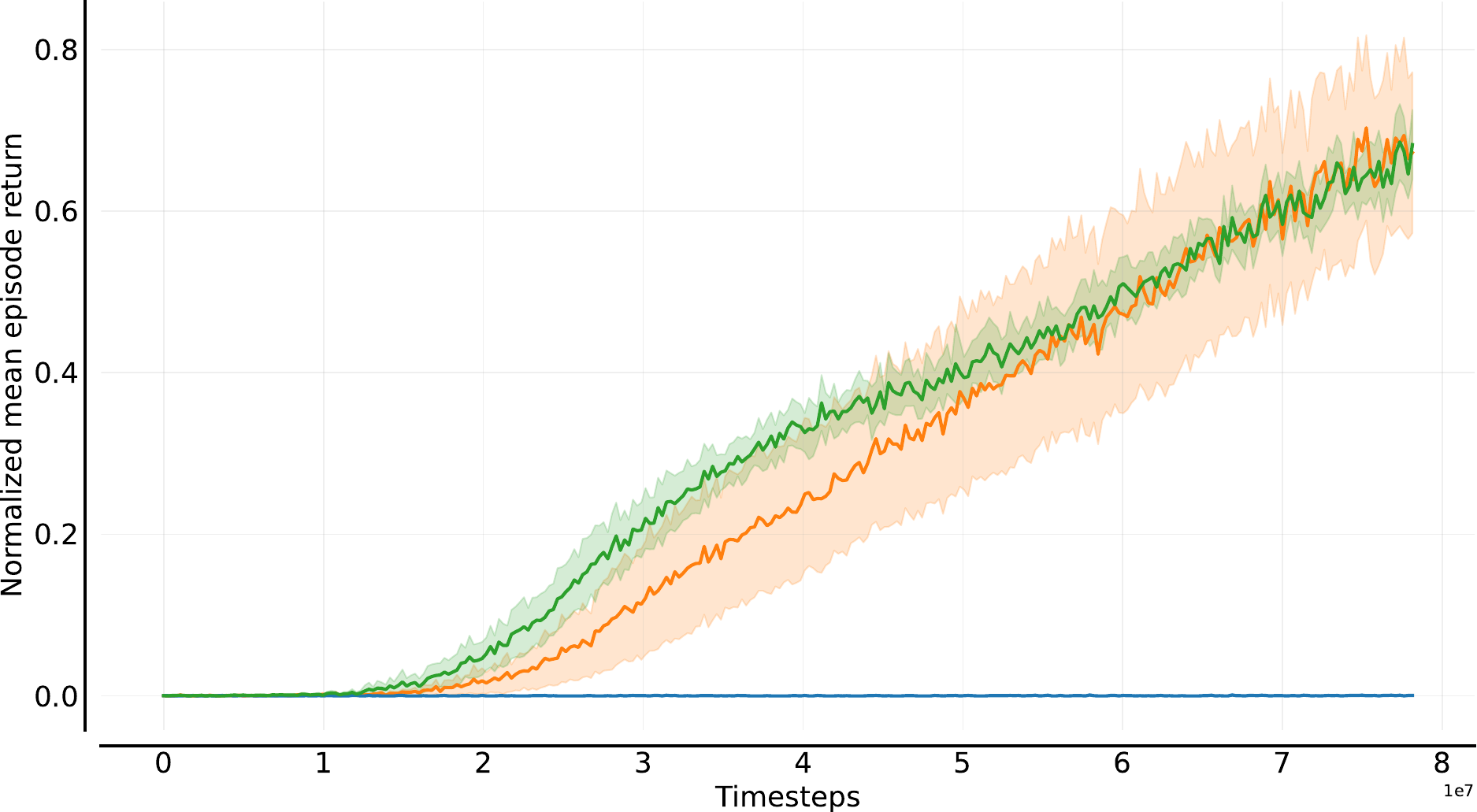}
    \caption*{\tiny RWARE Medium 4ag Hard}
\end{subfigure}\hfill
\begin{subfigure}{0.2\textwidth}
    \includegraphics[width=\linewidth]{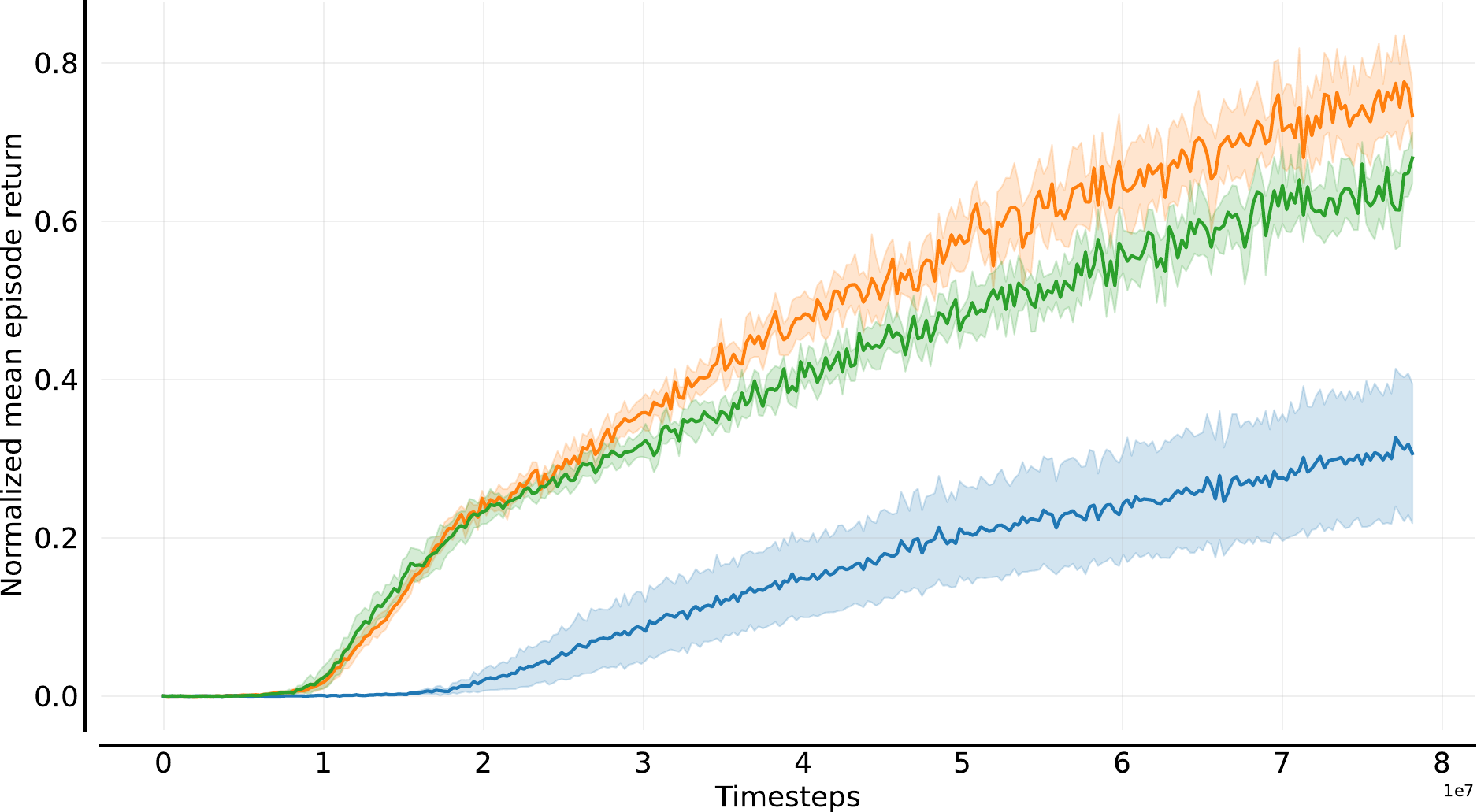}
    \caption*{\tiny RWARE Medium 6ag}
\end{subfigure}\hfill
\begin{subfigure}{0.2\textwidth}
    \includegraphics[width=\linewidth]{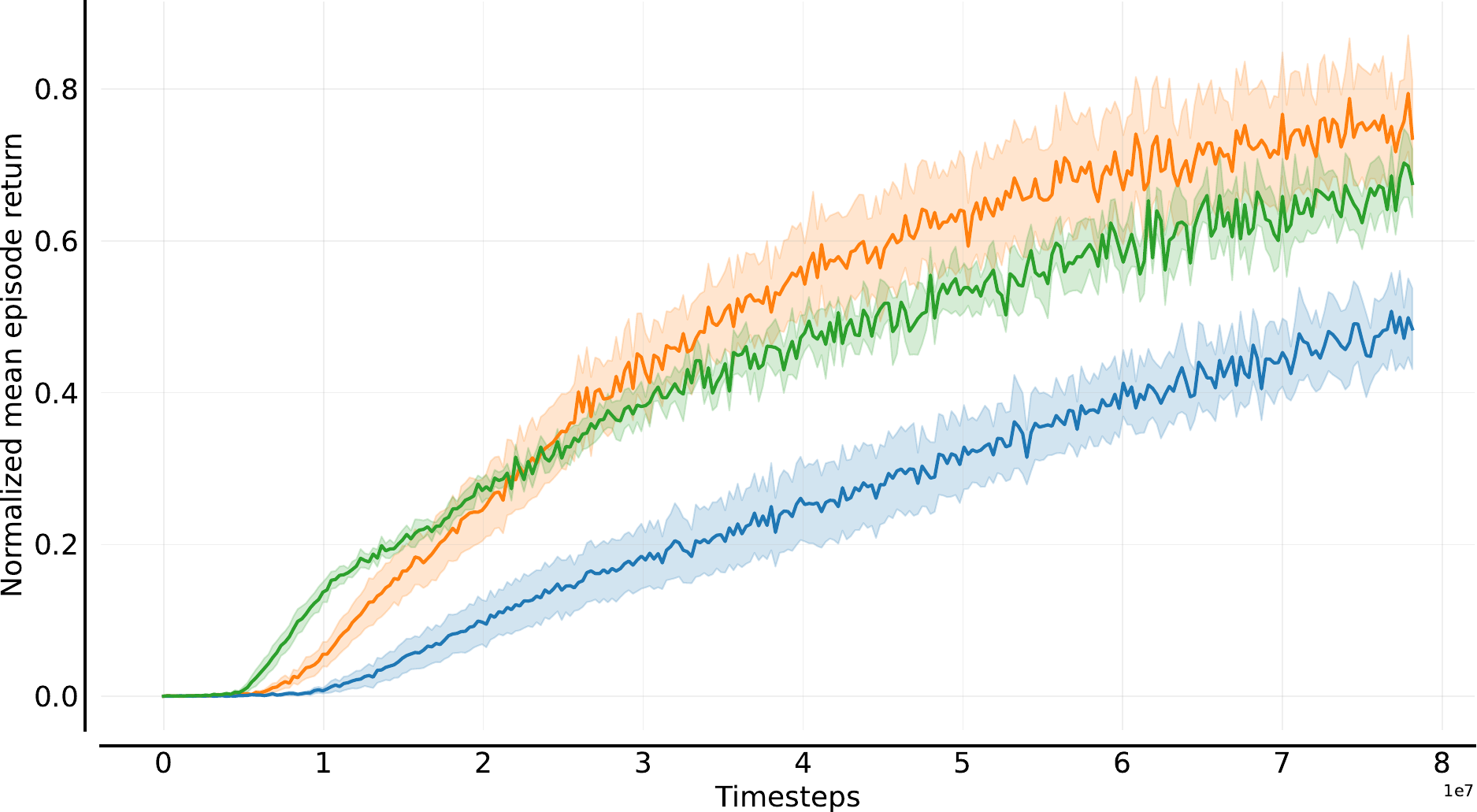}
    \caption*{\tiny RWARE Small 4ag}
\end{subfigure}\hfill
\par\smallskip 

\begin{subfigure}{0.2\textwidth}
    \includegraphics[width=\linewidth]{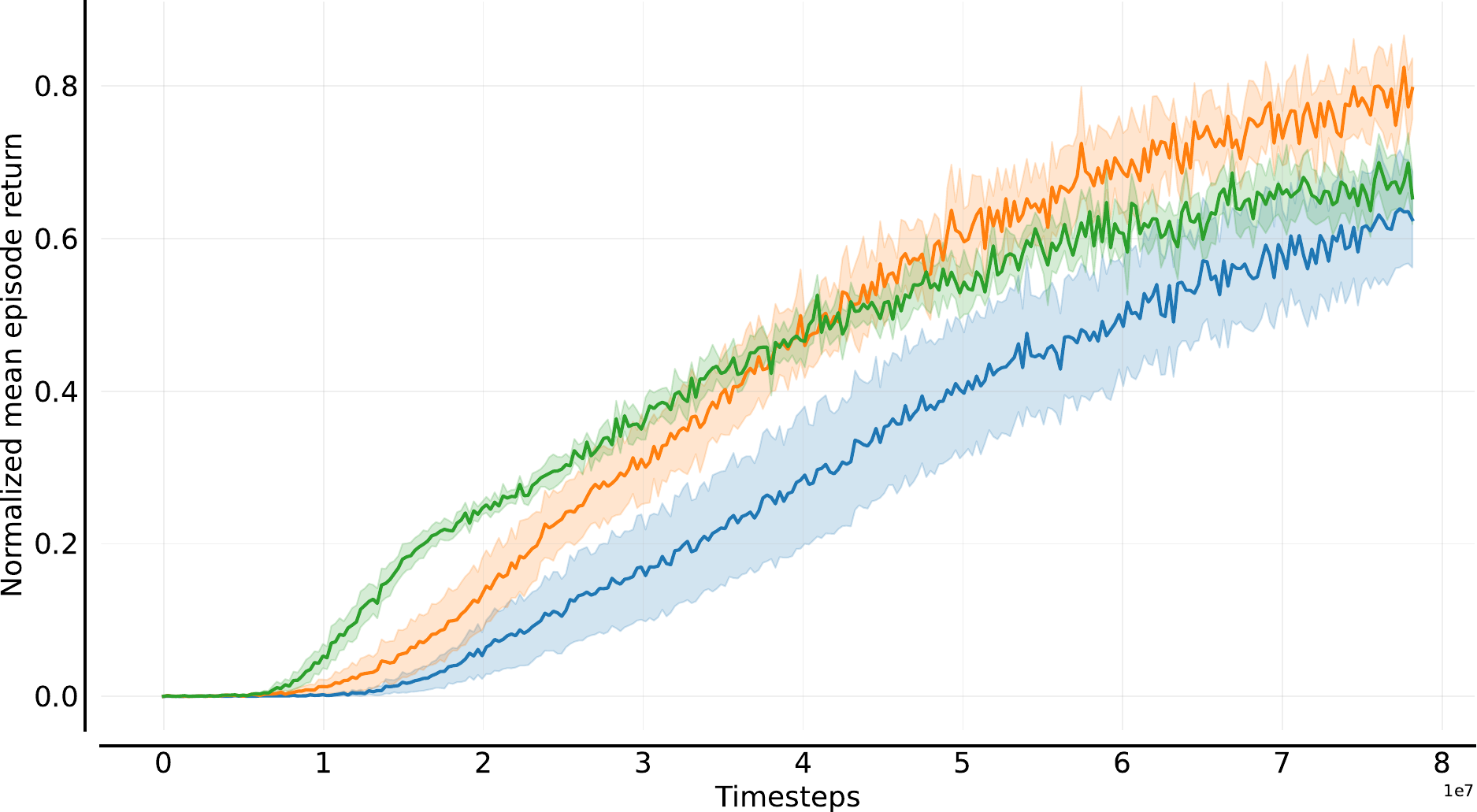}
    \caption*{\tiny RWARE Small 4ag Hard}
\end{subfigure}\hfill
\begin{subfigure}{0.2\textwidth}
    \includegraphics[width=\linewidth]{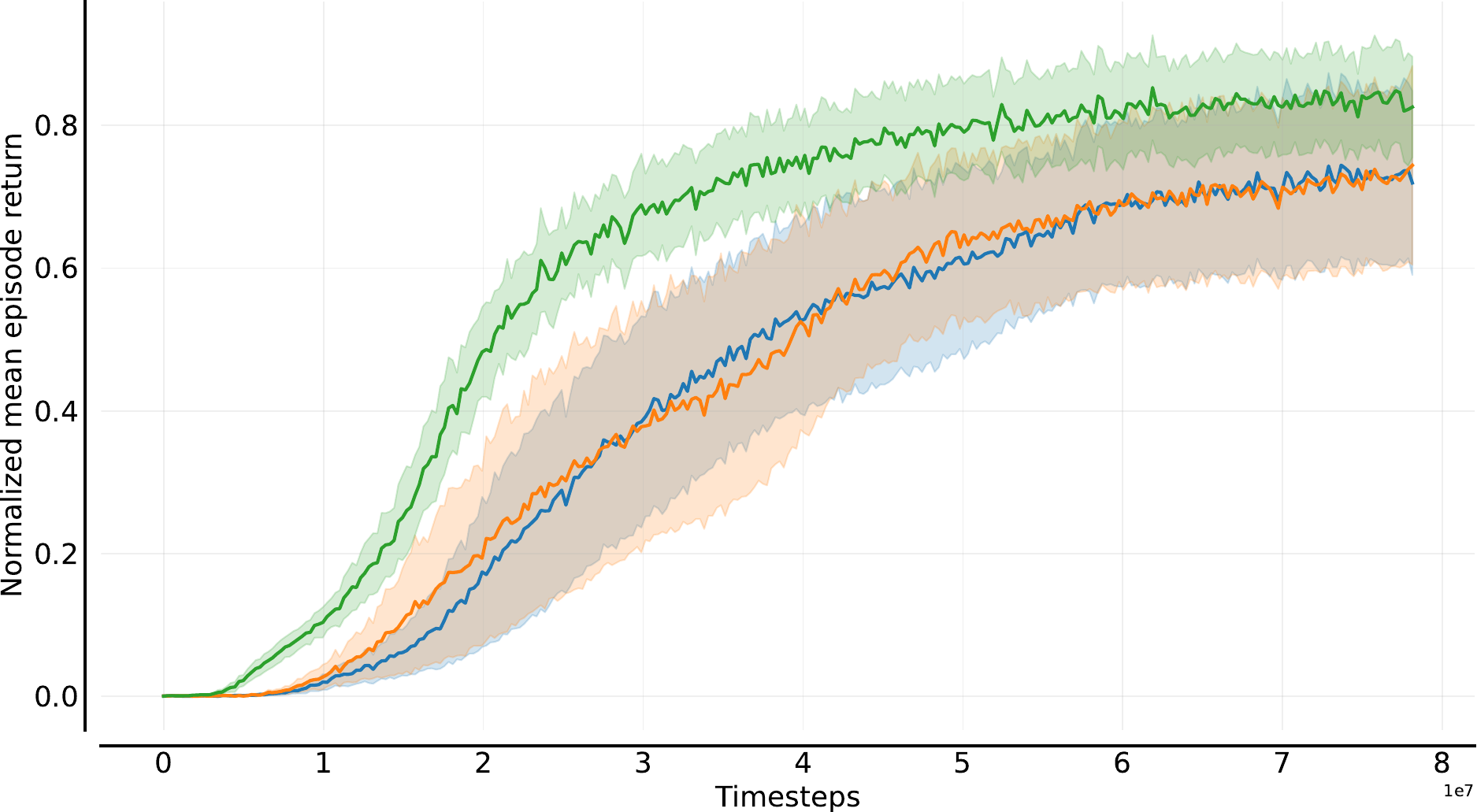}
    \caption*{\tiny RWARE Tiny 2ag}
\end{subfigure}\hfill
\begin{subfigure}{0.2\textwidth}
    \includegraphics[width=\linewidth]{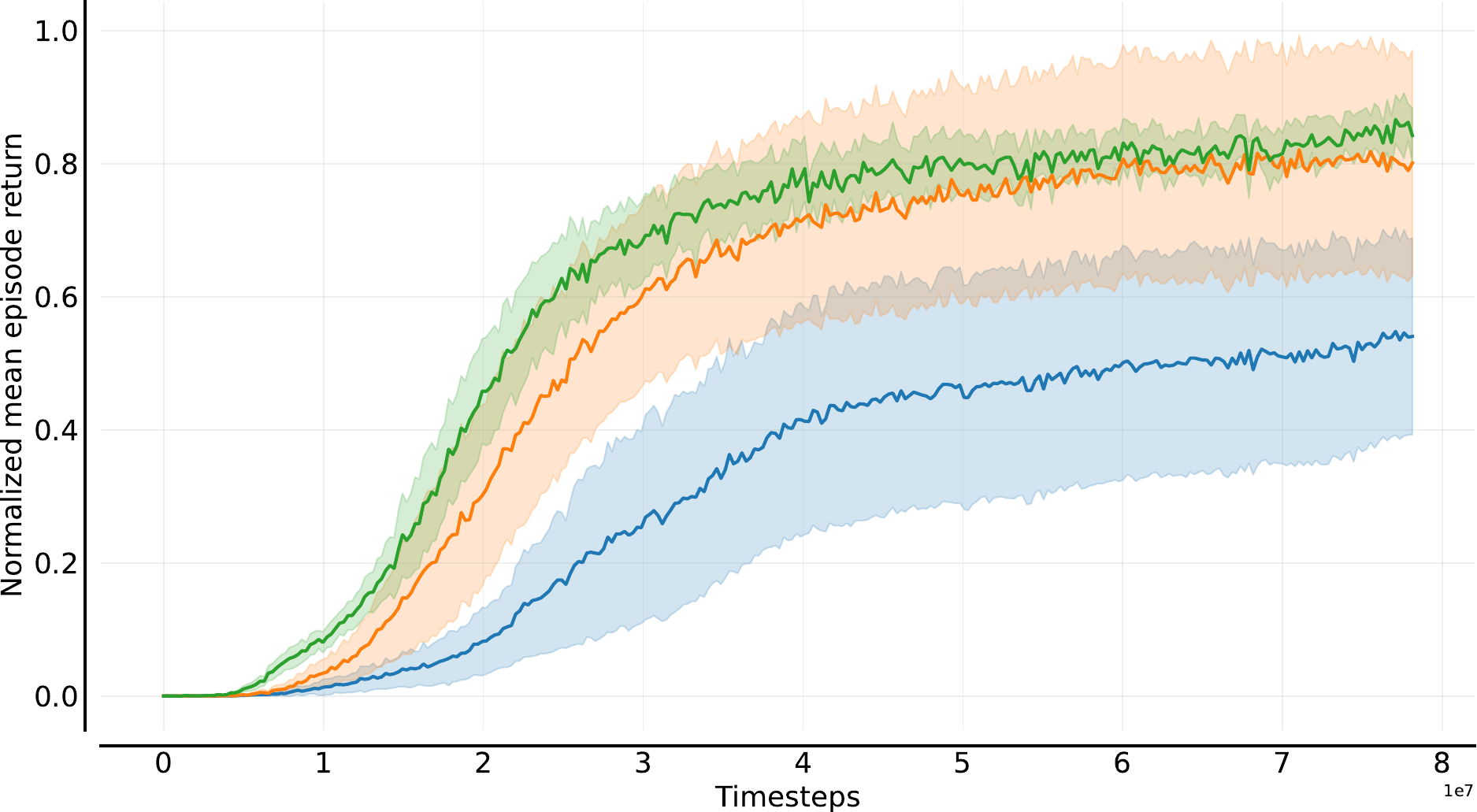}
    \caption*{\tiny RWARE Tiny 2ag Hard}
\end{subfigure}\hfill
\begin{subfigure}{0.2\textwidth}
    \includegraphics[width=\linewidth]{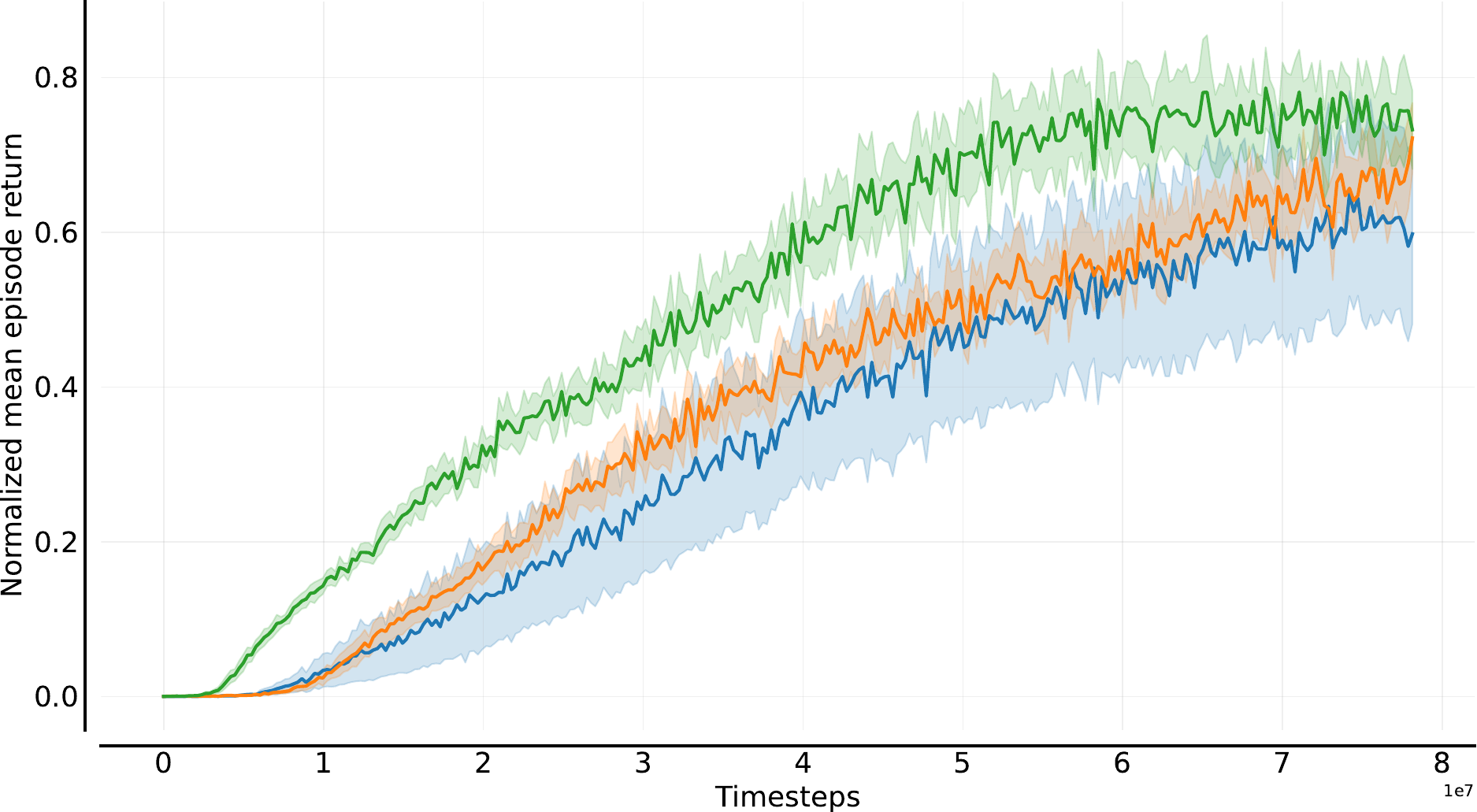}
    \caption*{\tiny RWARE Tiny 4ag}
\end{subfigure}\hfill
\begin{subfigure}{0.2\textwidth}
    \includegraphics[width=\linewidth]{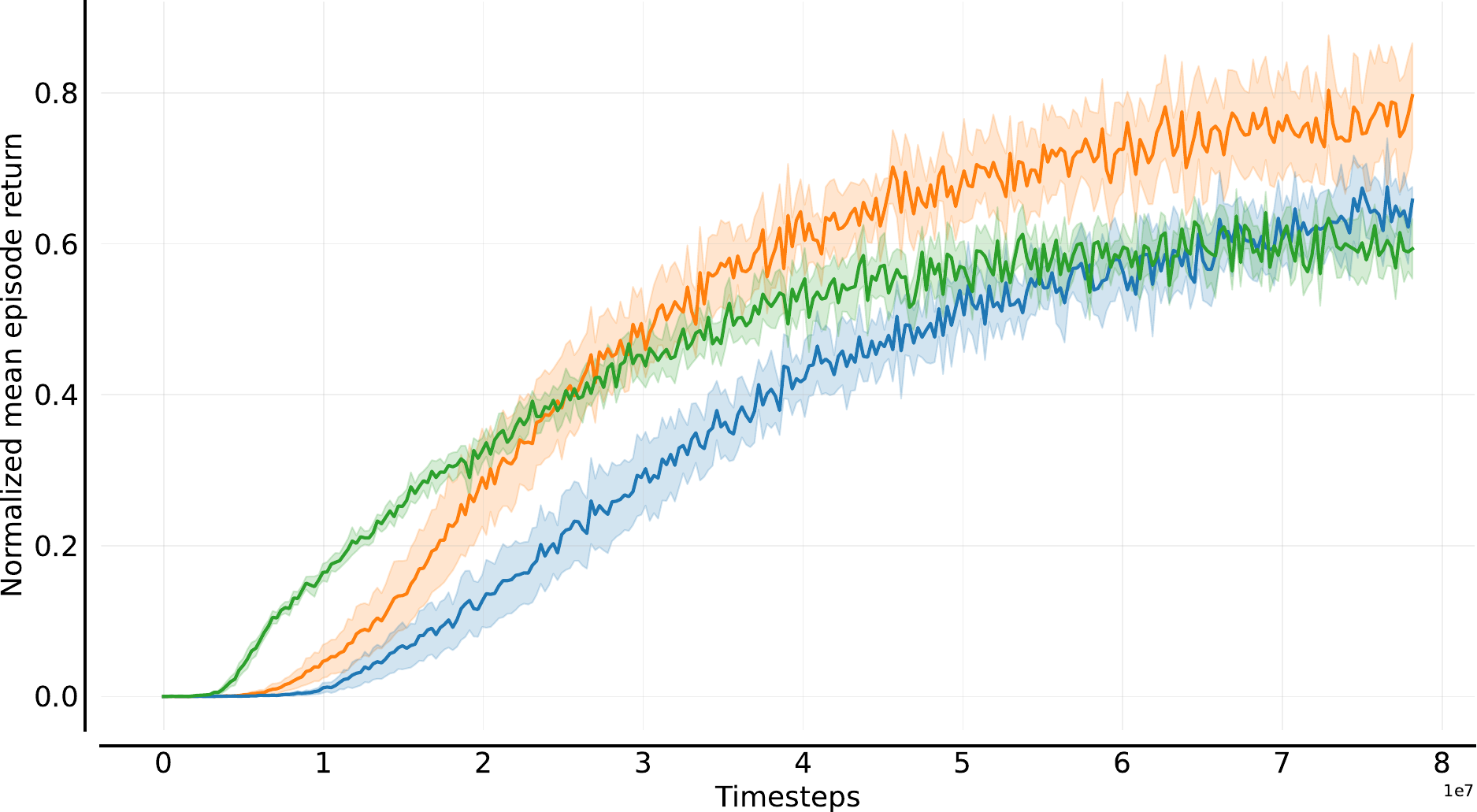}
    \caption*{\tiny RWARE Tiny 4ag Hard}
\end{subfigure}\hfill
\par\smallskip 

\begin{subfigure}{0.2\textwidth}
    \includegraphics[width=\linewidth]{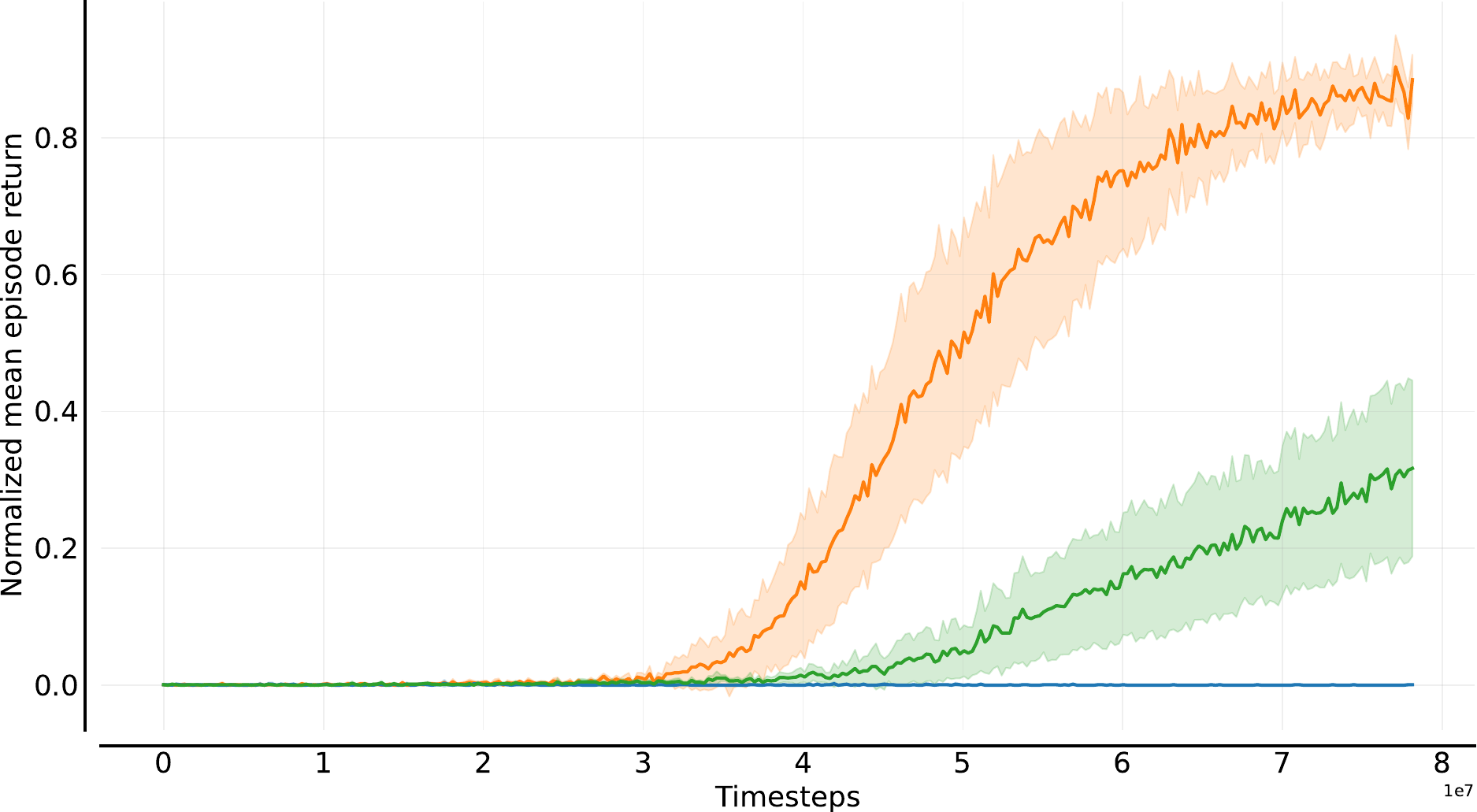}
    \caption*{\tiny RWARE XLarge 4ag}
\end{subfigure}\hfill
\begin{subfigure}{0.2\textwidth}
    \includegraphics[width=\linewidth]{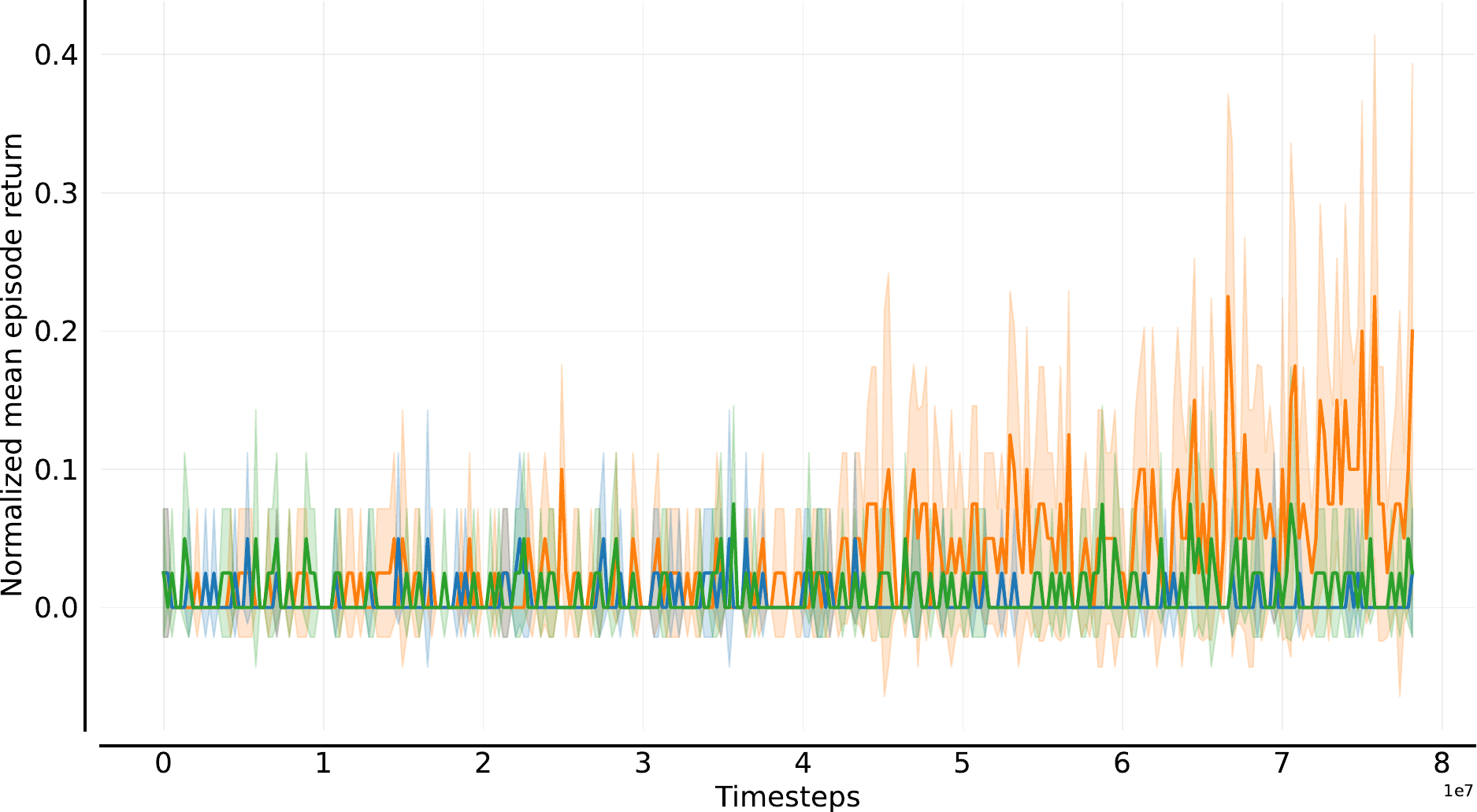}
    \caption*{\tiny RWARE XLarge 4ag Hard}
\end{subfigure}\hfill
\begin{subfigure}{0.2\textwidth}
    \includegraphics[width=\linewidth]{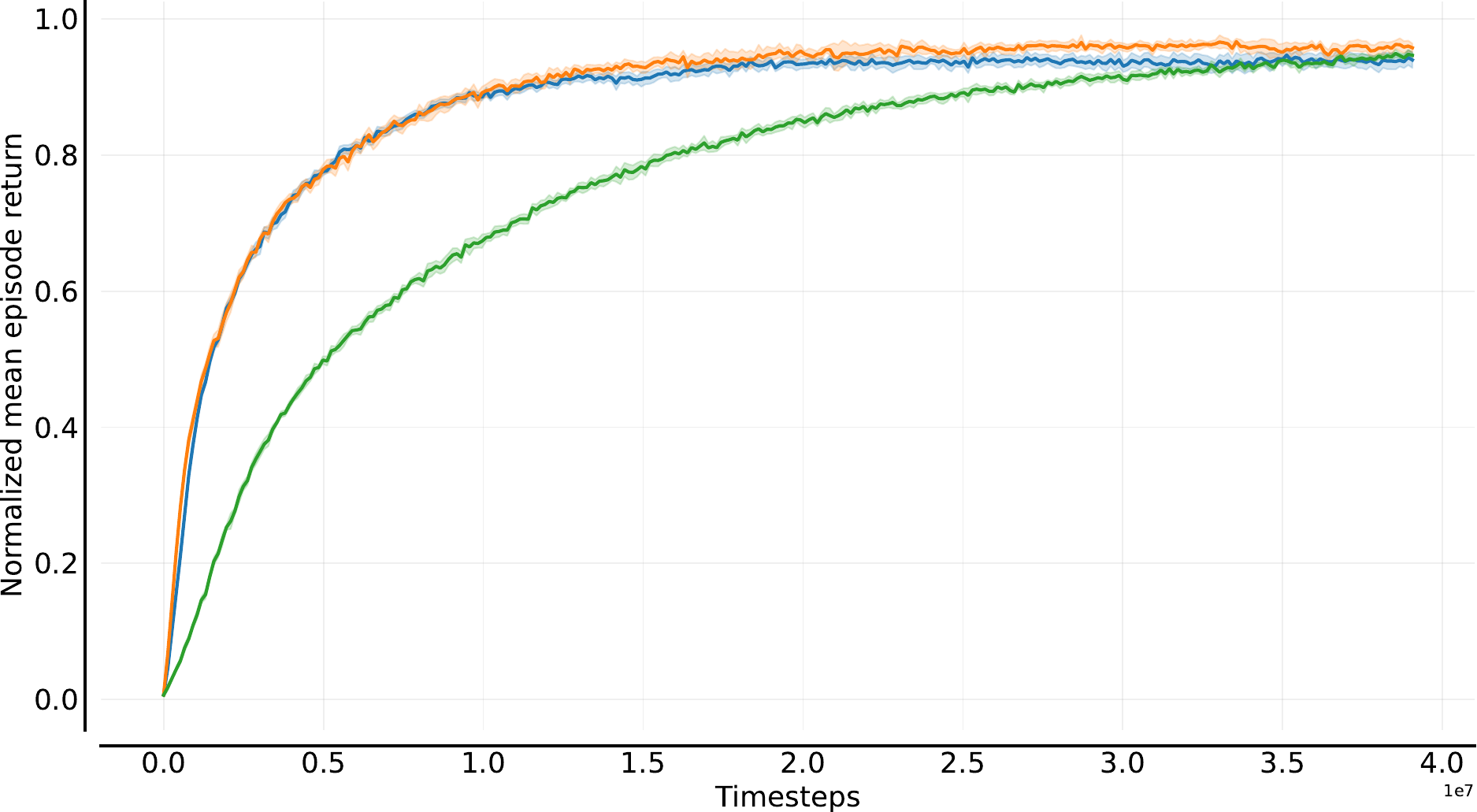}
    \caption*{\tiny Search and Rescue 100tg 2ag}
\end{subfigure}\hfill
\begin{subfigure}{0.2\textwidth}
    \includegraphics[width=\linewidth]{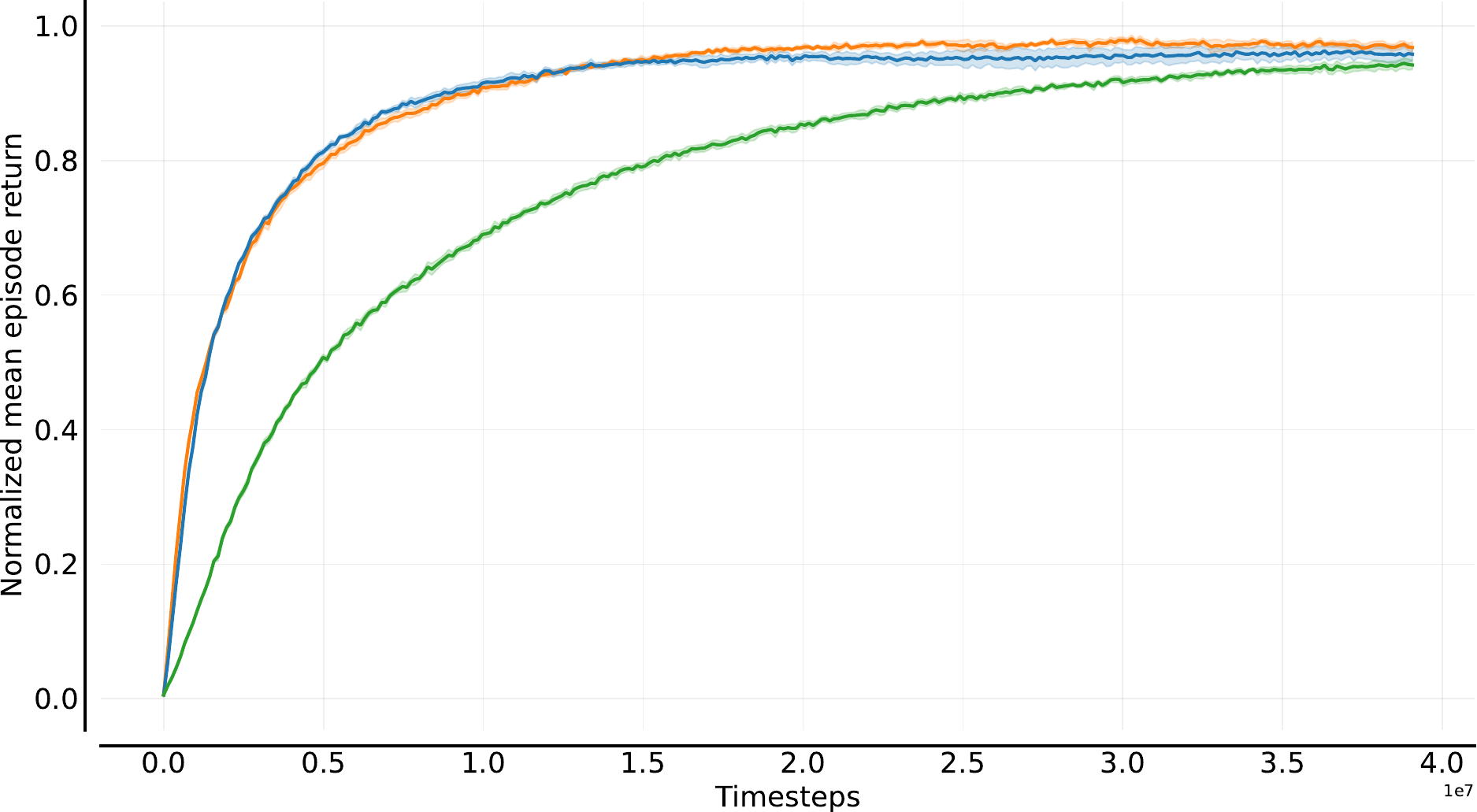}
    \caption*{\tiny Search and Rescue 200tg 4ag}
\end{subfigure}\hfill
\begin{subfigure}{0.2\textwidth}
    \includegraphics[width=\linewidth]{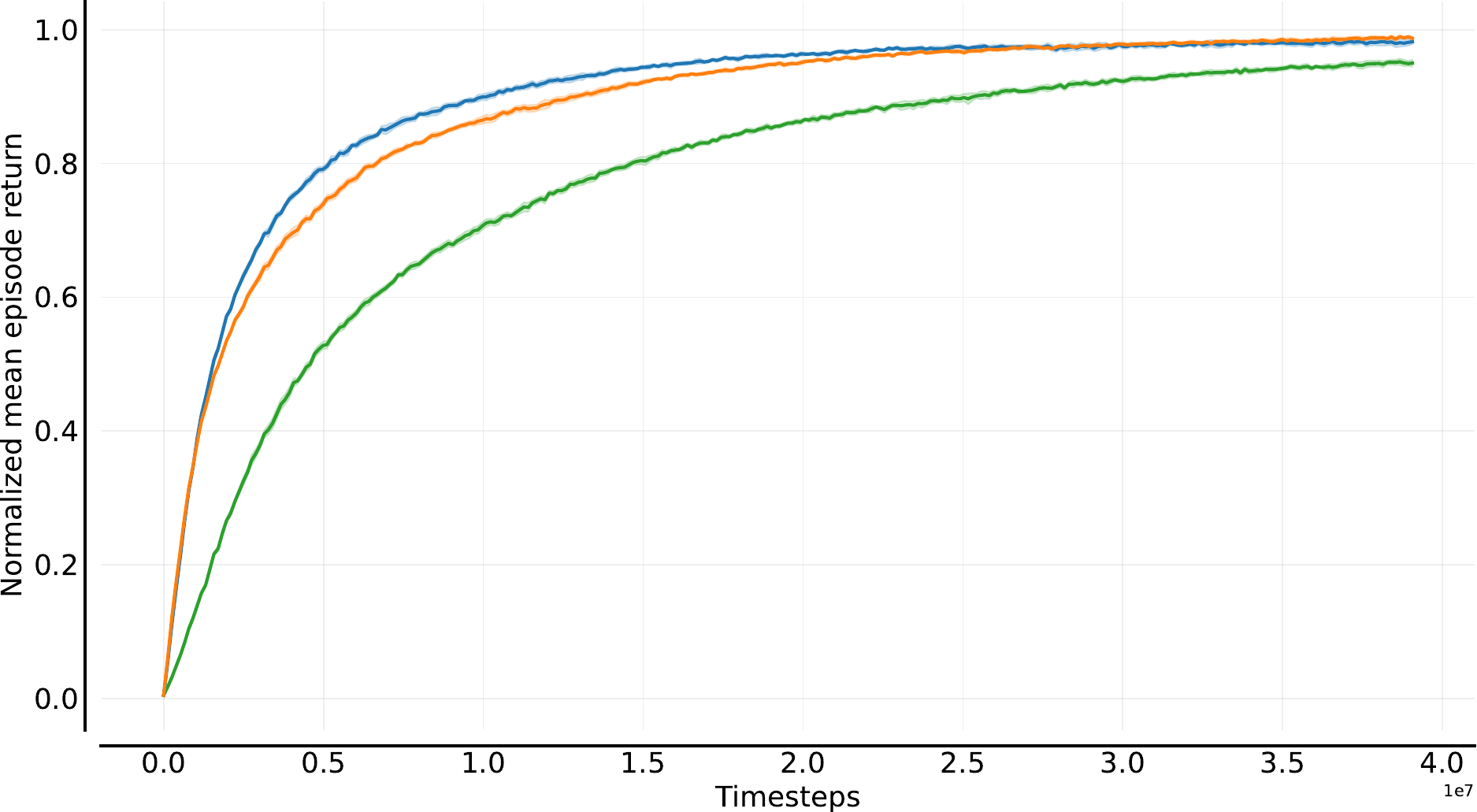}
    \caption*{\tiny Search and Rescue 300tg 6ag}
\end{subfigure}\hfill
\par\smallskip 

\begin{subfigure}{0.2\textwidth}
    \includegraphics[width=\linewidth]{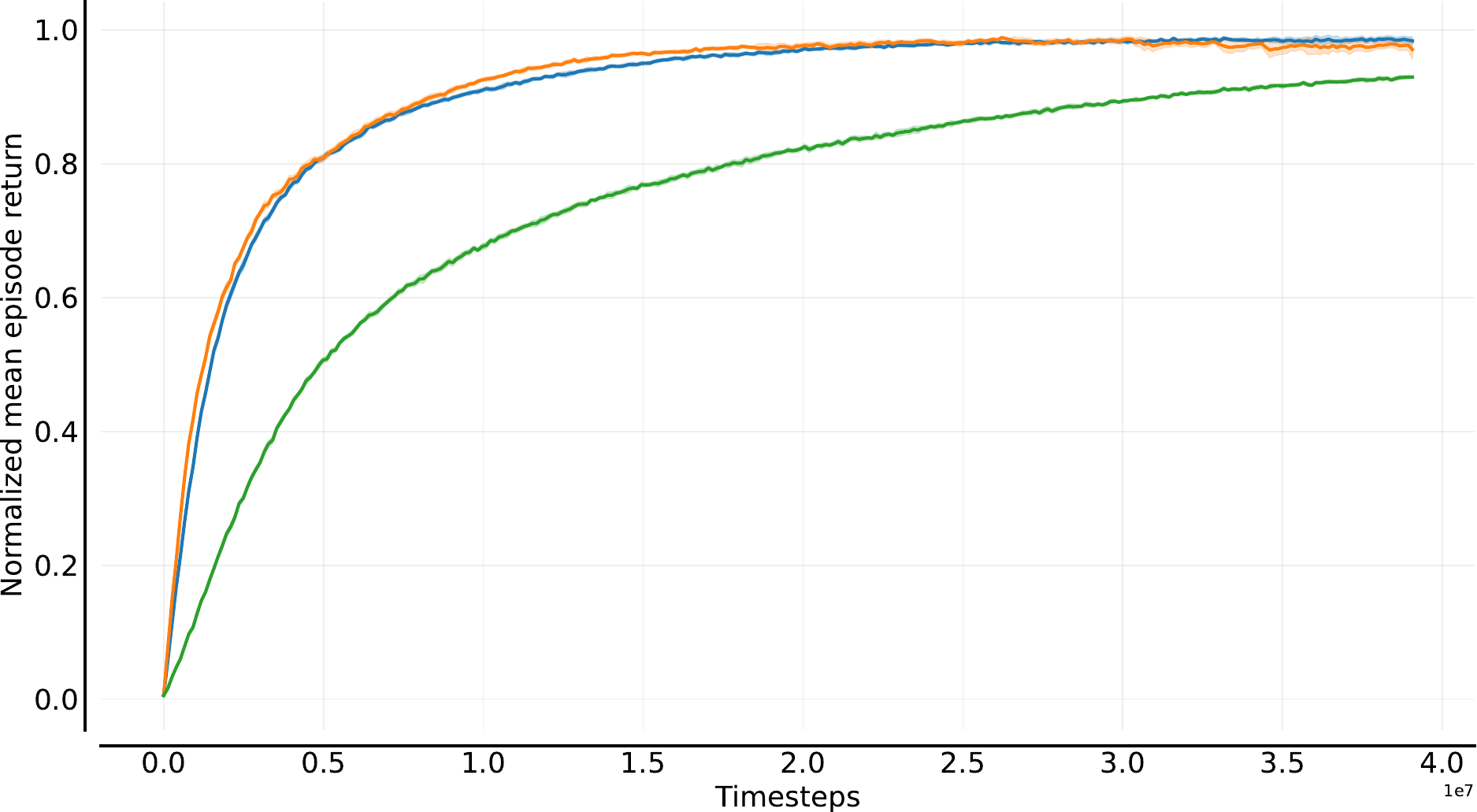}
    \caption*{\tiny Search and Rescue 400tg 8ag}
\end{subfigure}\hfill
\begin{subfigure}{0.2\textwidth}
    \includegraphics[width=\linewidth]{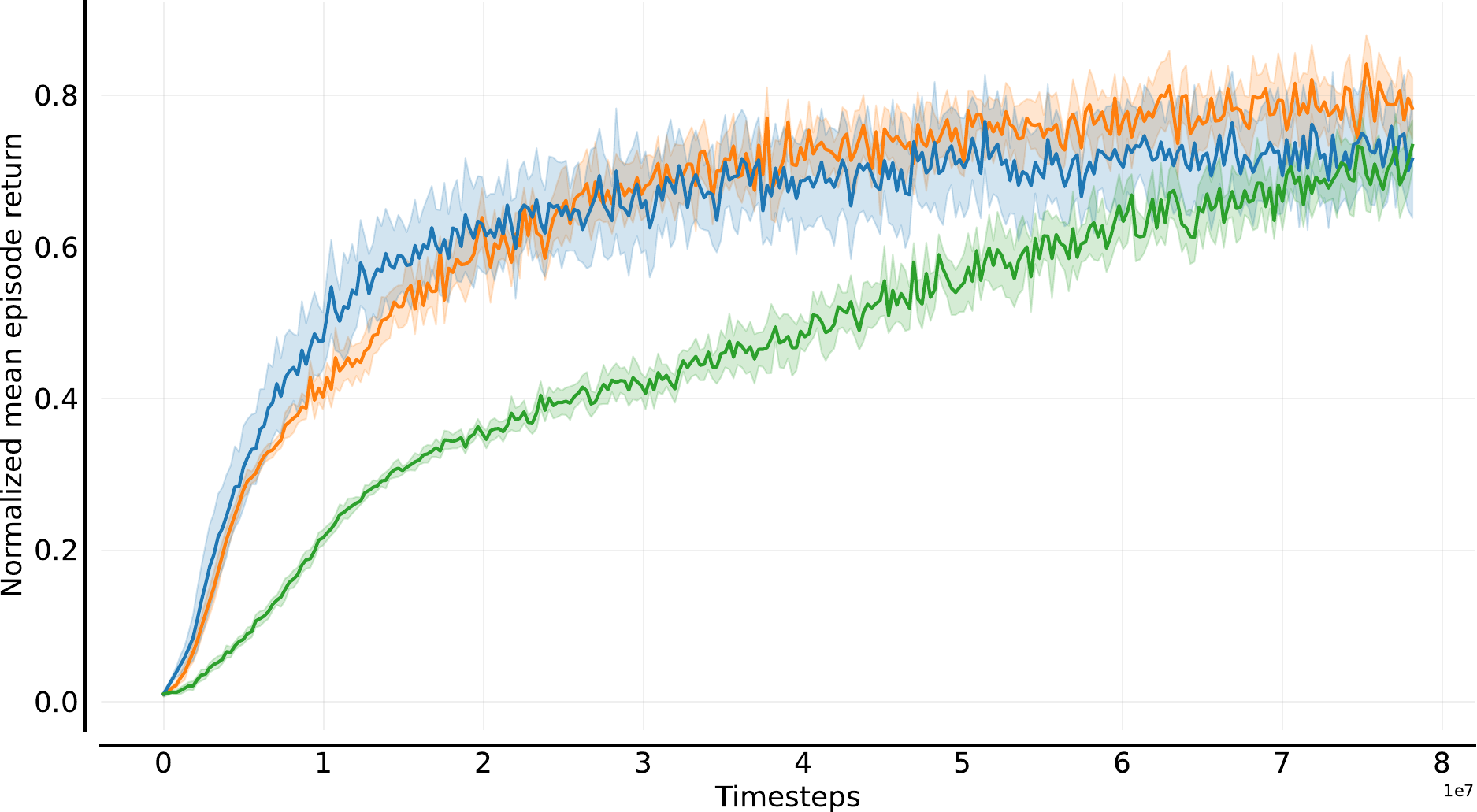}
    \caption*{\tiny SMAX 10m vs 11m}
\end{subfigure}\hfill
\begin{subfigure}{0.2\textwidth}
    \includegraphics[width=\linewidth]{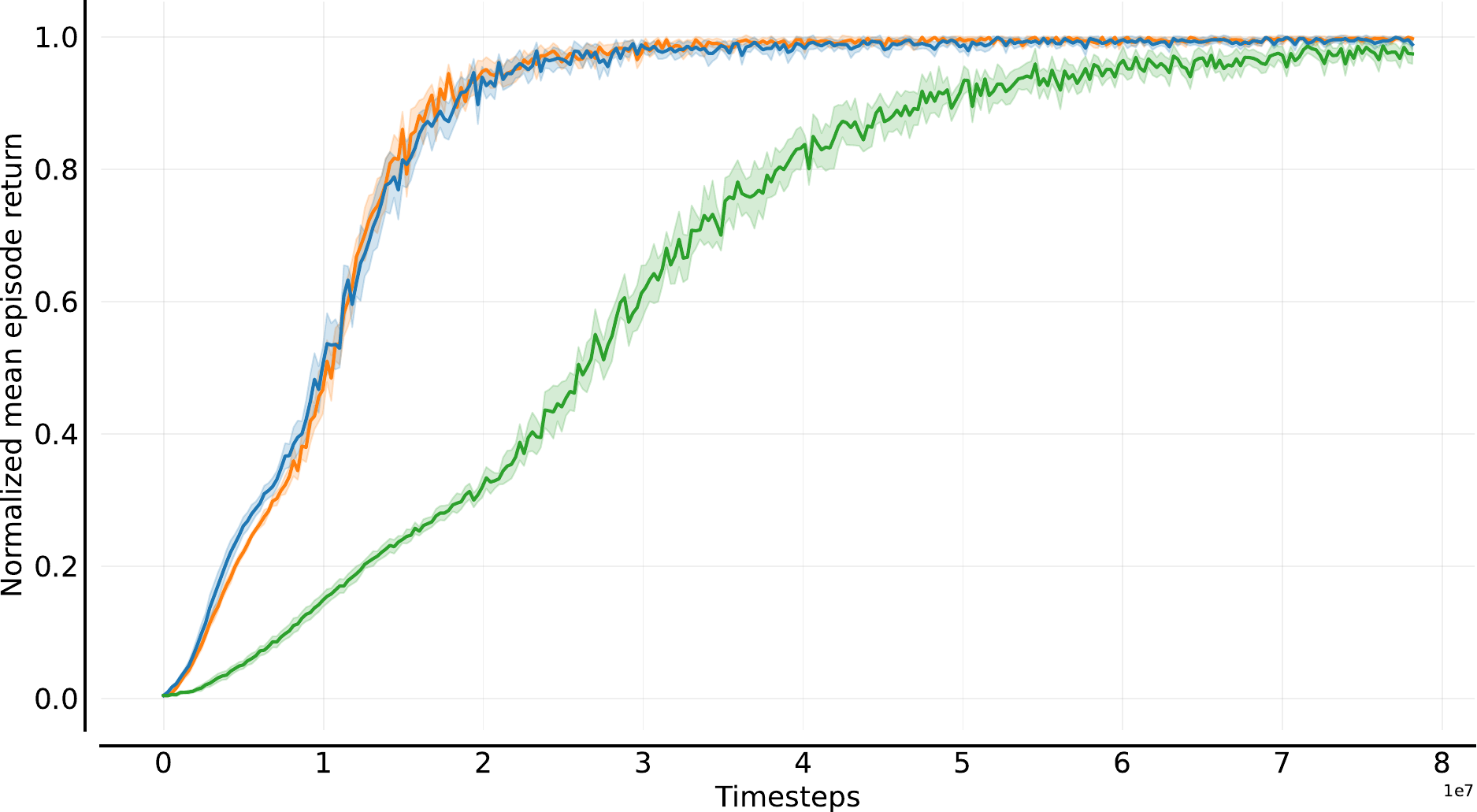}
    \caption*{\tiny SMAX 27m vs 30m}
\end{subfigure}\hfill
\begin{subfigure}{0.2\textwidth}
    \includegraphics[width=\linewidth]{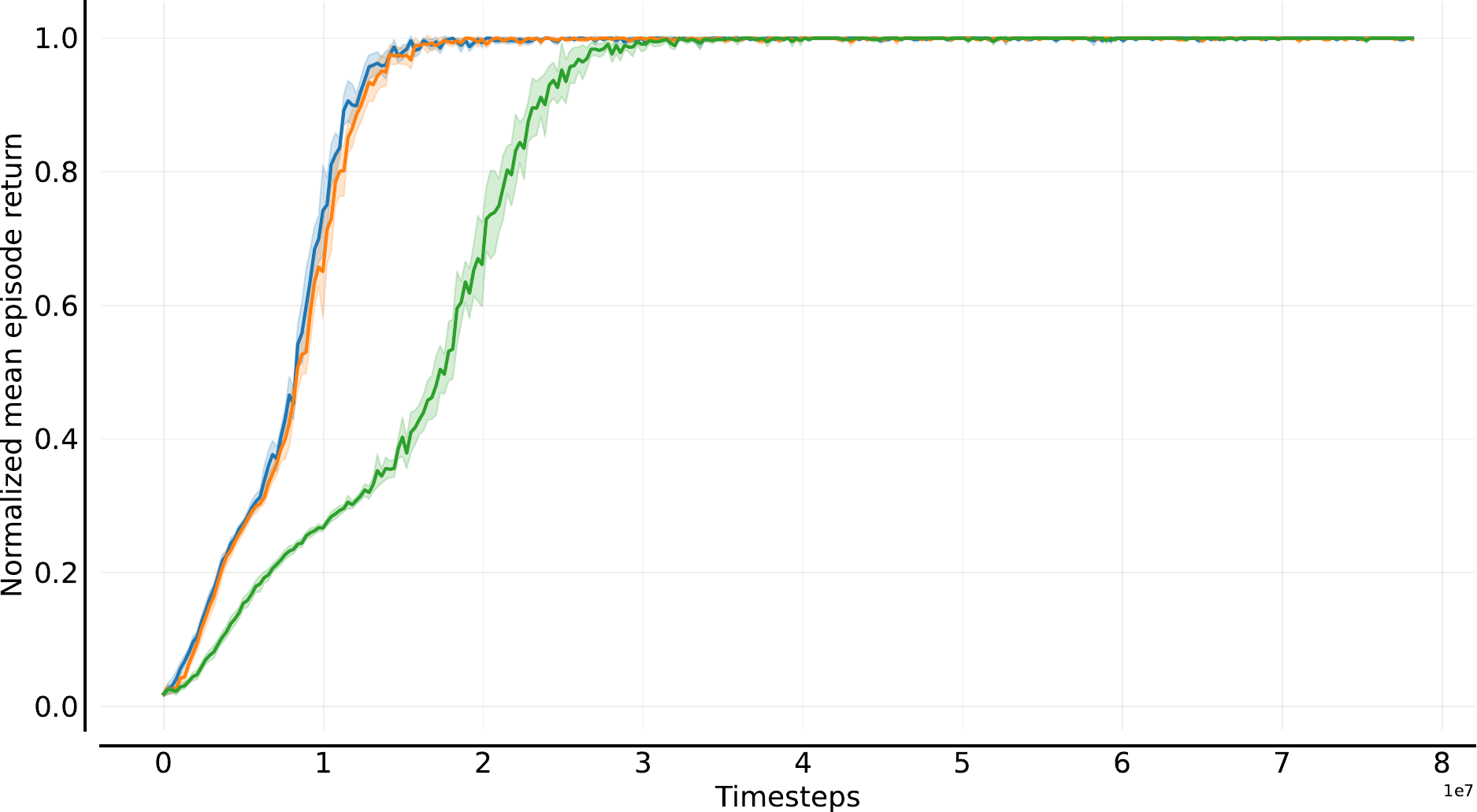}
    \caption*{\tiny SMAX 2s3z}
\end{subfigure}\hfill
\begin{subfigure}{0.2\textwidth}
    \includegraphics[width=\linewidth]{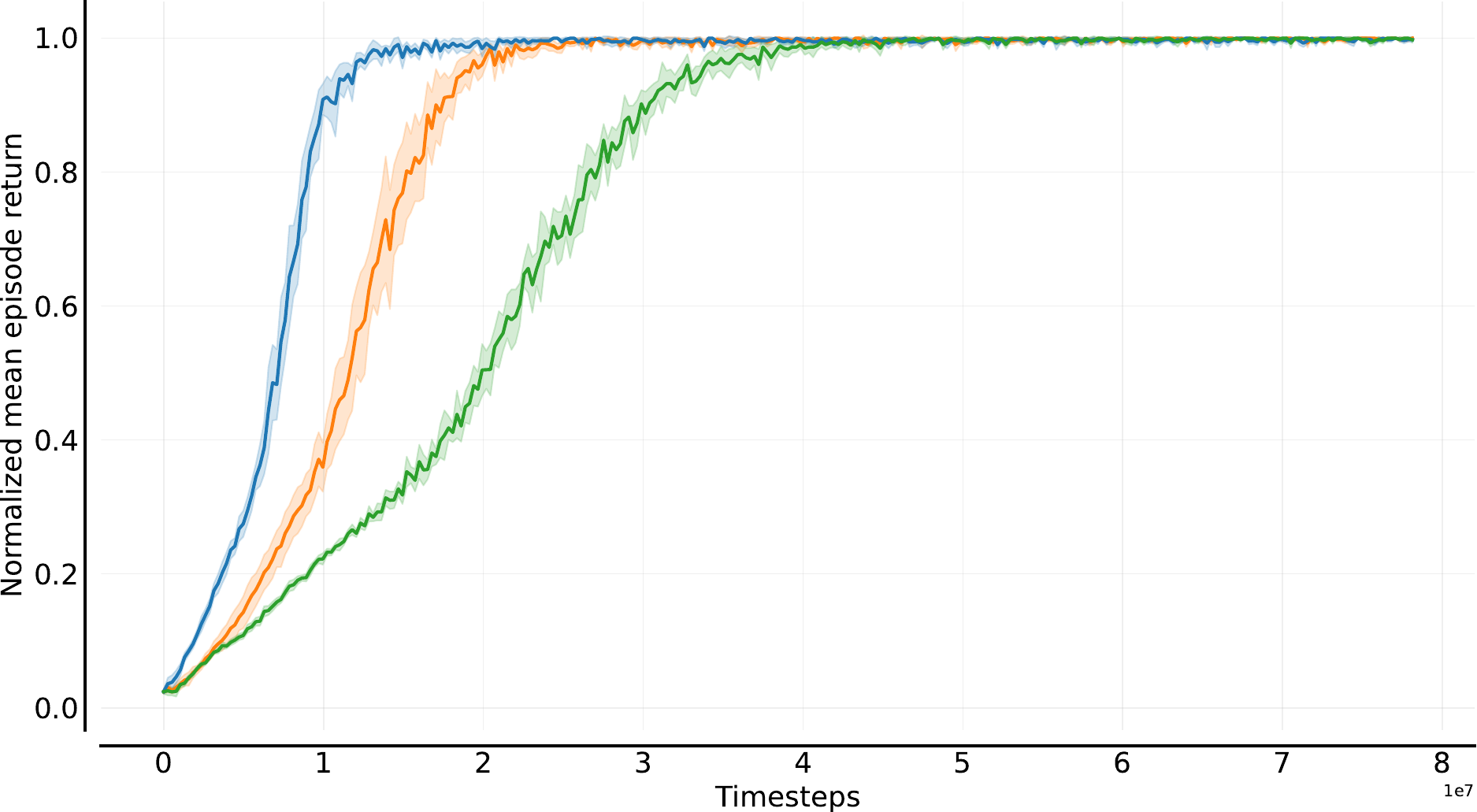}
    \caption*{\tiny SMAX 3s vs 5z}
\end{subfigure}\hfill
\par\smallskip 

\begin{subfigure}{0.2\textwidth}
    \includegraphics[width=\linewidth]{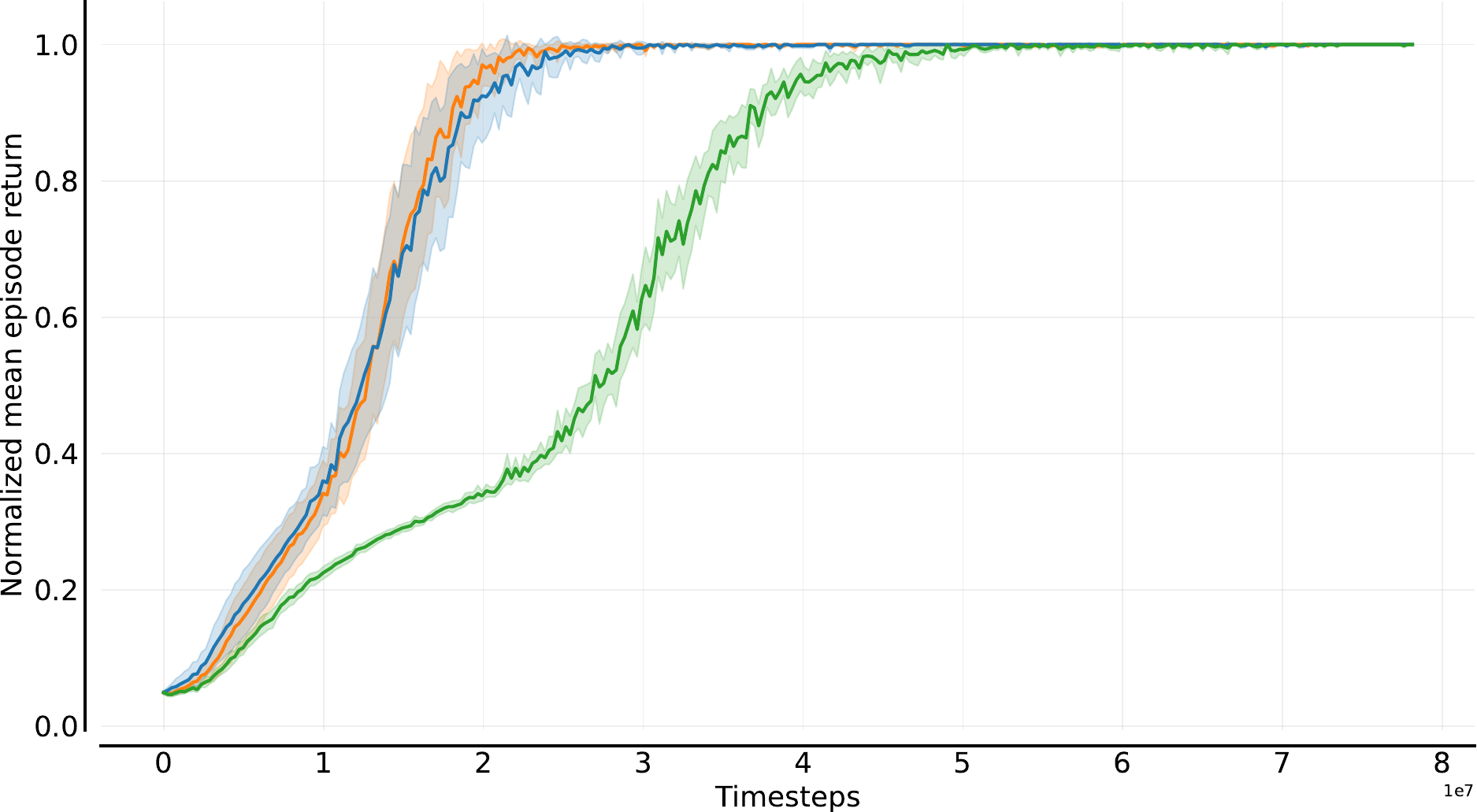}
    \caption*{\tiny SMAX 3s5z}
\end{subfigure}\hfill
\begin{subfigure}{0.2\textwidth}
    \includegraphics[width=\linewidth]{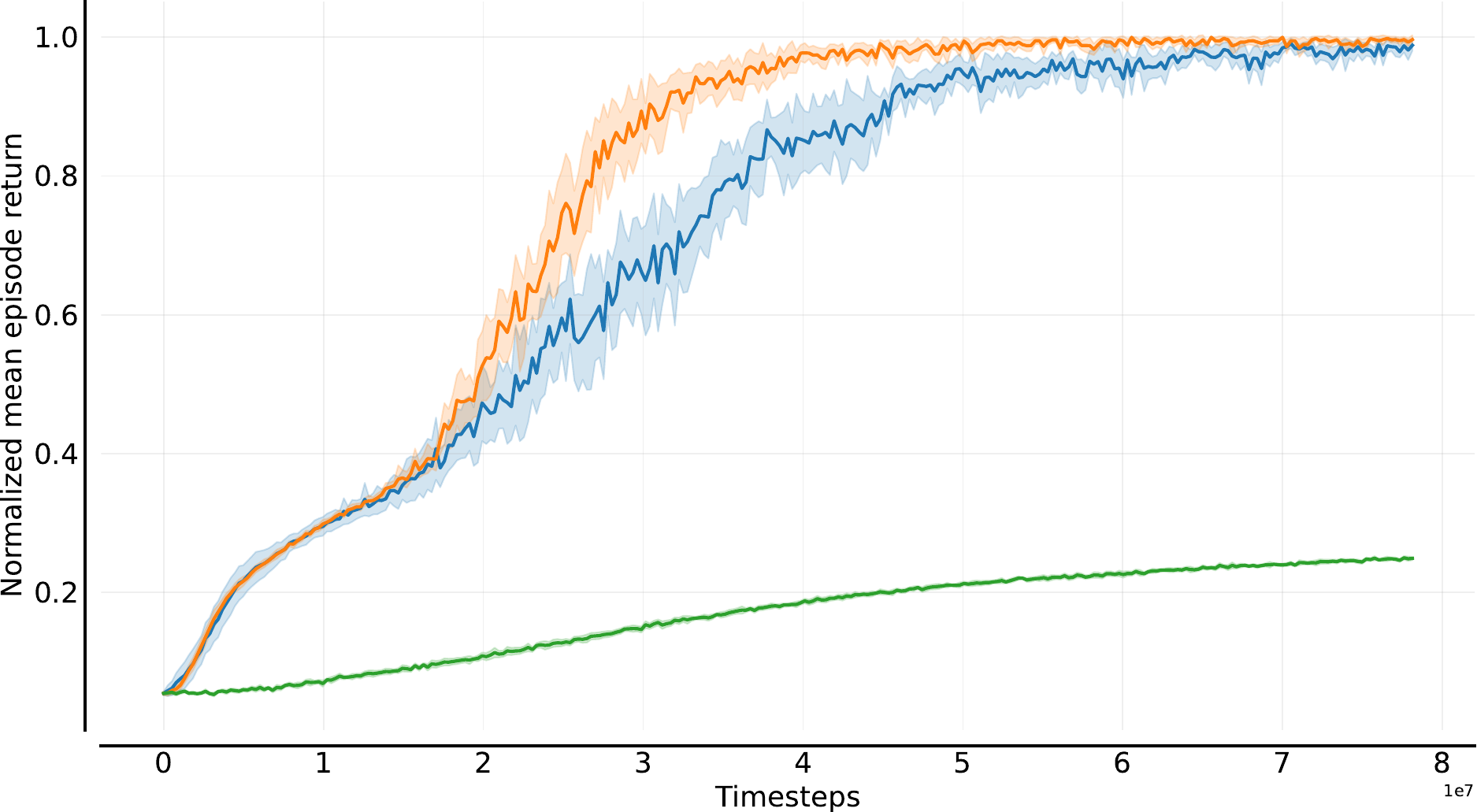}
    \caption*{\tiny SMAX 3s5z vs 3s6z}
\end{subfigure}\hfill
\begin{subfigure}{0.2\textwidth}
    \includegraphics[width=\linewidth]{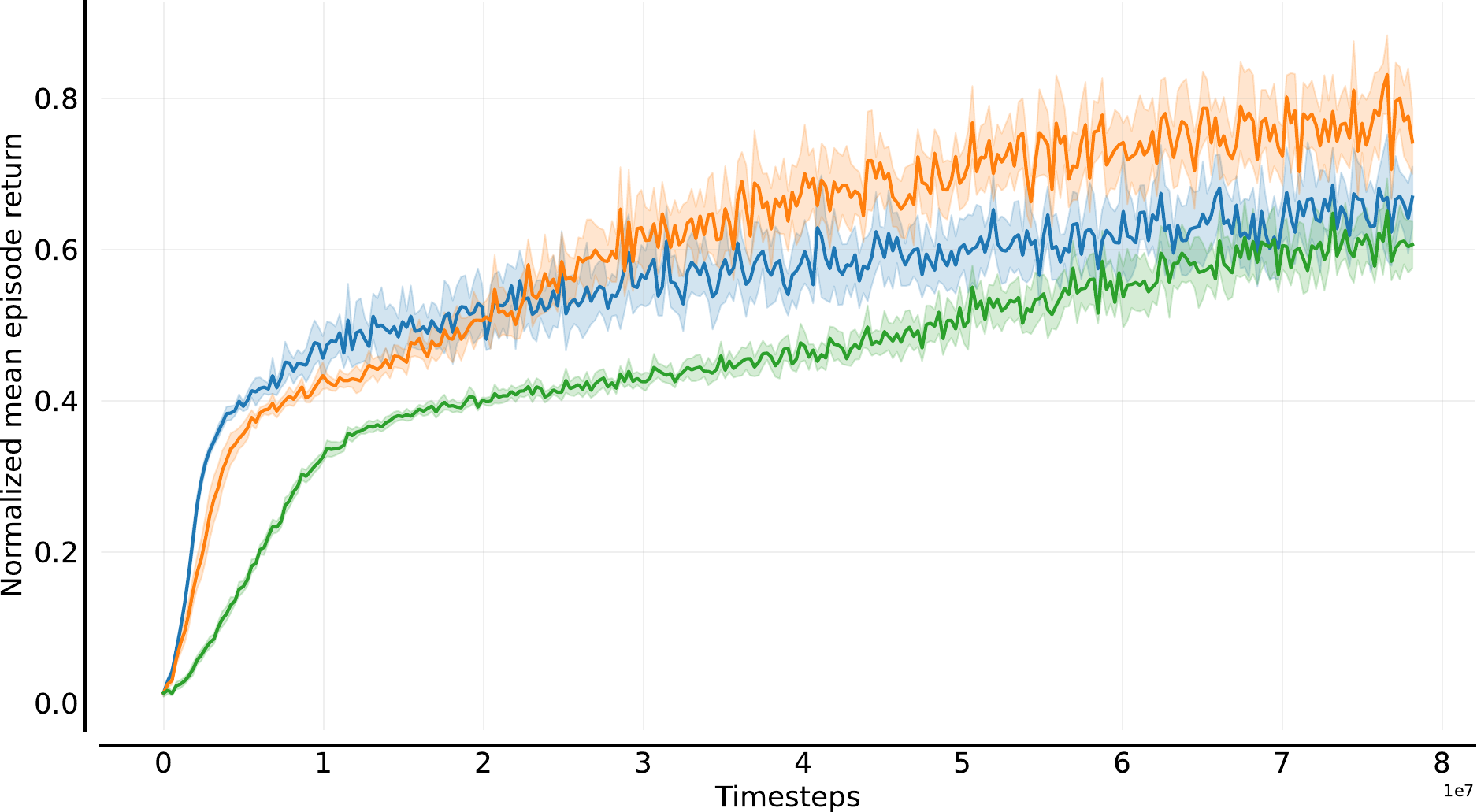}
    \caption*{\tiny SMAX 5m vs 6m}
\end{subfigure}\hfill
\begin{subfigure}{0.2\textwidth}
    \includegraphics[width=\linewidth]{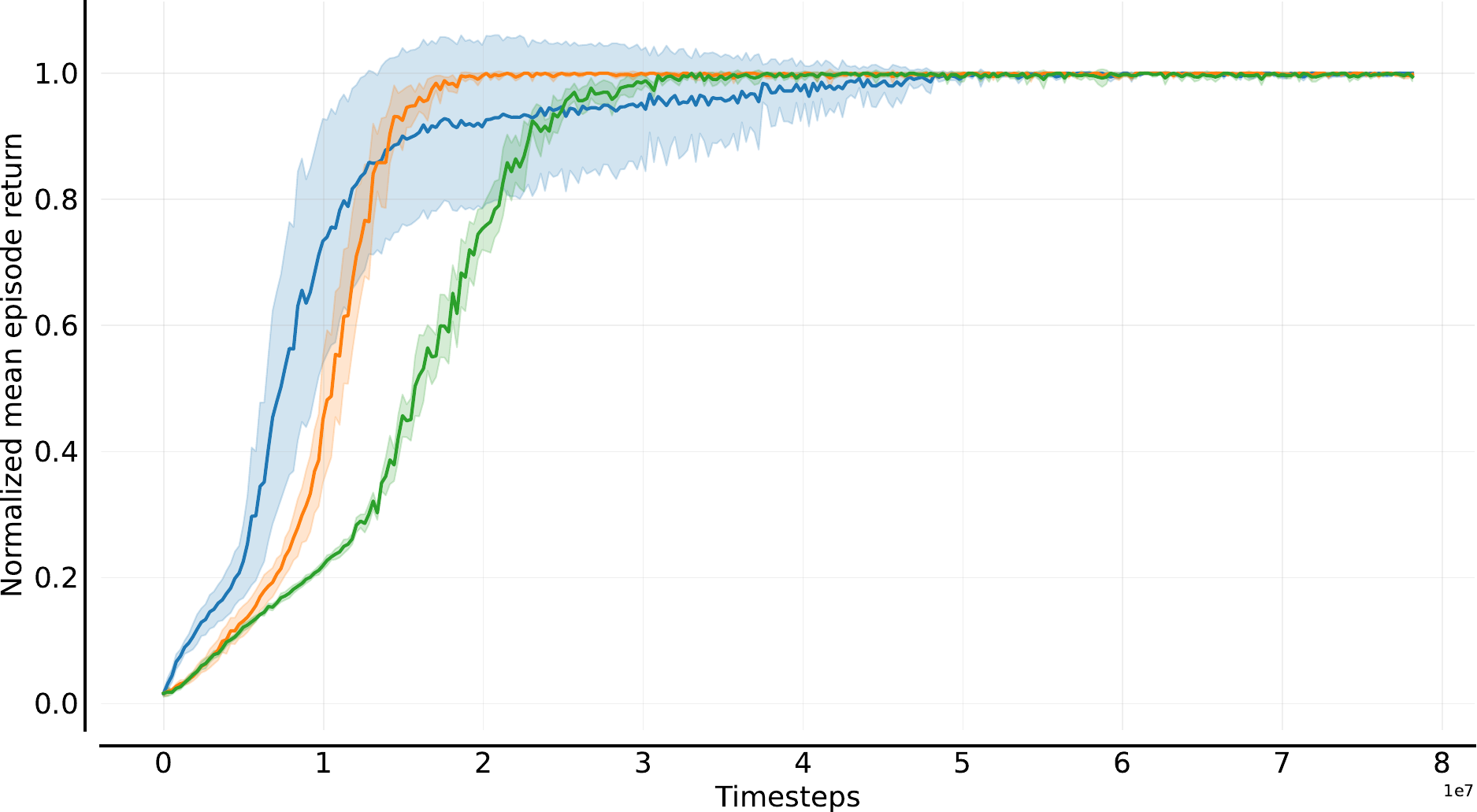}
    \caption*{\tiny SMAX 6h vs 8z}
\end{subfigure}\hfill
\begin{subfigure}{0.2\textwidth}
    \includegraphics[width=\linewidth]{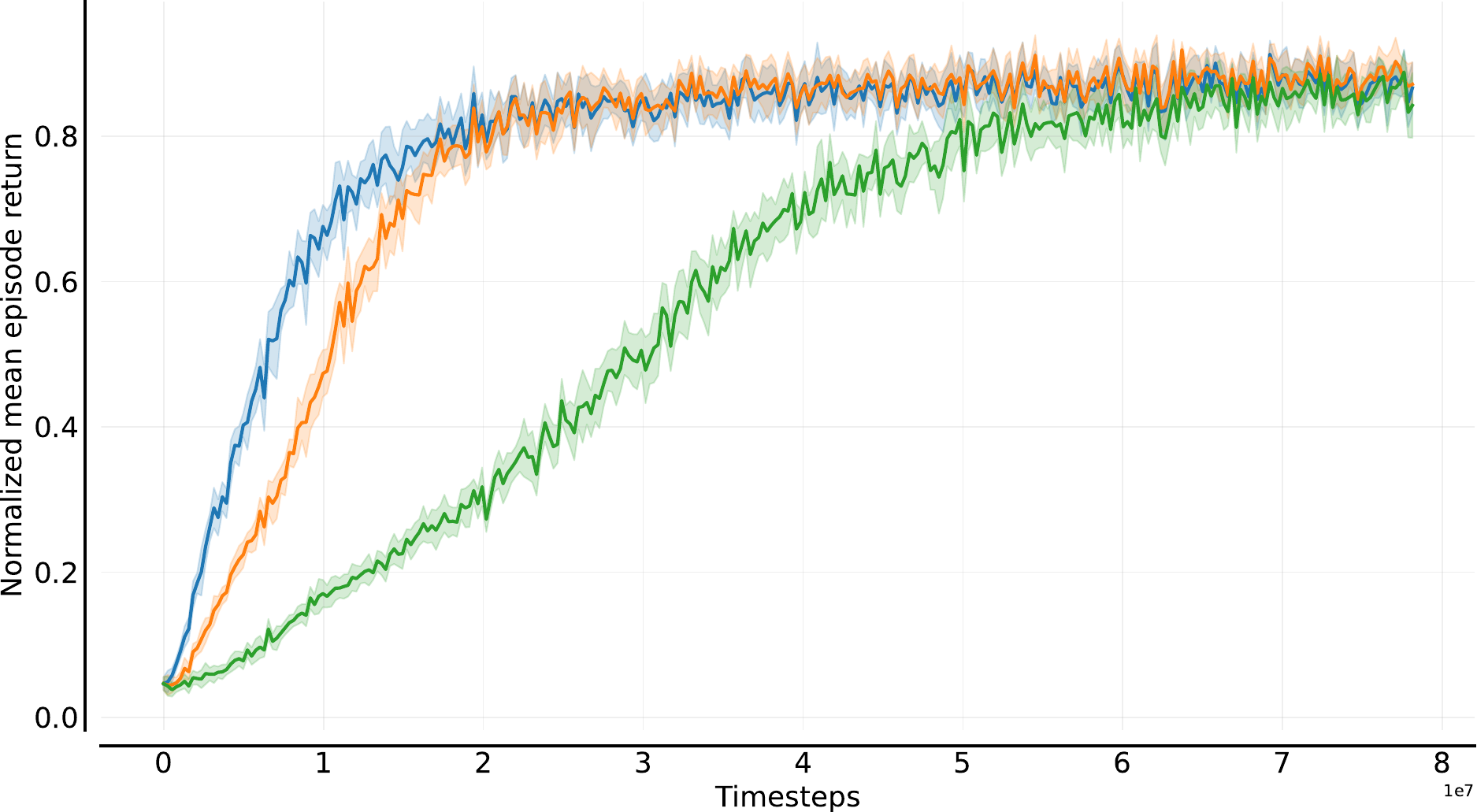}
    \caption*{\tiny SMAX SMACv2 10 Units}
\end{subfigure}\hfill
\par\smallskip 

\begin{subfigure}{0.2\textwidth}
    \includegraphics[width=\linewidth]{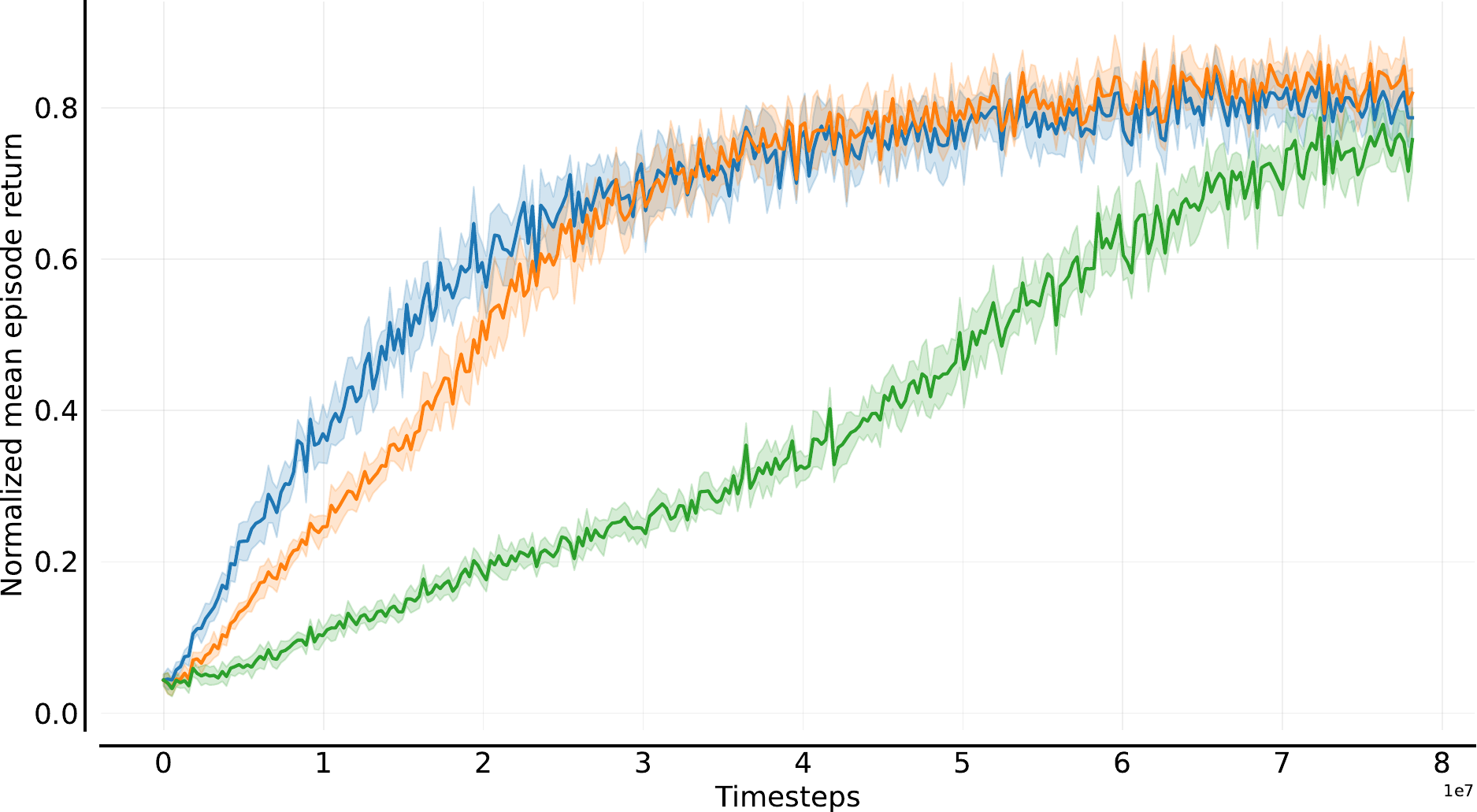}
    \caption*{\tiny SMAX SMACv2 20 Units}
\end{subfigure}\hfill
\begin{subfigure}{0.2\textwidth}
    \includegraphics[width=\linewidth]{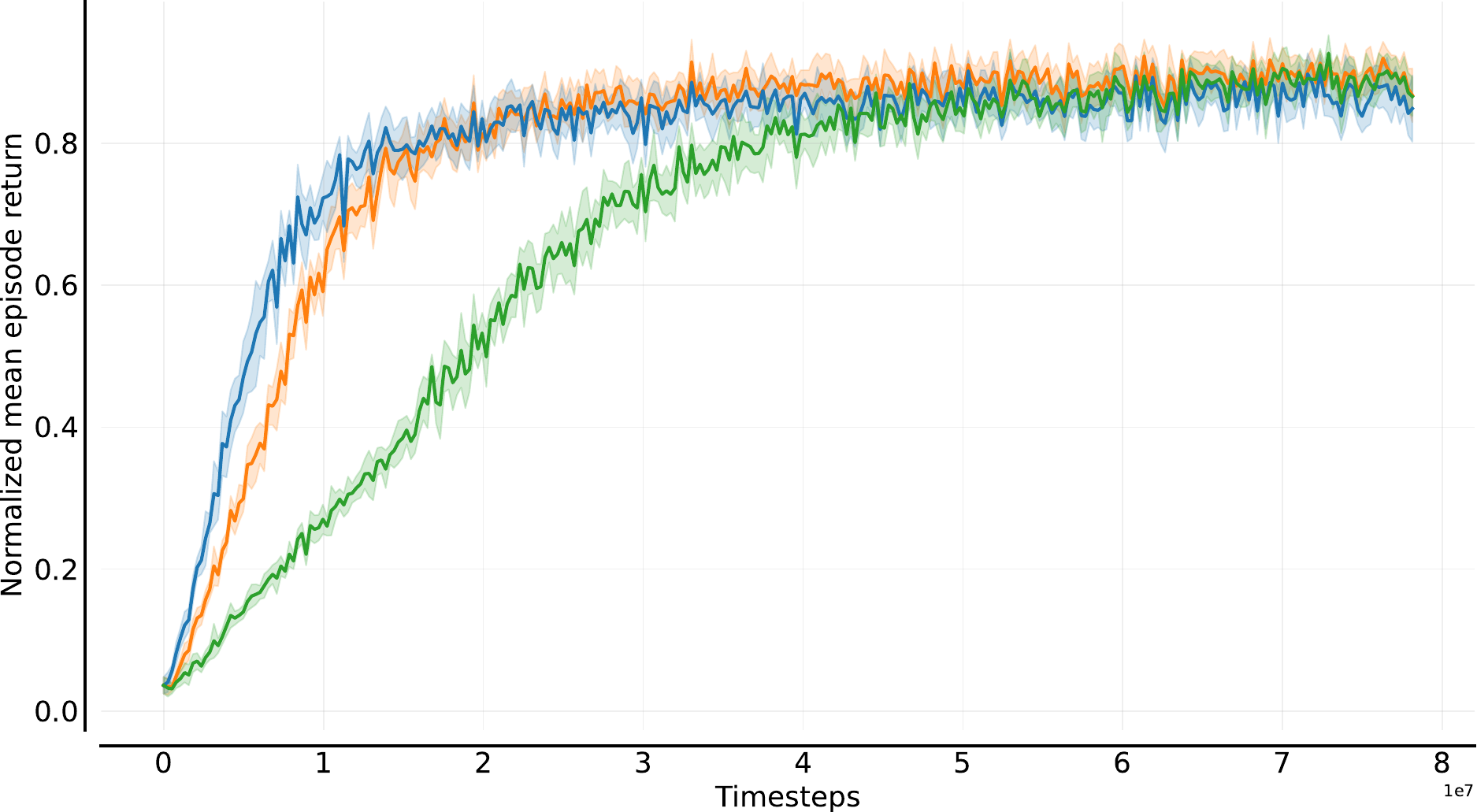}
    \caption*{\tiny SMAX SMACv2 5 Units}
\end{subfigure}\hfill
\begin{subfigure}{0.2\textwidth}
    \includegraphics[width=\linewidth]{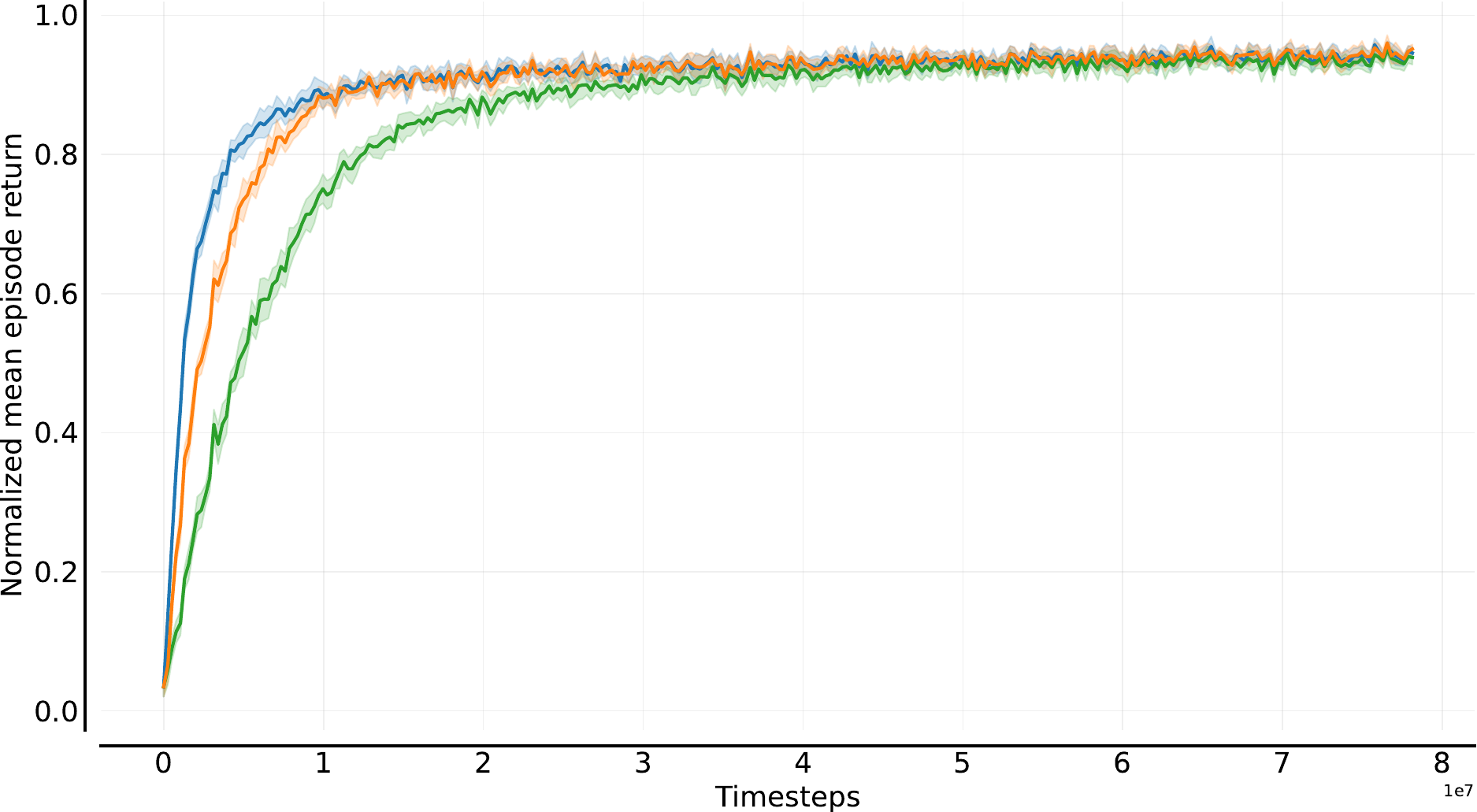}
    \caption*{\tiny Connector 10x10x10a}
\end{subfigure}\hfill
\begin{subfigure}{0.2\textwidth}
    \includegraphics[width=\linewidth]{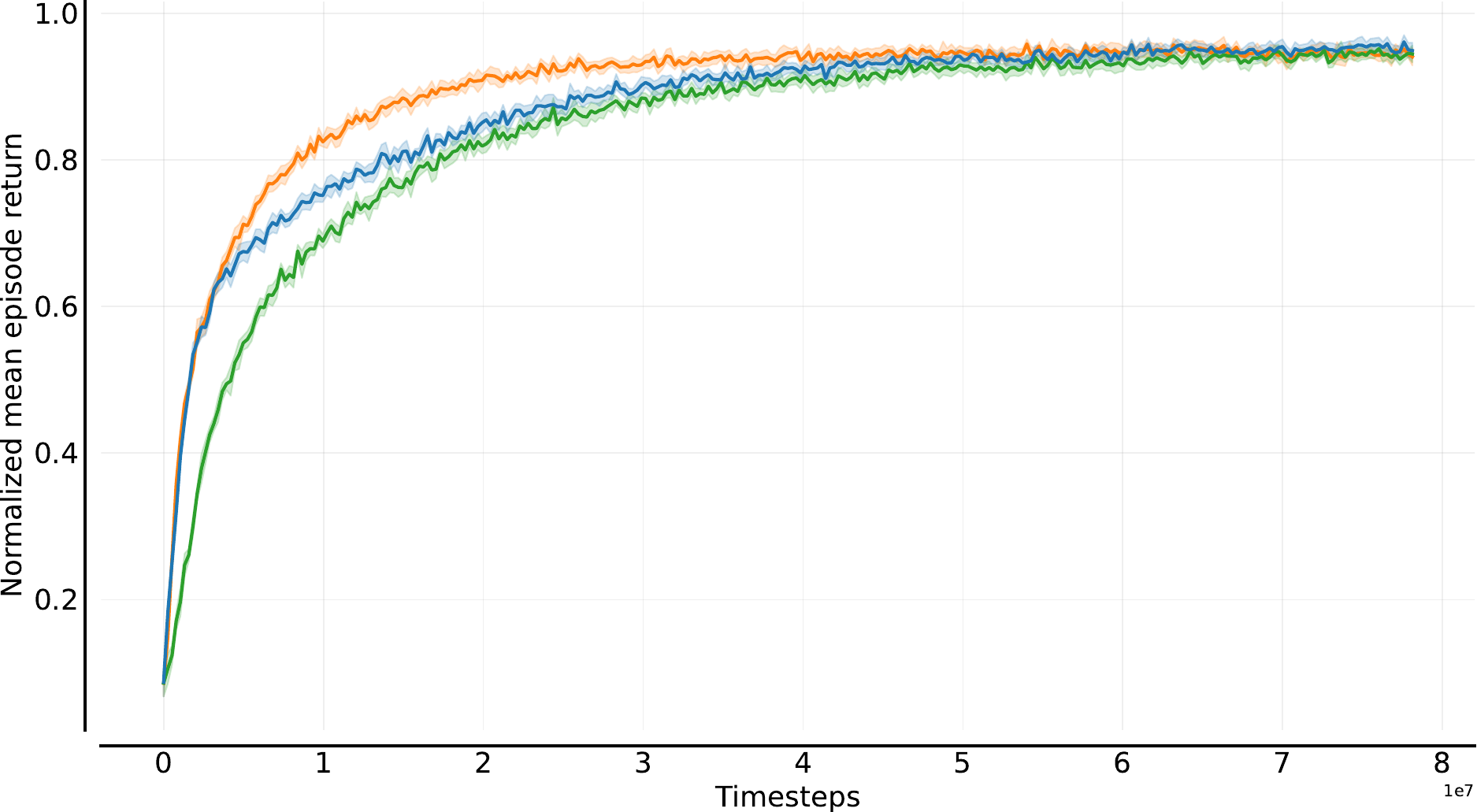}
    \caption*{\tiny Connector 15x15x23a}
\end{subfigure}\hfill
\begin{subfigure}{0.2\textwidth}
    \includegraphics[width=\linewidth]{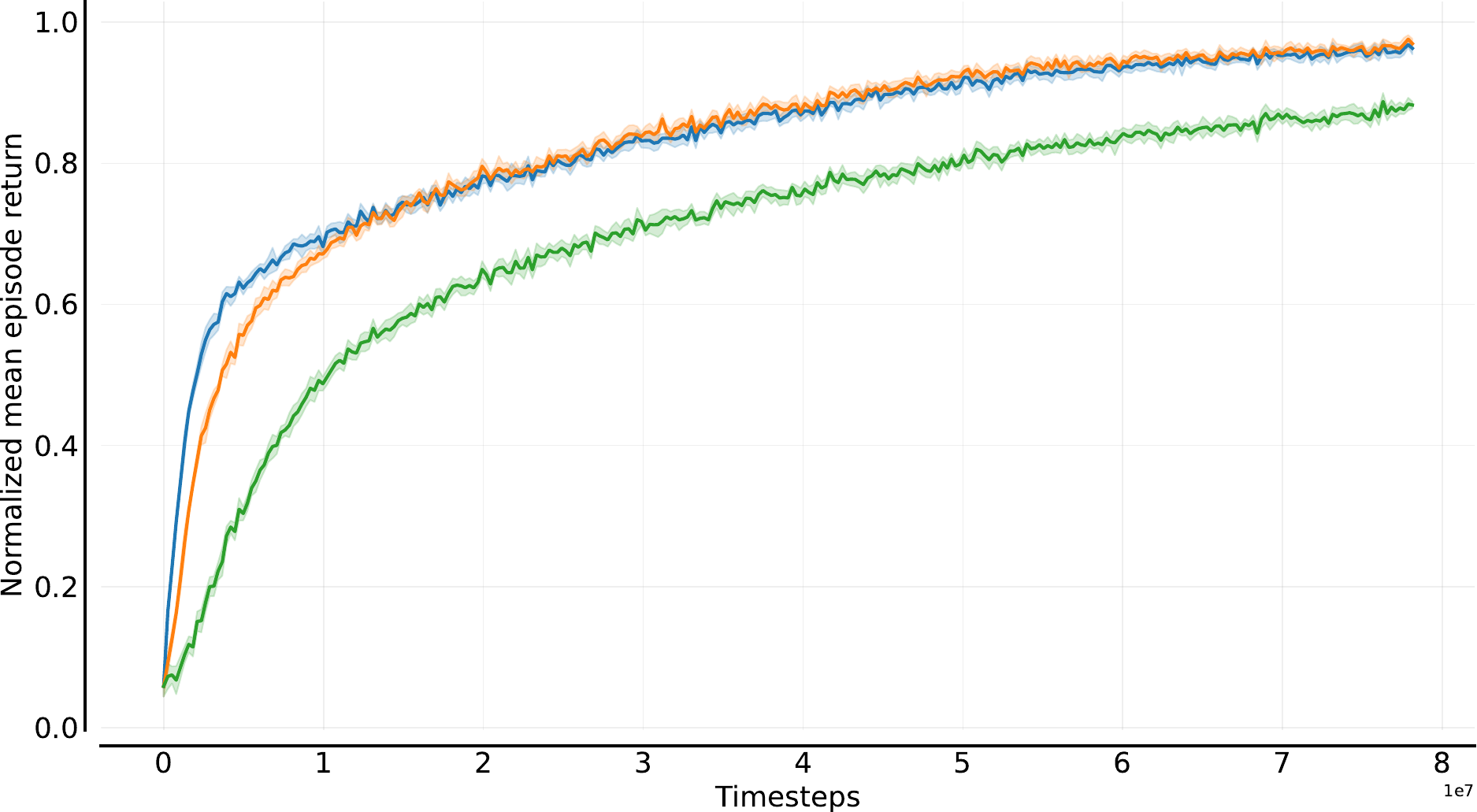}
    \caption*{\tiny Connector 18x18x33a}
\end{subfigure}\hfill
\par\smallskip 

\begin{subfigure}{0.2\textwidth}
    \includegraphics[width=\linewidth]{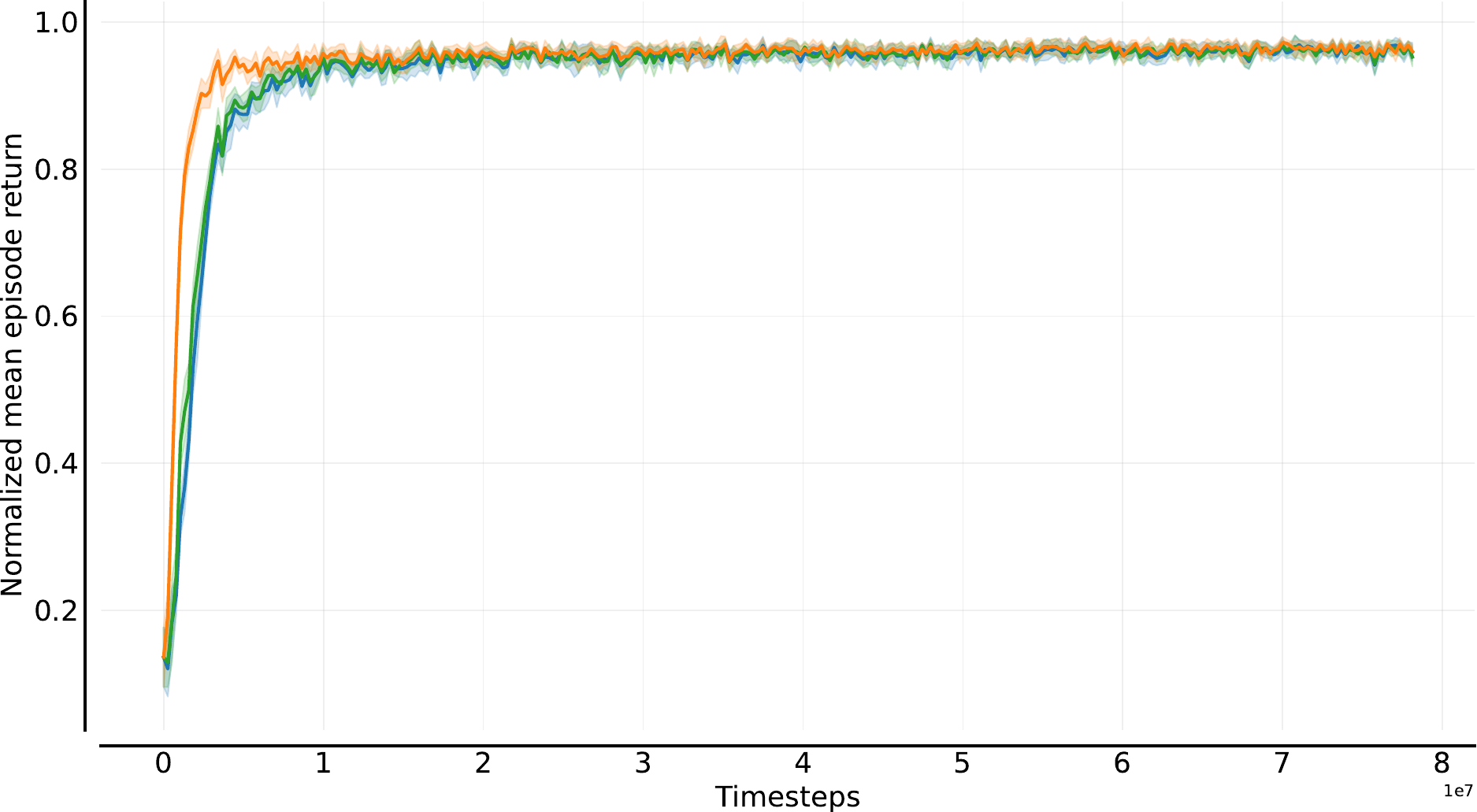}
    \caption*{\tiny Connector 5x5x3a}
\end{subfigure}\hfill
\begin{subfigure}{0.2\textwidth}
    \includegraphics[width=\linewidth]{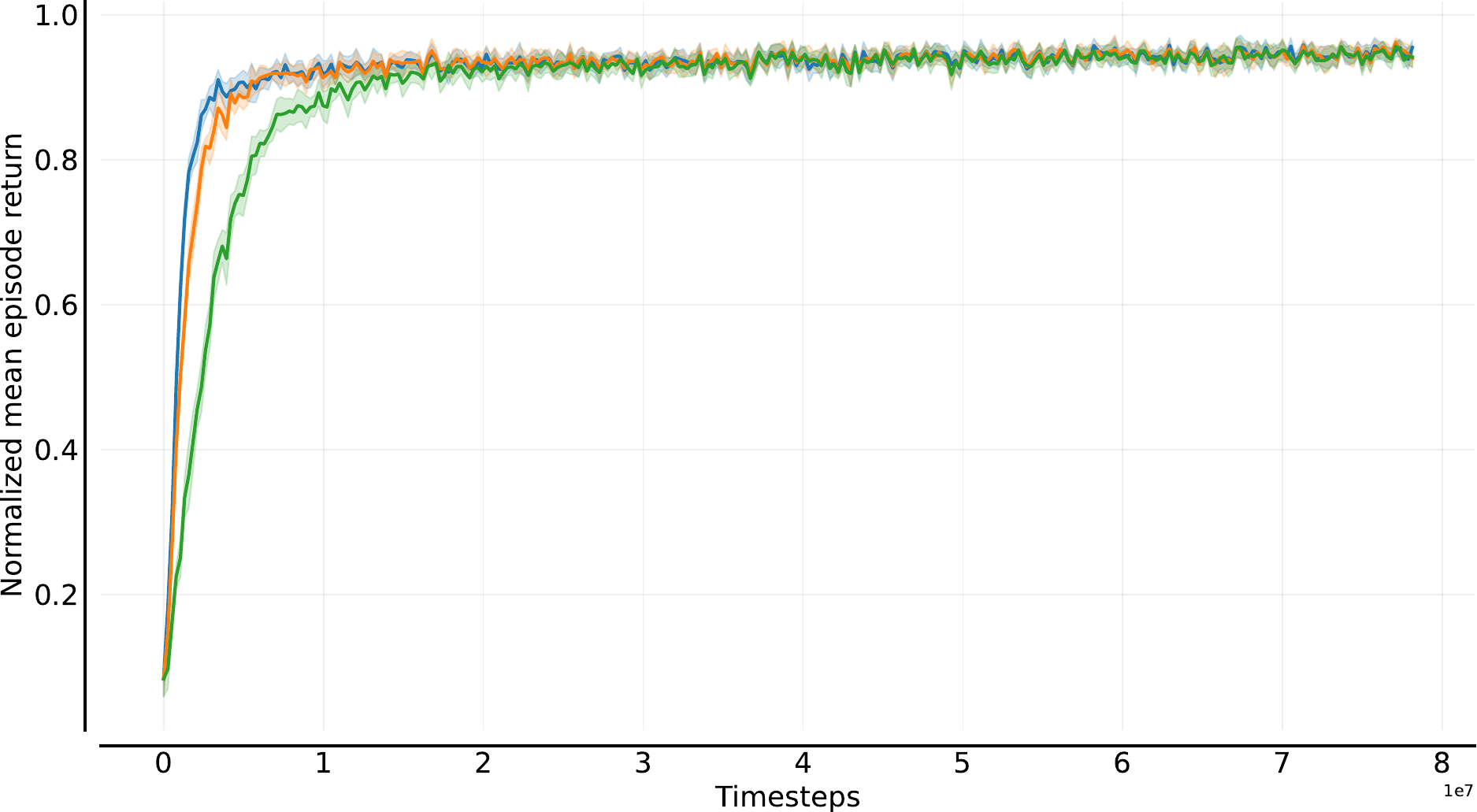}
    \caption*{\tiny Connector 7x7x5a}
\end{subfigure}\hfill
\begin{subfigure}{0.2\textwidth}
    \phantom{x} 
\end{subfigure}\hfill
\begin{subfigure}{0.2\textwidth}
    \phantom{x}
\end{subfigure}\hfill
\begin{subfigure}{0.2\textwidth}
    \phantom{x}
\end{subfigure}
\par\smallskip 

\par
\captionof{figure}{\textit{Per-task learning performance.} Plots depict the mean across 10 independent random seeds (shaded regions denote 95\% confidence intervals). Per-task min-max normalization is used.}
\label{fig:all_curves}


\newpage
\section*{NeurIPS Paper Checklist}

\begin{enumerate}

\item {\bf Claims}
    \item[] Question: Do the main claims made in the abstract and introduction accurately reflect the paper's contributions and scope?
    \item[] Answer: \answerYes{} 
    \item[] Justification: The abstract and introduction state the paper's main contributions: the MARS objective, its ratio-space properties, the PaxMen and AeroJAX benchmarks, and the 47-task empirical evaluation. The claims are scoped to cooperative CTDE multi-agent policy-gradient reinforcement learning and are supported by the formal surrogate-level results in Section~\ref{section:multi_agent_ratio_symmetry} and the experiments in Section~\ref{section:results}.

\item {\bf Limitations}
    \item[] Question: Does the paper discuss the limitations of the work performed by the authors?
    \item[] Answer: \answerYes{} 
    \item[] Justification: The paper discusses limitations in Appendix~\ref{appendix:limitations}, including the scope of the theoretical guarantees, benchmark coverage, target-ratio tuning, and open questions for broader multi-agent settings.

\item {\bf Theory assumptions and proofs}
    \item[] Question: For each theoretical result, does the paper provide the full set of assumptions and a complete (and correct) proof?
    \item[] Answer: \answerYes{} 
    \item[] Justification: The paper states the assumptions for each theoretical result in the proposition statements, including the positive ratio domain $r>0$, finite joint advantage estimate $\widehat{\mathbf{A}}$, and positive penalty strength $\alpha>0$. Full proofs are provided in Appendix~\ref{appendix:proofs_derivations}, including the MAPPO clipped-gradient derivation, the MARS penalty derivation, the extinction-barrier result, the no-gradient-truncation result, and the target-ratio alignment derivation.

    \item {\bf Experimental result reproducibility}
    \item[] Question: Does the paper fully disclose all the information needed to reproduce the main experimental results of the paper to the extent that it affects the main claims and/or conclusions of the paper (regardless of whether the code and data are provided or not)?
    \item[] Answer: \answerYes{} 
    \item[] Justification: The paper describes the experimental setup in Section~\ref{section:results}, including the evaluated algorithms, benchmark suite, tuning protocol, number of Optuna trials, shared policy-parameter setting, and evaluation scope. Appendix~\ref{appendix:section:hyperparameters} provides the hyperparameter search spaces, fixed hyperparameters, training lengths, and evaluation protocol, while Appendix~\ref{appendix:section:algorithm_pseudocode} provides pseudocode for MARS. Full code, compute details, and exact final tuned hyperparameters for all results are provided in the Supplementary Materials to assist in reproducibility.

\item {\bf Open access to data and code}
    \item[] Question: Does the paper provide open access to the data and code, with sufficient instructions to faithfully reproduce the main experimental results, as described in supplemental material?
    \item[] Answer: \answerYes{} 
    \item[] Justification: Full code, compute details, and exact final tuned hyperparameters for all results are provided in the supplementary materials to assist in reproducibility.

\item {\bf Experimental setting/details}
    \item[] Question: Does the paper specify all the training and test details (e.g., data splits, hyperparameters, how they were chosen, type of optimizer) necessary to understand the results?
    \item[] Answer: \answerYes{} 
    \item[] Justification: Section~\ref{section:results} summarizes the training setup, benchmark suite, baseline algorithms, tuning protocol, and evaluation design. Appendix~\ref{appendix:section:hyperparameters} specifies the hyperparameter search spaces, tuning budgets, fixed hyperparameters, optimizer choice, evaluation protocol, seed usage, and training lengths needed to understand the reported results. Final tuned hyperparameters for all experiments are provided in the supplementary materials.

\item {\bf Experiment statistical significance}
    \item[] Question: Does the paper report error bars suitably and correctly defined or other appropriate information about the statistical significance of the experiments?
    \item[] Answer: \answerYes{} 
    \item[] Justification: Section~\ref{section:results} reports 95\% confidence intervals for aggregate performance, learning curves, and probability-of-improvement estimates. The figure and table captions specify the uncertainty estimates used, including stratified bootstrap confidence intervals for aggregate metrics and confidence intervals across independent random seeds for learning curves, following the evaluation recommendations of \citet{agarwal2021deep}. Per-task learning curve results in Appendix \cref{appendix:section:per_task_performance} also report 95\% confidence intervals.

\item {\bf Experiments compute resources}
    \item[] Question: For each experiment, does the paper provide sufficient information on the computer resources (type of compute workers, memory, time of execution) needed to reproduce the experiments?
    \item[] Answer: \answerYes{} 
    \item[] Justification: The supplementary material reports accelerator configurations and compute records for the main experiments, and Appendix \cref{appendix:section:hyperparameters} reports tuning lengths, final evaluation lengths, and long-horizon ablation lengths. The full evaluation requires substantial compute because it spans 47 tasks, multiple random seeds, Optuna tuning, and billion-step ablations.
    
\item {\bf Code of ethics}
    \item[] Question: Does the research conducted in the paper conform, in every respect, with the NeurIPS Code of Ethics \url{https://neurips.cc/public/EthicsGuidelines}?
    \item[] Answer: \answerYes{} 
    \item[] Justification: The research conforms to the NeurIPS Code of Ethics. The work uses simulated reinforcement-learning environments, does not involve human subjects or private data, and the supplementary code is anonymized for review.

\item {\bf Broader impacts}
    \item[] Question: Does the paper discuss both potential positive societal impacts and negative societal impacts of the work performed?
    \item[] Answer: \answerYes{} 
    \item[] Justification: The paper studies foundational cooperative MARL optimization in simulation. Potential positive impacts include more stable training for cooperative multi-agent systems. Potential negative impacts are limited to downstream misuse of improved multi-agent coordination methods, particularly in simulated navigation or combat-inspired domains. The work does not release deployed systems, pretrained agents for real-world control, or human-subject data.
    
\item {\bf Safeguards}
    \item[] Question: Does the paper describe safeguards that have been put in place for responsible release of data or models that have a high risk for misuse (e.g., pre-trained language models, image generators, or scraped datasets)?
    \item[] Answer: \answerNA{} 
    \item[] Justification: The paper does not release high-risk pretrained models, scraped datasets, human-subject data, or generative models requiring controlled release safeguards. The supplementary material contains simulation environments, training code, and hyperparameter records.

\item {\bf Licenses for existing assets}
    \item[] Question: Are the creators or original owners of assets (e.g., code, data, models), used in the paper, properly credited and are the license and terms of use explicitly mentioned and properly respected?
    \item[] Answer: \answerYes{} 
    \item[] Justification: Existing assets are credited in the paper and supplementary archive. The archive includes license files for the modified Mava codebase and released environment packages, and a top-level notice identifies upstream projects and licenses used or adapted.

\item {\bf New assets}
    \item[] Question: Are new assets introduced in the paper well documented and is the documentation provided alongside the assets?
    \item[] Answer: \answerYes{} 
    \item[] Justification: The paper introduces the MARS implementation and the PaxMen and AeroJAX environment packages. The supplementary archive provides source code, package metadata, licenses, README documentation, hyperparameter records, and representative commands for running experiments with these assets.

\item {\bf Crowdsourcing and research with human subjects}
    \item[] Question: For crowdsourcing experiments and research with human subjects, does the paper include the full text of instructions given to participants and screenshots, if applicable, as well as details about compensation (if any)? 
    \item[] Answer: \answerNA{} 
    \item[] Justification: The work does not involve crowdsourcing, user studies, or human-subject experiments.

\item {\bf Institutional review board (IRB) approvals or equivalent for research with human subjects}
    \item[] Question: Does the paper describe potential risks incurred by study participants, whether such risks were disclosed to the subjects, and whether Institutional Review Board (IRB) approvals (or an equivalent approval/review based on the requirements of your country or institution) were obtained?
    \item[] Answer: \answerNA{} 
    \item[] Justification: The work does not involve crowdsourcing nor research with human subjects.

\item {\bf Declaration of LLM usage}
    \item[] Question: Does the paper describe the usage of LLMs if it is an important, original, or non-standard component of the core methods in this research? Note that if the LLM is used only for writing, editing, or formatting purposes and does \emph{not} impact the core methodology, scientific rigor, or originality of the research, declaration is not required.
    \item[] Answer: \answerNA{} 
    \item[] Justification: The core method development, theory, experiments, and environments do not use LLMs as a methodological component.

\end{enumerate}

\end{document}